\documentclass{article}

\usepackage[preprint]{neurips_2026}

\usepackage[utf8]{inputenc} % allow utf-8 input
\usepackage[T1]{fontenc}    % use 8-bit T1 fonts
\usepackage{hyperref}       % hyperlinks
\usepackage{url}            % simple URL typesetting
\usepackage{booktabs}       % professional-quality tables
\usepackage{amsfonts, amsmath, amsfonts}       % blackboard math symbols
\usepackage{nicefrac}       % compact symbols for 1/2, etc.
\usepackage{microtype}      % microtypography
\usepackage{xcolor}         % colors
\usepackage{graphicx}         % plots
\usepackage{wrapfig}         % fig side to text
\usepackage{subcaption}
\usepackage{fontawesome5}

\newcount\Comments  % 0 suppresses notes to selves in text
\definecolor{darkgreen}{rgb}{0,0.5,0}
\definecolor{purple}{rgb}{1,0,1}

\newcommand{\mynote}[3]{\ifnum\Comments=1\fbox{\bfseries\sffamily\color{#3} {#1}}{\small\textsf{\emph{\color{#3}{ #2}}}}\fi}

\newcommand{\takeaway}[1]{%
    \vspace{0.2ex}\noindent\fbox{\textbf{\textbf{Takeaway:}}} {#1}\vspace{0.2ex}%
}

\title{Measuring Cross-Modal Synergy: A Benchmark for VLM Explainability}

\author{%
  Joël Roman Ky\textsuperscript{1} \quad
  Salah Ghamizi\textsuperscript{1, 2} \quad
  Maxime Cordy\textsuperscript{1} \\
  \textsuperscript{1}University of Luxembourg, Luxembourg \quad \\
  \textsuperscript{2}Luxembourg Institute of Health (LIH), Luxembourg \quad \\
  \texttt{joel.ky@uni.lu}, \texttt{salah.ghamizi@uni.lu} , \texttt{maxime.cordy@uni.lu}
}

\begin{document}

\maketitle

\begin{abstract}
% Sentence 1: Topic
Vision-Language Models (VLMs) map complex visual inputs to semantic spaces, but interpreting the cross-modal reasoning of VLMs currently relies on post-hoc explainers evaluated via unimodal perturbation metrics. 
We expose a limitation in this paradigm: because multimodal datasets contain language priors and modality biases, VLMs frequently exhibit cross-modal redundancy, allowing them to answer visual queries using text alone. Consequently, unimodal metrics penalize faithful explainers, triggering an evaluation collapse where visual and textual rankings fundamentally contradict each other. %(Kendall's $\tau = -0.06$). 
To resolve this, we introduce Synergistic Faithfulness ($\mathcal{F}_{syn}$), a scalable metric rooted in the Shapley Interaction Index that strictly isolates the joint Harsanyi dividend between modalities, serving as a highly accurate surrogate ($\rho = 0.93$) while achieving up to $88\times$ computational speedup. 
Evaluating 9 distinct XAI methods across 3 VLM architectures and 3 benchmark datasets, reveals that explainers proposed for VLMs heavily over-index on visual salience and significantly underperform adapted attention-based methods in capturing true cross-modal synergy. 
By decoupling visual plausibility from cross-modal faithfulness, this work provides a rigorous evaluation framework required to safely audit VLM reasoning in high-stakes deployments. 
%Our code for the metric and the benchmark have been released at \url{https://github.com/serval-uni-lu/xai-vlms-benchmark}

\begin{center}
    \faGithub \hspace{0.5em} \url{https://github.com/serval-uni-lu/xai-vlms-benchmark/}
\end{center}

\end{abstract}

\section{Introduction}\label{introduction}

%\joel{Repasse sur tout le papier pour bien distinguer XAI, VLM, metric}

Vision-Language Models (VLMs) have achieved unprecedented capabilities by mapping high-dimensional visual inputs into the rich semantic space of Large Language Models (LLMs) \cite{wang2025internvl3, comanici2025gemini, team2026qwen3}. As these models are increasingly deployed in high-stakes real-world domains \cite{li2024mmro, cui2024survey, xiao2025comprehensive}, understanding their complex cross-modal reasoning has transitioned from a theoretical curiosity to a regulatory imperative. Frameworks such as the EU AI Act explicitly mandate traceability and the "right to explanation" for high-risk AI systems. Meeting these regulatory demands requires a crucial distinction between \textit{interpretability}, which consists in designing models that are inherently mathematically transparent and \textit{explainability} (XAI), which provides post-hoc insights into opaque "black box" algorithms. Given the massive scale and non-linear complexity of modern VLMs, inherent interpretability remains largely elusive, making post-hoc XAI the primary mechanism for auditing model behavior. Researchers have consequently proposed a vast array of XAI methods, ranging from classical gradient approaches \cite{sundararajan2017axiomatic} to newly proposed VLM-native explainers \cite{li2025token}. However, deploying these explainers safely hinges entirely on rigorous evaluation. One must be able to prove that an explanation is mathematically \textit{faithful} to the VLM's actual internal logic, rather than just visually plausible to a human observer. If evaluation metrics are flawed, there is a risk of deploying dangerously ungrounded models masked by convincing, yet false, explanations.

\begin{wrapfigure}{R}{0.55\textwidth} % [t] forces it to the top of the page
    \vspace{-0.5cm}
    \centering
    
    % --- ROW 1: THE OR-GATE (FAILURE CASE) ---
    \begin{subfigure}[b]{0.58\linewidth}
        \centering
        \includegraphics[width=\linewidth]{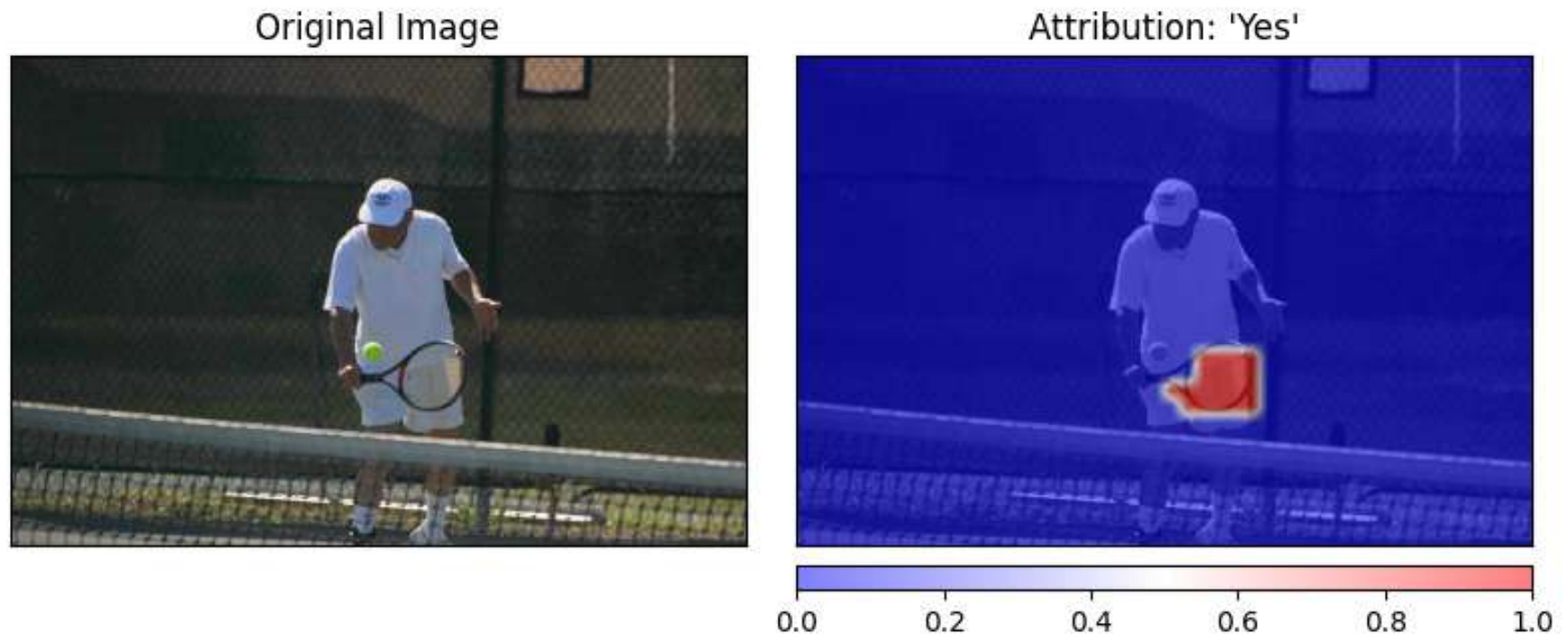} % Replace with your filename
        \caption{"Is there a tennis racket?"}
        \label{fig:motivating_or_heatmap}
    \end{subfigure}
    \hfill
    \begin{subfigure}[b]{0.39\linewidth}
        \centering
        \includegraphics[width=\linewidth]{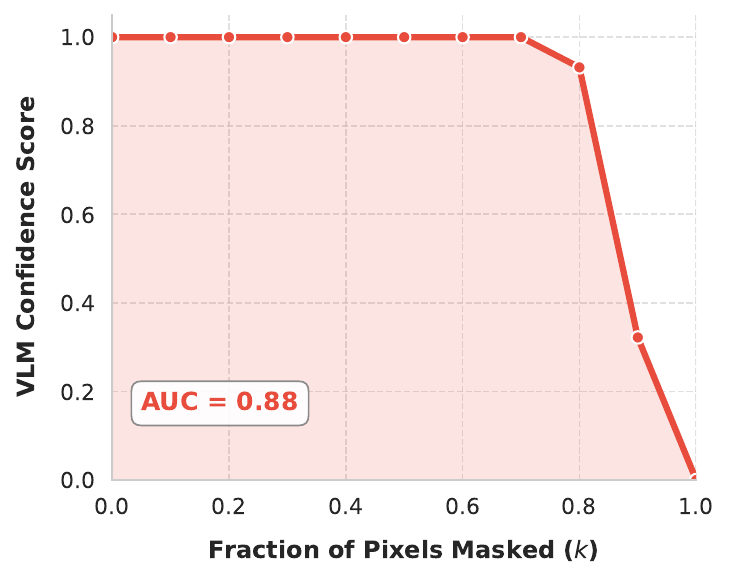} % Replace with your filename
        \caption{Unimodal Deletion}
        \label{fig:motivating_or_curve}
    \end{subfigure}

    % \vspace{0.6cm} % Adds breathing room between the two rows

    % % --- ROW 2: THE AND-GATE (SUCCESS CASE) ---
    % \begin{subfigure}[b]{0.58\linewidth}
    %     \centering
    %     \includegraphics[width=\linewidth]{figs/or_36.pdf} % Replace with your filename
    %     \caption{"Is there a frisbee in the image?"}
    %     \label{fig:motivating_and_heatmap}
    % \end{subfigure}
    % \hfill
    % \begin{subfigure}[b]{0.38\linewidth}
    %     \centering
    %     \includegraphics[width=\linewidth]{figs/deletion_curve_or_gate_36.pdf} % Replace with your filename
    %     \caption{Unimodal Deletion}
    %     \label{fig:motivating_and_curve}
    %     %\vspace{-0.5cm}
    % \end{subfigure}
    
    \caption{Limitations of unimodal faithfulness metrics using an oracle explainer (i.e., the ground-truth human bounding box). On this instance, the unimodal deletion curve remains nearly flat despite perfect visual grounding. This results in a penalized score, as a higher AUC in deletion metrics incorrectly implies a poor explanation.}
    \label{fig:motivating_example}
\end{wrapfigure}

Historically, the prevailing assumption in multimodal explainability has been that the faithfulness of an explainer can be evaluated using standard unimodal perturbation. Metrics like Insertion and Deletion Area Under the Curve (AUC) \cite{petsiuk2018rise} assume that if an explainer accurately identifies important input features (visual patches or text tokens), systematically masking those features will cause a rapid degradation in VLM confidence. While highly effective for pure vision or language models, applying this unimodal metric to highly entangled VLMs introduces a vulnerability. 
Indeed, because multimodal datasets frequently contain severe language priors and modality biases \cite{chen2024we, li2024naturalbench, zheng2025mllms, chen2024quantifying, lee2025vlind, liu2024insight}, VLMs often exhibit high cross-modal redundancy \cite{li2023evaluating, tong2024eyes}. 
Therefore, deleting the important visual patches identified by an explainer will not drop VLM confidence if the prompt text contains sufficient residual information  as demonstrated in Fig. \ref{fig:motivating_example} (here the word "tennis racket"). This paradox makes standard unimodal metrics unable to distinguish perfectly faithful explainers from random explainers, and triggers an evaluation collapse: our empirical analysis reveals that evaluating the same set of XAI methods on the same dataset yields completely contradictory rankings depending on whether visual or textual perturbation is used. Thus, current benchmarking heavily measures an explainer's susceptibility to dataset-specific modality biases rather than its true algorithmic faithfulness.

To resolve this, we advocate a transition from measuring isolated feature importance to measuring \textit{cross-modal synergy}. We introduce the Synergistic Faithfulness metric ($\mathcal{F}_{syn}$), a novel surrogate for the Shapley Interaction Index (SII), the theoretical gold standard in cooperative game theory for measuring player interaction (here, a player is the features of a given modality). This metric explicitly computes the 2-way Harsanyi dividend between visual and textual features. By approximating this joint interaction across a continuous perturbation trajectory, $\mathcal{F}_{syn}$ circumvents the intractable $\mathcal{O}(2^{N})$ ($N$ the number of players) combinatorial complexity of exact SII computation, while maintaining a strong Spearman correlation $\rho = 0.92$ with SII. Our approach thus provides a rigorous surrogate for exact Shapley interaction while achieving up to $88\times$ empirical speedup, allowing the metric to scale efficiently to modern, high-resolution VLMs. 

Equipped with this metric, we conduct an evaluation of state-of-the-art multimodal XAI methods, encompassing 3 benchmarks across 3 prominent VLM architectures (LLaVA-1.5, Qwen2.5-VL, and InternVL-3.5) and 9 distinct explainers. To compare these explainers on a fair ground, we employ a Linear Mixed-effects Model (LMM) to statistically isolate explainer performance from dataset and model-induced variance. Our results demonstrate the superiority of attention-based explainers, with AttnLRP\cite{achtibat2024attnlrp} achieving the highest statistically significant coefficients ($\beta=0.029$), ahead SoTA VLM-native approaches (LLaVA-CAM \cite{zhang2025redundancy}, TAM \cite{li2025token}) with an LMM coefficient $\beta=0.008$.

Our findings challenge the prevailing assumption that newly designed "VLM-native" explainers are inherently superior. Instead, we reveal that these native methods over-index on visual salience, failing to capture true mathematical cross-modal synergy when compared to adapted LLM-based algorithms. 
In summary, our primary contributions and empirical insights are:
{\setlength{\leftmargini}{1.5em}%
\setlength{\labelsep}{1em}%
\begin{itemize}
    \item We demonstrate the failure of unimodal faithfulness metrics in multimodal contexts, proving that cross-modal redundancy generates statistically unstable and contradictory explainer rankings.
    \item We propose Synergistic Faithfulness ($\mathcal{F}_{syn}$), an efficient metric that isolates true text-to-vision grounding by measuring the Shapley Interaction Index across continuous perturbation trajectories.
    \item We conduct a comprehensive benchmark over 3 benchmark datasets, 9 XAI methods and 3 VLM architectures. By isolating explainer fixed effects via an LMM, we reveal that attention-based explainers outperform newly proposed VLM-native explainers, which we empirically prove prioritize raw feature localization at the expense of cross-modal interaction.
\end{itemize}}

\section{Related Work}\label{related_work}
\subsection{Explainable AI for VLMs}
Post-hoc XAI methods for VLMs has evolved across three generations. \textbf{i) Classical feature attribution} methods (e.g., Input$\times$Gradient \cite{simonyan2013deep}, Integrated Gradients \cite{sundararajan2017axiomatic}) were designed for discriminative classification. When applied to deep, autoregressive transformers, they frequently suffer from gradient shattering and instability. \textbf{ii) Attention-based methods} (e.g., Attention Rollout \cite{abnar2020quantifying}, Grad$\times$Rollout \cite{chefer2021generic}, AttnLRP \cite{achtibat2024attnlrp}) leverage transformer mechanics to trace information flow. However, inherent noise from Vision Transformer (ViT) layers frequently bleeds into cross-attention modules, often yielding diffuse multimodal attributions. Recently, \textbf{iii) VLM-native explainers} (e.g., LLaVA-CAM, TAM) have emerged, utilizing targeted activation maps to explicitly ground text tokens to visual features. A central benchmarking objective is rigorously verifying whether these native methods truly capture cross-modal synergy better than their predecessors.

\subsection{Evaluation of Multimodal XAI}
Evaluating XAI generally splits into plausibility and faithfulness \cite{jacovi2020towards}. Plausibility metrics (e.g., pointing game, IoU \cite{zhang2018top, li2025token}) measure human alignment, but risk rewarding visually salient object detection over the model's actual reasoning. To audit true internal reasoning, evaluation must focus on \textit{faithfulness}, historically measured via feature perturbation (e.g., Insertion/Deletion AUC \cite{petsiuk2018rise}, SRG (Symmetric Relevance Gain) \cite{bluecherdecoupling}). However, standard perturbation metrics evaluate modalities in strict isolation \cite{agarwal2025rethinking}. Modern VLMs possess strong modality priors, frequently relying on textual co-occurrences \cite{li2023evaluating} to compensate for visual blind spots \cite{tong2024eyes}. Grounded in Partial Information Decomposition \cite{williams2010nonnegative}, such overlapping priors act as redundant features. Because unimodal metrics ignore cross-modal dependencies, they systematically penalize explainers that highlight redundant features, inadvertently rewarding shortcuts rather than verifying true cross-modal synergy.

\subsection{Shapley Interaction Index}
To rigorously isolate cross-modal feature interactions, researchers utilize cooperative game theory. While standard Shapley Values \cite{roth1988introduction} measure individual marginal contributions, the Shapley Interaction Index (SII) \cite{grabisch1999axiomatic} utilizes the Harsanyi dividend to measure pure synergy between features. Recent works have applied SII as an \textit{interpretability method} to extract attributions for early, low-resolution models like ViLT or CLIP \cite{wenderoth2025measuring, banieckiexplaining}. These studies utilize SII strictly as a post-hoc explainer, not as a benchmarking metric. Furthermore, exact SII computation requires $\mathcal{O}(2^{N})$ forward passes, making it intractable for the combinatorial token spaces of modern high-resolution VLMs. To overcome this computational bottleneck, we propose a scalable metric that approximates SII, enabling large-scale benchmarking of XAI methods for VLMs.

% \section{Limitations of unimodal faithfulness}\label{limitations}
% \input{sections/limitations}

\section{Method}\label{method}
\subsection{Notation}
\label{sec:notation}

Let $f: \mathcal{V} \times \mathcal{T} \to [0, 1]$ denote a VLM that maps an image $I = \{p_1, \ldots, p_m\} \in \mathcal{V}$ (comprising $m$ visual patches) and a text prompt $T = \{t_1, \ldots, t_n\} \in \mathcal{T}$ (comprising $n$ tokens) to a confidence score for a target prediction. We define a \textit{zero-state} baseline for both modalities, denoted as $I_\emptyset$ (an aggressively blurred or zero-padded image) and $T_\emptyset$ (masked text tokens). Let $\mathcal{E}$ be a post-hoc explainer that outputs an attribution map scoring the importance of each visual patch and text token. Using the rankings derived from $\mathcal{E}$, we define $I_k \subseteq I$ and $T_k \subseteq T$ (for any continuous threshold $k \in [0, 1]$) as the subsets consisting of the top $k$-proportion of the most highly attributed visual patches and textual tokens, respectively.

We formally distinguish between unimodal and multimodal faithfulness evaluation metrics. We define a unimodal faithfulness evaluation metric as a function mapping a subset of single-modality features to a scalar faithfulness score. Using $\mathcal{P}(\cdot)$ to denote the power set, a visual metric $\mu_{I}(\mathcal{E},  f): \mathcal{P}(I) \rightarrow \mathbb{R}$ evaluates the visual attribution $I_k$ in strict isolation, verifying its faithfulness while treating the text as a frozen, static context. While this metric strictly depends on the evaluated explainer $\mathcal{E}$ and the specific VLM $f$, we denote it simply as $\mu_I$ for notational brevity. Specifically, the standard unimodal faithfulness metrics, Deletion AUC, Insertion AUC and SRG are defined for the visual modality as:

\begin{align}
    \mu_{I}^{del} &= \int_{0}^{1} f(I \setminus I_k, T)\, dk &
    \mu_{I}^{ins} &= \int_{0}^{1} f(I_\emptyset \cup I_k, T)\, dk  &
    \mu_{I}^{srg} = \mu_{I}^{ins} - \mu_{I}^{del}
\end{align}
Accordingly, perfect explainer minimizes confidence upon feature removal and maximizes it upon introduction ($\mu_{I}^{del} \to 0, \mu_{I}^{ins} \to 1$), whereas the worst explainer yields the exact inverse ($\mu_{I}^{del} \to 1, \mu_{I}^{ins} \to 0$). 
For text, one can define a similar metric $\mu_{T}: \mathcal{P}(T) \rightarrow \mathbb{R}$ evaluates the textual attribution $T_k$ against a static image context. By symmetry, the corresponding textual metrics, $\mu_{T}^{del}$, $\mu_{T}^{ins}$ and $\mu_{T}^{srg}$ evaluate the removal and addition of $T_k$ against an unperturbed image $I$.

By contrast, a true multimodal faithfulness metric must evaluate the two modalities together. We formally define it as a joint mapping $\mu_{I \times T}: \mathcal{P}(I) \times \mathcal{P}(T) \rightarrow \mathbb{R}$ that simultaneously maps subsets from both modality spaces to a unified scalar. Rather than performing independent ablations, $\mu_{I \times T}$ evaluates the joint perturbation to explicitly isolate the synergistic interaction between $I_k$ and $T_k$, measuring how the modalities fuse to produce a prediction.

\subsection{Limitations of unimodal faithfulness metrics}
\label{sec:limitations}
To demonstrate why unimodal faithfulness metrics are vulnerable when applied to highly entangled VLMs, we examine their behavior under boundary conditions. Consider a highly capable VLM $f$ with perfect confidence (s.t. $f(I,T) = 1.0$) operating under perfect cross-modal redundancy (e.g., the model can answer any query using any modality without the other). Under perturbation of a single modality, the remaining modality is entirely sufficient to preserve model confidence. This yields the following boundary behavior across all perturbation thresholds $k$: $f(I \setminus I_k, T) = 1.0$ and $f(I_\emptyset \cup I_k, T) = 1.0$. If we evaluate a theoretically perfect visual explainer (one that correctly highlights the exact ground-truth features) using standard unimodal metrics, we get:

\begin{align}
    \mu_{I}^{del} &= \int_{0}^{1}  f(I \setminus I_k, T) \, dk  = 1.0 &
    \mu_{I}^{ins} &= \int_{0}^{1}  f(I_\emptyset \cup I_k, T)\, dk = 1.0
\end{align}

This results in an evaluation paradox: the deletion metric assigns the perfect explainer the \textit{worst} possible score ($\mu_I^{del}=1.0$), failing to recognize the importance of the visual features because the model's confidence is sustained by the redundant text. Conversely, the visual insertion metric evaluates features by adding them to a masked image while keeping the text unperturbed. Because the redundant text alone already maximizes the model's confidence at step zero ($f(I_\emptyset, T) = 1.0$), the insertion curve remains entirely flat at $1.0$. Consequently, the metric assigns a \textit{perfect} insertion score to \textit{any} explainer, even a completely random one, rendering it powerless to distinguish genuine attribution from noise. Due to the bidirectional symmetry of this VLM, this failure applies equally to the textual evaluation metrics ($\mu_{T}^{del}$ and $\mu_{T}^{ins}$).

This paradox has further practical implications. Real-world multimodal datasets are notoriously heterogeneous. Recent studies demonstrate that evaluation benchmarks frequently suffer from severe modality biases, where a significant portion of questions can be answered using only textual co-occurrences or language priors \cite{chen2024we, li2024naturalbench, zheng2025mllms, chen2024quantifying, lee2025vlind, liu2024insight, li2023evaluating}. Thus, unimodal metrics cannot distinguish between a poor explainer and a highly faithful explainer operating on a redundant instance. This inability to isolate pure synergistic interaction makes unimodal evaluation unstable for modern VLMs, yielding inconsistent rankings as empirically demonstrated later in our experiments (Section \ref{sec:unimodal_rank_results}).

\subsection{Synergistic Faithfulness Metric}

To rigorously quantify the true cross-modal interactions between visual and textual features, rather than evaluating them as isolated unimodal signals, we must model their joint contributions. From a game-theoretic perspective, VLM inference can be formalized as a cooperative game. In this formulation, the \textit{game} is the model's prediction task (e.g., maximizing the model's confidence score for a target generation). The \textit{players} collaborating in this game are the individual input features (e.g., patches and tokens). Given a coalition of features $S$, an input image $I$ comprising $m$ visual patches ($|I| = m$) and a text prompt $T$ comprising $n$ tokens ($|T| = n$), the complete set of players is the multimodal feature set $\mathcal{M} = I \cup T$. We define a characteristic value function $\nu: \mathcal{P}({\mathcal{M}}) \to [0, 1]$, where $\nu(S)$ yields the VLM's confidence score $f$ when only the players in coalition $S \subseteq \mathcal{M}$ are \textit{unmasked} (retaining their original input values), and all other players $\mathcal{M} \setminus S$ are masked to the zero-state.

While standard Shapley Values \cite{roth1988introduction} evaluate the marginal contribution of a single player, evaluating cross-modal grounding requires measuring the interaction between a specific visual patch $p_i \in I$ and a text token $t_j \in T$. The gold-standard metric for this is the Shapley Interaction Index (SII) \cite{grabisch1999axiomatic}:

\begin{equation}
    \Phi_{ij}(\nu) = \sum_{S \subseteq \mathcal{M} \setminus \{p_i, t_j\}} \frac{|S|!(|\mathcal{M}|-|S|-2)!}{(|\mathcal{M}|-1)!} \Delta_{i,j}\nu(S)
\end{equation}

where $S$ acts as a variable context coalition of background features, and $\Delta_{i,j}\nu(S)$ represents the 2-player Harsanyi dividend:
\begin{equation}
    \Delta_{i,j}\nu(S) = \nu(S \cup \{p_i, t_j\}) - \nu(S \cup \{p_i\}) - \nu(S \cup \{t_j\}) + \nu(S)
\end{equation}

Standard unimodal perturbation fails because it strictly evaluates modalities in isolations, structurally ignoring the joint Harsanyi dividend ($\Delta_{i,j}$) entirely. However, exactly computing the SII to capture this dividend requires $\mathcal{O}(2^{m+n})$ forward passes, which is intractable for high-resolution VLMs. 

To resolve this, we approximate the interaction by evaluating the Harsanyi dividend dynamically along a continuous perturbation trajectory. For any threshold step $k \in [0, 1]$, we compute the synergistic interaction in both directions. First, we evaluate the top of the boolean lattice (\textit{synergistic deletion}) by progressively masking the top $k$-proportion of features from the full state, calculating the joint degradation alongside the independent marginal degradations. Second, we evaluate the bottom of the lattice (\textit{synergistic insertion}) by iteratively revealing the top $k$-proportion of features from the zero-state:

\begin{align}
    del_{joint}(k) &= f(I \setminus I_k, T \setminus T_k) & ins_{joint}(k) &= f(I_\emptyset \cup I_k, T_\emptyset \cup T_k) \\
    del_{img}(k) &= f(I \setminus I_k, T) & ins_{img}(k) &= f(I_\emptyset \cup I_k, T_\emptyset) \\
    del_{txt}(k) &= f(I, T \setminus T_k) & ins_{txt}(k) &= f(I_\emptyset, T_\emptyset \cup T_k)
\end{align}
 
By applying the Harsanyi structure to these perturbation steps, the synergistic deletion and insertion bounds reflect the pure interaction of the manipulated features:
\begin{align}
    syn_{del}(k) &= del_{joint}(k) - del_{img}(k) - del_{txt}(k) + f(I, T) \\
    syn_{ins}(k) &= ins_{joint}(k) - ins_{img}(k) - ins_{txt}(k) + f(I_\emptyset, T_\emptyset)
\end{align}

To evaluate an explainer across the entire feature distribution, we integrate the synergy scores across the perturbation threshold $k$ (approximated via a Riemann sum over $K$ discrete steps):
\begin{align}
    AUC_{ins} &= \int_{0}^{1} syn_{ins}(k) \, dk &
    AUC_{del} &= \int_{0}^{1} syn_{del}(k) \, dk
\end{align}

To unify these complementary integrals into a single scalar, we define the final \textbf{Synergistic Faithfulness metric}:
\begin{equation}
    \mathcal{F}_{syn} = \frac{AUC_{ins} + AUC_{del}}{2}
\end{equation}

Thus, $\mathcal{F}_{syn}$ serves as a concrete instantiation of our multimodal metric formulation $\mu_{I \times T}: \mathcal{P}(I) \times \mathcal{P}(T) \rightarrow \mathbb{R}$. This formulation guarantees that an explainer achieves a high score if and only if it identifies feature subsets that are strictly synergistic across modalities, mitigating the vulnerabilities induced by cross-modal redundancy while operating in computationally feasible $\mathcal{O}(K)$ time.

\section{Experiments}\label{experiments}
To empirically validate our claims, we design a benchmarking framework with three primary objectives: (1) to demonstrate the fallacious signals provided by traditional unimodal faithfulness evaluation metrics, (2) to show that our proposed synergistic faithfulness metric accurately and efficiently approximates exact SII computation, and (3) to deploy $\mathcal{F}_{syn}$ at scale to rigorously benchmark the true cross-modal reasoning of multimodal XAI methods.

\subsection{Experimental Setup} \label{sec:benchmark_setup}

\textbf{Explainers.} We evaluate 9 distinct post-hoc XAI methods across three categories: i) \textbf{Classical:} Random attribution (baseline), Integrated Gradients \cite{sundararajan2017axiomatic}, Input$\times$Gradients \cite{simonyan2013deep}, and GradCAM \cite{selvaraju2017grad}; ii) \textbf{Attention-based:} Rollout \cite{abnar2020quantifying}, Grad$\times$Rollout \cite{chefer2021generic}, and AttnLRP \cite{achtibat2024attnlrp}; iii) \textbf{VLM-native:} LLaVA-CAM \cite{zhang2025redundancy} and TAM \cite{li2025token}. We exclude explainers restricted strictly to visual attribution (e.g., IGOS \cite{xing2025large}, Eagle \cite{chen2025mllms}), as our metric requires corresponding textual attributions for joint perturbation. Furthermore, we omit early game-theoretic explainers (e.g., MultiSHAP \cite{wang2025multishap}, FIxLIP \cite{banieckiexplaining}) designed for shallow dual-encoders, as they hit intractable $\mathcal{O}(2^n)$ complexity on modern autoregressive sequence lengths.

\textbf{Models.} To ensure architecture-agnostic findings, we evaluate three prominent VLM architectures: LLaVA-1.5-7B \cite{liu2024improved}, Qwen2.5-VL-3B \cite{bai2025qwen3}, and InternVL-3.5-2B \cite{wang2025internvl3}. Targeting the 2B-7B parameter regime aligns with established evaluation standards \cite{li2025token, chen2025mllms}. This scale strikes a balance: it exhibits the emergent reasoning and heavy language priors necessary to trigger the redundancy paradox, while remaining efficient enough to make massive perturbation sweeps computationally tractable.

\textbf{Benchmark Datasets.} We construct our benchmark across three datasets specifically chosen because they are curated to suppress unimodal language priors: RePOPE \cite{neuhaus2025repope}, CVBench \cite{tong2024cambrian}, and MMStar \cite{chen2024we}. Conversely, we exclude standard captioning datasets (e.g., MSCOCO \cite{lin2014microsoft,chen2015microsoft}). Because measuring Shapley synergy mathematically requires dynamic interplay between specific semantic query tokens and visual patches, utilizing fixed prompts (e.g., "Describe this image") causes textual attribution to become trivial, rendering cross-modal evaluation fundamentally uninformative.

\subsection{Ranking instability of unimodal metrics}\label{sec:unimodal_rank_results}

\begin{figure}[t]
  \centering
  \includegraphics[width=0.7\linewidth]{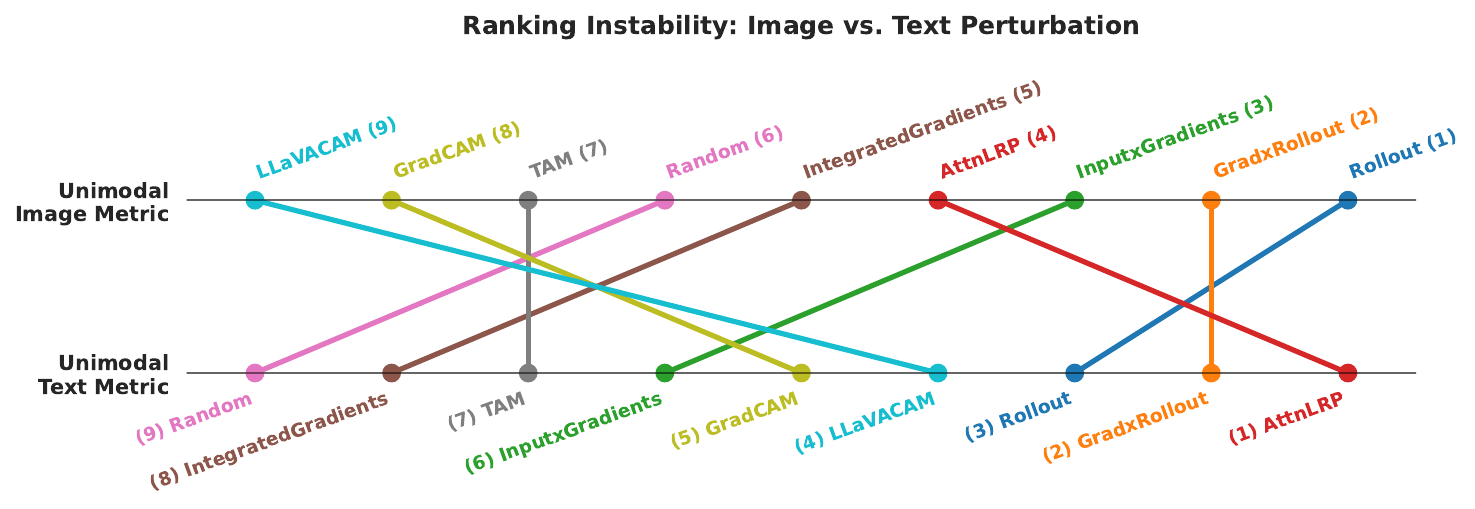}
  \caption{Ranking instability of explainers when evaluated using isolated unimodal perturbation. The crossing lines demonstrate the lack of agreement between visual and textual faithfulness metrics.}
  \label{fig:rank_instability}
\end{figure}

To empirically validate the vulnerability of unimodal faithfulness metrics established in Section \ref{sec:limitations}, we compute Kendall's rank correlation ($\tau$) between the explainer rankings obtained after using the visual faithfulness metric $\mu_I^{srg}$ and those obtained using the corresponding textual metric $\mu_T^{srg}$. If these metrics captured reasoning on non-redundant benchmarks, rankings should strongly correlate across modalities. Instead, we observe a severe evaluation collapse. Globally, the rank correlation is indistinguishable from random noise ($\tau = 0.167$, $p = 0.61$). This inter-modality contradiction (Fig. \ref{fig:rank_instability}) confirms that an explainer deemed faithful by visual perturbation is frequently ranked as unfaithful by textual perturbation.

Furthermore, the degree of this ranking instability is sensitive to the dataset's cross-modal redundancy. On RePOPE, where the questions to detect object hallucination allow models to fall back on residual unimodal biases, ranking agreement remains weak ($\tau = 0.22$, $p = 0.48$). Even on MMStar, which demand strict joint reasoning, the metrics exhibit a moderate but marginal correlation ($\tau = 0.50$, $p = 0.08$). This task-dependent fluctuation demonstrates that unimodal metrics primarily fail to provide a stable faithfulness evaluation. Extensive rank shift visualizations are in Appendix \ref{appendix_benchmark}.

\takeaway{The ranking inconsistencies across modalities demonstrate the weakness of unimodal faithfulness metrics and the need for synergistic multimodal faithfulness metrics.}

\subsection{Validation of the Synergistic Faithfulness Metric}

Before deploying the synergistic faithfulness metric ($\mathcal{F}_{syn}$) at scale, we validate its accuracy and efficiency across three settings. First, we use a controlled ground-truth environment to confirm it correctly isolates pure cross-modal synergy. Second, we prove $\mathcal{F}_{syn}$ strongly correlates with the Shapley Interaction Index (SII) on high-resolution VLMs by comparing it against a macro-coalition formulation of SII. Finally, we compute the true SII without macro-coalitions in an exact atomic setting, proving our metric fully preserves mathematical fidelity while eliminating severe computational bottlenecks.

\subsubsection{Controlled Ground-Truth Validation}

Validating empirically $\mathcal{F}_{syn}$ requires an environment where true cross-modal dependencies are known by construction. Standard post-hoc evaluations on deep Vision-Language Models (VLMs) lack verifiable ground-truth explanations, making it difficult to definitively prove whether a metric strictly rewards cross-modal faithfulness.

To establish an independent validity check, we construct a controlled evaluation environment utilizing the GridVQA dataset \cite{belsare2026gridvqa} and a custom white-box synthetic VLM. We utilize the $D_{pure}$ subset of GridVQA, which is explicitly designed to eliminate unimodal shortcut learning and force joint cross-modal reasoning, thereby providing mathematically verifiable ground-truth feature masks. To guarantee that the evaluated model relies exclusively on these correct reasoning pathways, we implement a synthetic VLM governed strictly by the ground-truth tokens. During the perturbation phase of the metric, this model calculates a textual score ($T_{score}$) and a visual score ($V_{score}$) based purely on the rate of unmasked ground-truth features. The model then outputs a joint probability score defined as $T_{score} \times V_{score}$. Under this architecture, if an explainer masks the true features, cross-modal synergy mathematically breaks down, forcing the model to output a low target logit.

We evaluate three synthetic explainers within this controlled environment to observe the behavior of $\mathcal{F}_{syn}$: (i) \textit{Oracle:} An explainer that perfectly targets the ground-truth multimodal features; (ii) \textit{Random:} An explainer that uniformly masks features across both modalities; and (ii) \textit{Anti-Oracle:} An explainer that actively avoids ground-truth features, targeting only background information. A theoretically valid multimodal metric must rank these baselines strictly as Oracle $>$ Random $>$ Anti-Oracle. Our empirical evaluation confirms this expected behavior: the Oracle explainer achieves a Synergistic Faithfulness score of $\mathcal{F}_{syn} = 0.82$, the Random explainer scores $0.35$, and the Anti-Oracle establishes the lower bound at $0.07$. These results provide independent empirical proof that $\mathcal{F}_{syn}$ successfully identifies and strictly rewards true causal localization. Having established this baseline validity, we now apply $\mathcal{F}_{syn}$ to evaluate post-hoc explainers on complex, real-world deep VLMs.

\takeaway{By defining a verifiable environment without modality biases using GridVQA dataset, we prove that $\mathcal{F}_{syn}$ identifies and rewards true cross-modal synergy, ranking a perfect Oracle explainer significantly above Random and Anti-Oracle baselines.}

% \subsubsection{Validation of the Synergistic Faithfulness Metric}
\subsubsection{Correlation with Macro-Coalition SII}
\label{sec:macro_correlations}

\begin{figure}[t]
  \centering
  \begin{minipage}[t]{0.48\linewidth}
    \centering
    \includegraphics[height=50mm]{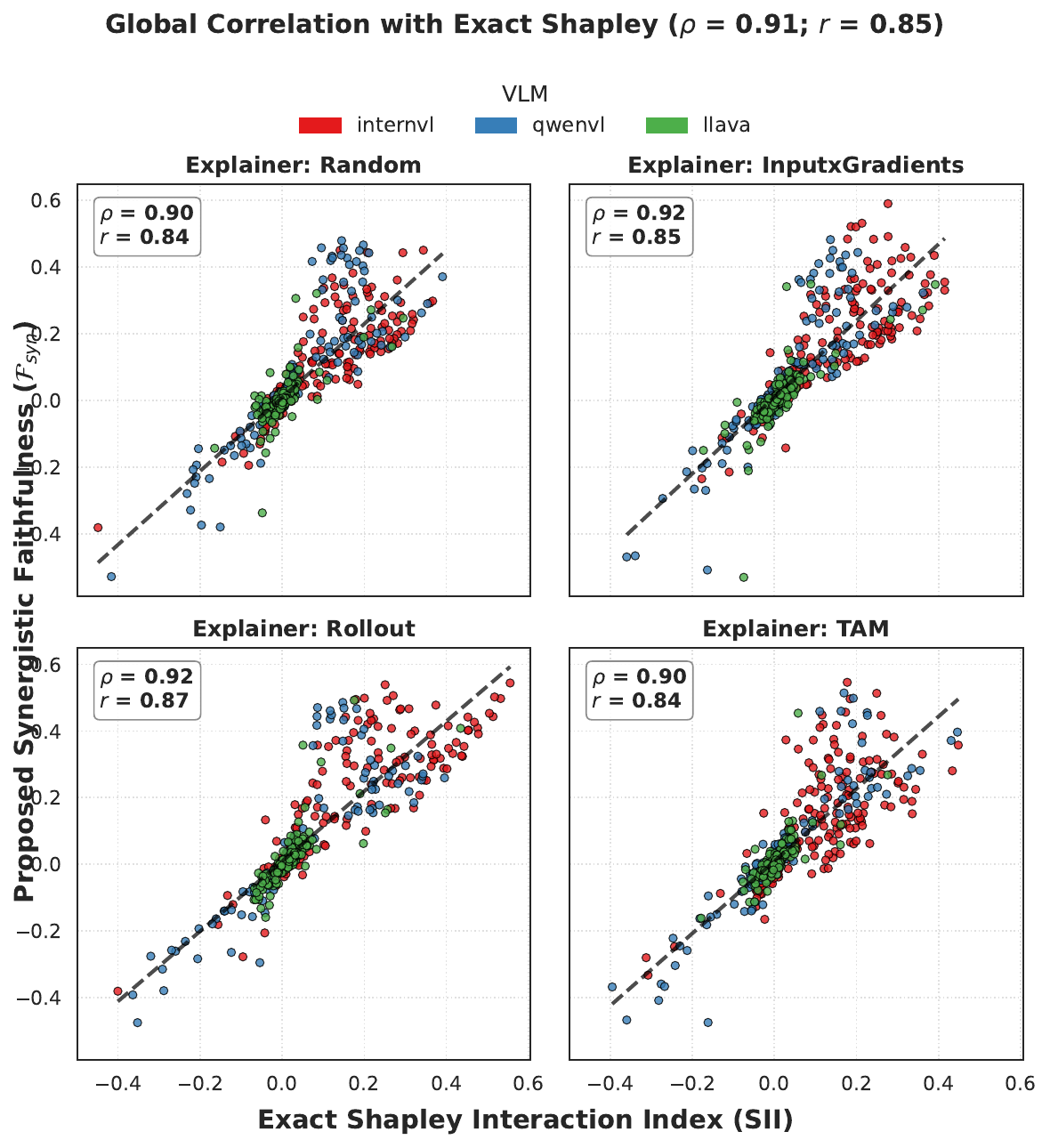}
    \caption{Correlation between the ground-truth SII baseline and the synergistic faithfulness ($\mathcal{F}_{syn}$). The strong rank alignment is consistently preserved across fundamentally different explainer types and VLM architectures.}
    \label{fig:correlation_subplots}
  \end{minipage}
  \hfill
  \begin{minipage}[t]{0.48\linewidth}
    \centering
    \includegraphics[height=4cm]{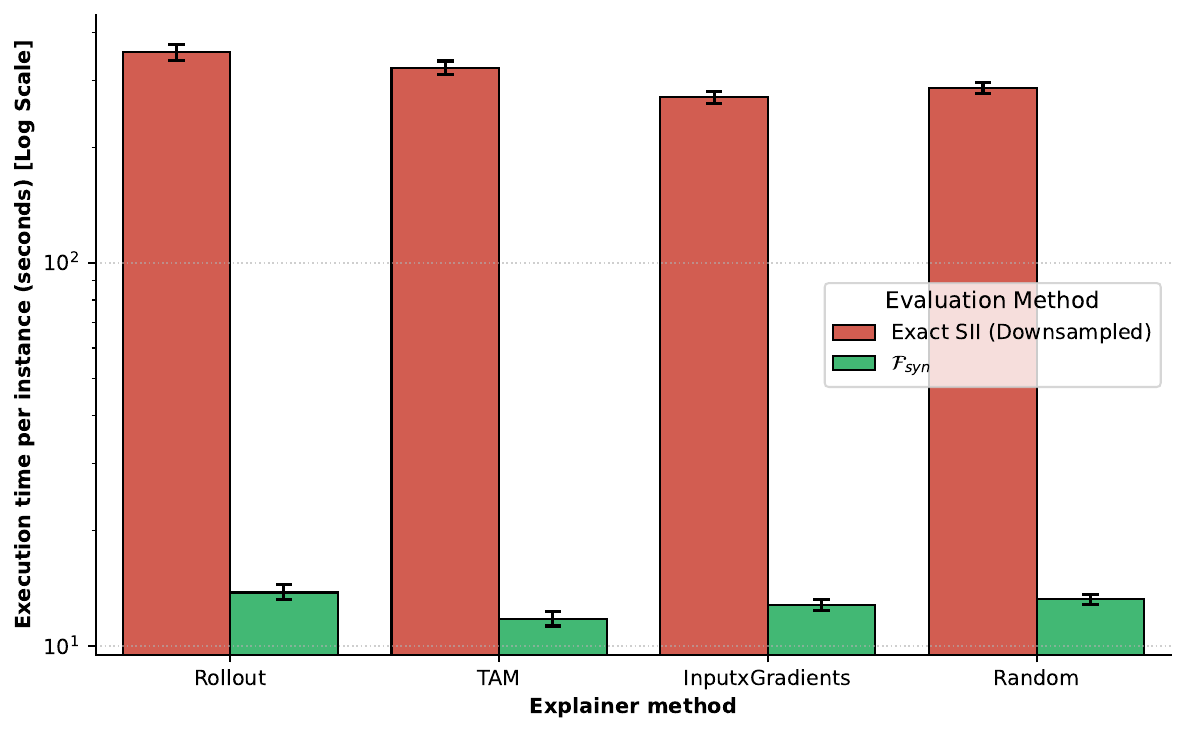}
    \caption{Empirical execution time per instance (log scale). $\mathcal{F}_{syn}$ achieves a $24\times$ speedup over the exact Shapley approximator, reducing average compute time from 281s to 11s without compromising ordinal integrity.}
    \label{fig:runtime_comparison}
  \end{minipage}
\end{figure}

We first evaluate whether $\mathcal{F}_{syn}$ captures Shapley interactions across standard, high-resolution VLMs. Because exact SII is mathematically intractable at these resolutions, we utilize a macro-coalition SII baseline. For any perturbation step $k$, we aggregate the targeted features into primary foreground players and cluster the remaining background into macro-coalitions (detailed in Appendix \ref{app:macro_coalitions}). 

Computed across $N=200$ multimodal instances, $\mathcal{F}_{syn}$ exhibits a strong rank alignment with this baseline. As visualized in Fig. \ref{fig:correlation_subplots}, this correlation is perfectly preserved across fundamentally different explainer architectures: Random ($\rho=0.90$, $r=0.84$), Input$\times$Gradients ($\rho=0.92$, $r=0.85$), Rollout ($\rho=0.92$, $r=0.87$), and TAM ($\rho=0.90$, $r=0.84$). Crucially, $\mathcal{F}_{syn}$ achieves this efficiently. Because calculating the joint, visual, and textual bounds at step $k$ requires only 6 forward passes, the entire metric requires exactly $6K + 2$ forward passes. As demonstrated in  Fig. \ref{fig:runtime_comparison}, the macro-coalition SII required 281 seconds per image, whereas $\mathcal{F}_{syn}$ completed in 11 seconds (a \textbf{24$\times$ speedup}).

\takeaway{$\mathcal{F}_{syn}$ maintains a strong rank alignment ($\rho \ge 0.90$) with the macro-coalition SII baseline across diverse explainers while delivering a $24\times$ computational speedup, reducing per-image execution time from 281 to 11 seconds.}

\subsubsection{Correlation with Exact Atomic SII}
\label{sec:exact_correlations}

While macro-coalitions make standard resolutions tractable, they still rely on spatial approximation. To definitively prove that $\mathcal{F}_{syn}$ preserves the true atomic feature-ranking fidelity without being confounded by macroscopic grouping, we conduct a constrained full-enumeration experiment using QwenVL-2.5, which permits dynamic visual token adjustment, we downsample images to $32 \times 32$ pixels. This bounds the game-theoretic state space to $N \le 17$ atomic players, removing the need for macro-coalitions. Even under these constraints, computing the exact atomic SII required $2^{17} \approx 130,000$ forward passes per image. Evaluated across the explainers, $\mathcal{F}_{syn}$ maintains a near-perfect rank alignment with the exact atomic index (Spearman $\rho = 0.93$, Pearson $r = 0.91$, $p < 0.001$). Furthermore, while the exact SII required an average of 637.43 seconds per image, $\mathcal{F}_{syn}$ completed the evaluation in just 7.18 seconds, achieving an \textbf{88.7$\times$ speedup}. We provide further details and results in Appendix \ref{app:exact_sii_validation}.

\takeaway{By preserving the ordinal integrity of exact Shapley interaction on both atomic and architectural scales while accelerating computation by up to $88\times$, $\mathcal{F}_{syn}$ successfully unlocks the ability to evaluate multimodal explainers.}

\subsection{Explainers synergistic faithfulness benchmark results}
\label{sec:benchmark_results}

\begin{figure}[t]
  \begin{minipage}[b]{0.48\linewidth}
    \centering
    \includegraphics[width=\linewidth]{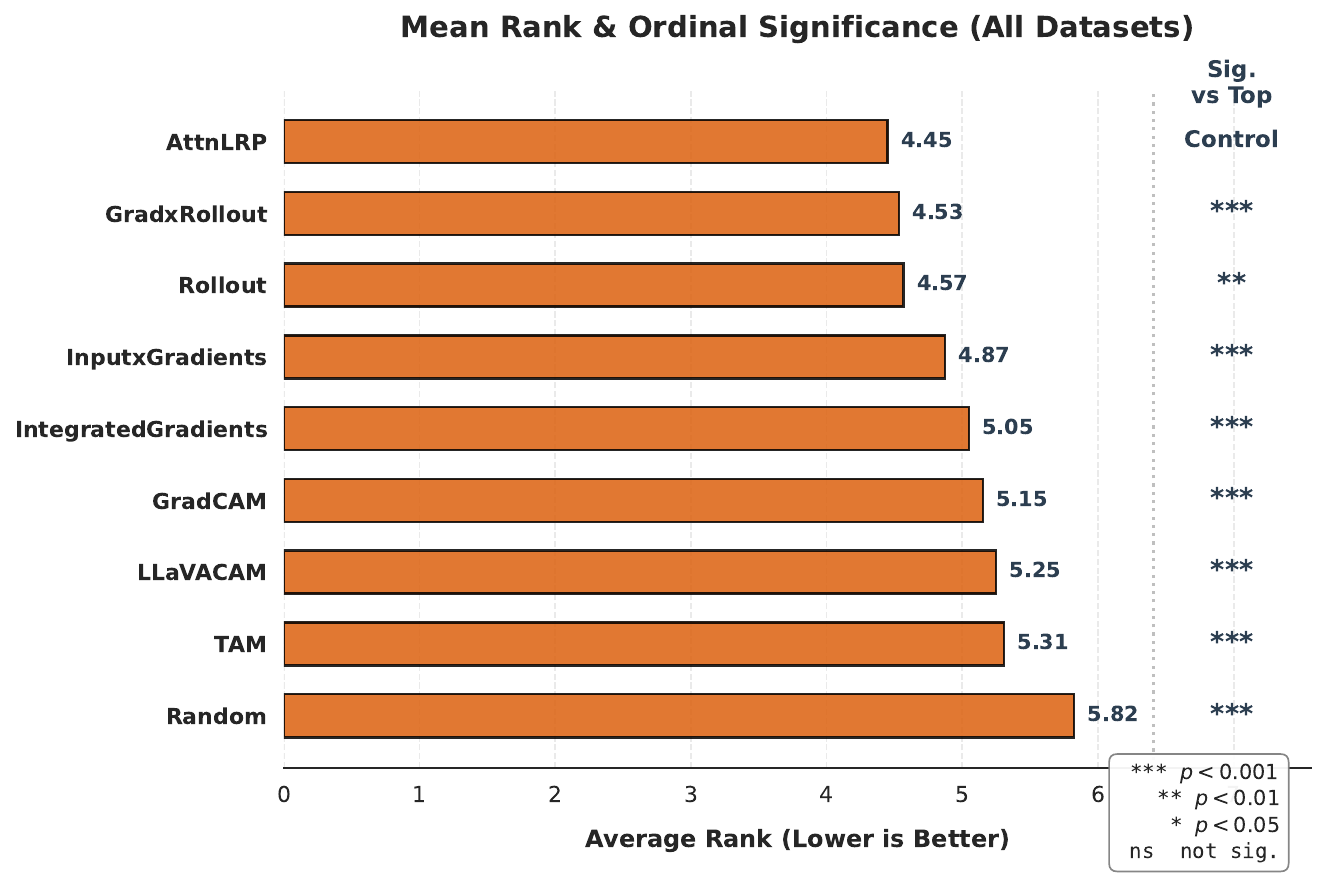}
    \caption{Average explainer rankings across all benchmark instances. Significance indicators denote the results of pairwise Wilcoxon signed-rank tests evaluated against the top-performing method.}
    \label{fig:wilcoxon_ranks}
  \end{minipage}
  \hfill
  \centering
  \begin{minipage}[b]{0.48\linewidth}
    \centering
    \includegraphics[width=\linewidth,trim=0 0 0 2.2em,clip]{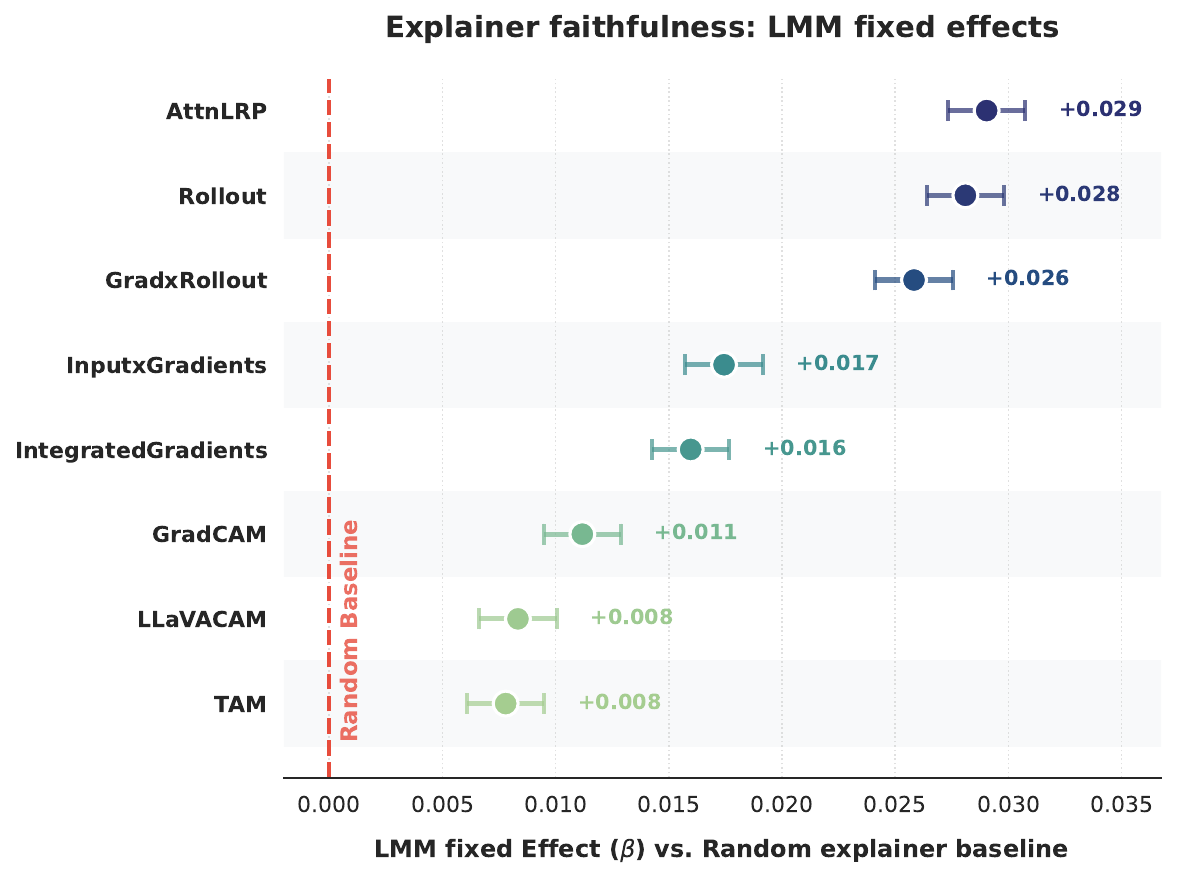}
    \caption{Forest plot of LMM fixed effects ($\beta$), estimating the algorithmic contribution of each explainer while controlling for dataset and model variance. The Random explainer serves as the reference baseline ($\beta=0$).}
    \label{fig:forest_plot}
  \end{minipage}
  
\end{figure}

\begin{table*}[t]
\centering
\caption{Benchmark Leaderboard. Scores represent average $\mathcal{F}_{syn}$ across models. The final column reports the LMM fixed effect ($\beta$), representing the true algorithmic superiority over the Random baseline, controlling for dataset and model variance. \textbf{Bold} indicates the best results and underline the second best.}
\label{tab:main_leaderboard}
% \resizebox{\textwidth}{!}{%
\begin{tabular}{l|ccc|c|c}
\toprule
\textbf{Explainer} & \textbf{CVBench}  & \textbf{MMStar} & \textbf{RePOPE} & \textbf{Overall} & \textbf{LMM $\beta$} \\
%&  &  &  & & \textbf{(Rank)} \\
\midrule
Random & 0.069$_{(\pm0.08)}$ & 0.096$_{(\pm0.11)}$ & 0.068$_{(\pm0.11)}$ & 0.073$_{(\pm0.10)}$ & 0.000 \\
\midrule
% \textit{Classical} & & & & \\
\addlinespace
GradCAM & 0.075$_{(\pm0.08)}$ & 0.103$_{(\pm0.11)}$ & 0.083$_{(\pm0.12)}$ & 0.084$_{(\pm0.11)}$ & 0.011 \\
Input$\times$Gradients & 0.092$_{(\pm0.10)}$ & 0.112$_{(\pm0.13)}$ & 0.083$_{(\pm0.12)}$ & 0.090$_{(\pm0.12)}$ & 0.017 \\
Integ. Gradients & 0.084$_{(\pm0.09)}$ & 0.109$_{(\pm0.12)}$ & 0.085$_{(\pm0.13)}$ & 0.089$_{(\pm0.12)}$  & 0.016 \\
\midrule
% \textit{Attention-based} & & & & \\
\addlinespace
Rollout & \underline{0.103$_{(\pm0.13)}$} & \underline{0.133$_{(\pm0.15)}$} & \underline{0.090$_{(\pm0.14)}$} & \underline{0.101$_{(\pm0.14)}$} & \underline{0.028} \\
Grad$\times$Rollout & \bf 0.110$_{(\pm0.13)}$ & \bf 0.137$_{(\pm0.16)}$ & 0.081$_{(\pm0.12)}$ & 0.099$_{(\pm0.13)}$ & 0.026 \\
AttnLRP & 0.099$_{(\pm0.11)}$ & 0.128$_{(\pm0.14)}$ & \bf 0.095$_{(\pm0.14)}$ & \bf 0.102$_{(\pm0.13)}$ & \bf 0.029 \\
\midrule
% \textit{VLM-Native} & & & & \\
\addlinespace
LLaVA-CAM & 0.074$_{(\pm0.08)}$ & 0.102$_{(\pm0.12)}$ & 0.079$_{(\pm0.11)}$ & 0.081$_{(\pm0.11)}$  & 0.008 \\
TAM & 0.082$_{(\pm0.09)}$ & 0.114$_{(\pm0.13)}$ & 0.070$_{(\pm0.11)}$ & 0.081$_{(\pm0.11)}$ & 0.008 \\
\bottomrule
\end{tabular}%
%}
\end{table*}

To evaluate the cross-modal faithfulness of the evaluated XAI methods, we applied the $\mathcal{F}_{syn}$ metric across the benchmark instances. The unadjusted average scores across different VLM architectures and datasets are detailed in Table \ref{tab:main_leaderboard}. To rigorously assess the ordinal consistency of explainer performance, we computed the mean rank of each method across all individual test instances. As visualized in Fig. \ref{fig:wilcoxon_ranks}, we conducted pairwise Wilcoxon signed-rank tests (with Bonferroni correction) to compare all methods against the top-performing explainer. This non-parametric analysis demonstrates that attention-based methods reach better ranking on average, establishing a statistically significant ordinal superiority over both classical and VLM-native approaches. 

However, direct averaging of scores and ranks across datasets with varying levels of redundancy may still obscure underlying contextual variance. To isolate the faithfulness of each explainer from this variance, we fitted a Linear Mixed-effects Model (LMM). By treating the \textit{Explainer} as a fixed effect, while modeling the VLM architecture, the dataset, and the specific multimodal instance as random effects, the LMM statistically sanitizes the performance scores from dataset difficulty and model capability. The full mathematical formulation of this model is provided in Appendix \ref{app:lmm_formulation}.

The resulting LMM fixed effects ($\beta$), visualized in Fig. \ref{fig:forest_plot}, indicate that attention-based XAI methods consistently outperform both classical and VLM-native approaches. Specifically, AttnLRP and Rollout achieved the highest statistically significant coefficients ($\beta = +0.029$ and $\beta = +0.028$ respectively, $p < 0.001$) relative to the Random baseline. Notably, VLM-native methods (TAM and LLaVA-CAM), despite being specifically designed for multimodal architectures, yielded marginal improvements over the random baseline ($\beta = +0.008$). While prior studies indicate these native methods frequently generate plausible visual bounding boxes, the LMM analysis demonstrates that they do not robustly capture cross-modal synergy when evaluated via Harsanyi interaction. This suggests that current VLM-native gradient routing mechanisms disproportionately prioritize visual salience over the cross-attention dynamics required for effective text-to-vision grounding. The comprehensive statistical report, including standard errors, $z$-scores, and exact $p$-values decomposed per dataset, is provided in Appendix \ref{appendix_statistical}. Further comprehensive benchmark results are provided in Appendix \ref{appendix_benchmark}, including qualitative results across all the explainers in Appendix \ref{appendix_quality}.

\takeaway{Our benchmark reveals that attention-based XAI methods are the most faithful algorithms for multimodal post-hoc explainability.}

% The quantitative LMM findings are strongly supported by qualitative visual inspection (Fig. \ref{fig:qualitative_heatmaps}). When visualizing the attributions for highly synergistic tasks—such as spatial reasoning in CVBench or interleaved logic in MMStar—LLM-adapted methods like AttnLRP successfully highlight both the critical visual patches and the specific query tokens driving the logic. In contrast, VLM-native methods like TAM frequently exhibit the "Clever Hans" effect. They act as sophisticated unimodal edge-detectors, producing visually appealing heatmaps on prominent objects in the image, even if those objects are entirely irrelevant to the specific textual query.

\section{Discussion \& Limitations}\label{discussion}

Our benchmark reveals a counterintuitive trend: specific VLM-native explainers (e.g., TAM) consistently underperform compared to attention-based explainers (e.g., AttnLRP, Rollout). We hypothesize this occurs because these native methods often over-index on visual salience, effectively functioning as unimodal object detectors rather than capturing joint cross-modal reasoning. To empirically evaluate this, we contrasted the spatial plausibility of these explainers on the COCO dataset against their cross-modal faithfulness on MMStar (detailed in Appendix \ref{app:coco_plausibility}). The results confirm that while methods like TAM excel at spatial localization (optimizing for human plausibility), their heatmaps are frequently unfaithful to the model's actual joint reasoning mechanism.

While $\mathcal{F}_{syn}$ successfully resolves the vulnerabilities of unimodal faithfulness metrics, our benchmark introduces specific limitations: \textbf{(1)} Approximating the continuous perturbation trajectory via a Riemann sum over $K$ discrete steps reduces $\mathcal{O}(2^{m+n})$ exact Shapley complexity to $\mathcal{O}(K)$ forward passes. While highly effective for offline benchmarking, $\mathcal{F}_{syn}$ remains more costly than traditional metrics and may not be suitable for hard real-time scenarios requiring fast response; \textbf{(2)} Our empirical validation is restricted to Visual Question Answering (VQA) formats using open-weight VLMs. Open-ended generation, broader modalities (audio/video), and -- more generally -- proprietary frontier models could not be considered as their black-box API (e.g. not revealing logits) prevents the direct use of explainers. We mitigate this by the diversity of VLMs we consider and the use of model-independent statistical tests (see LMM in Appendix \ref{app:lmm_formulation}). While our primary benchmark focuses on mid-scale VLMs (2B-7B), we successfully verified that our findings scale to higher-capacity models (up to 27B parameters) in Appendix \ref{app:larger_vlms}.

\textbf{Broader Impacts.} Current faithfulness evaluation metrics risk encouraging automation bias in high-stakes domains by rewarding plausible but mathematically unfaithful explanations. $\mathcal{F}_{syn}$ mitigates this vulnerability, equipping auditors to verify genuine cross-modal reasoning. Ultimately, the development of safe multimodal systems requires a shift away from isolated feature attribution toward XAI explicitly engineered to capture game-theoretic synergy.

\section{Conclusion }\label{conclusion}

Beyond the introduction of a scalable evaluation metric, we raise the necessity of shifting the definition of "faithful" explanation in multimodal settings. We have empirically demonstrated that standard unimodal metrics trigger an evaluation collapse. Notably, modality biases and redundancies in real-world multimodal datasets make traditional faithfulness metrics unable to distinguish faithful explainers from poor ones. To resolve this limitation, we introduced a novel and scalable game-theoretic metric that explicitly isolates the Harsanyi interaction dividend between modalities. By applying $\mathcal{F}_{syn}$ across a benchmark, we revealed a misalignment in modern XAI: visually localized attribution maps do not guarantee that a model engaged in joint reasoning. If we are to safely deploy VLMs, XAI evaluation must move beyond asking, "Did the explainer highlight the correct object?" and utilize metrics like $\mathcal{F}_{syn}$ to rigorously verify, "Did the model genuinely fuse these modalities to reach its prediction?"

\begin{ack}
This research was fonded in whole or in part by the Luxembourg National Research Fund (grants NCER22/IS/16570468/NCER-FT), by the Luxembourg National Research Fund (FNR), grants reference CORE C24/IS/18942843 and  C24/IS/18942843/FM2MRI and by BGL BNP Paribas Luxembourg. We thank Ulrick Ble, Anas Zilali and the BGL DataLab team for their collaboration and technical input throughout this work. The experiments  were performed on the Luxembourg national supercomputer MeluXina. The authors gratefully acknowledge the LuxProvide teams for their expert support. We also thank all reviewers for their precious comments.

\end{ack}

% \section*{References}

\bibliographystyle{unsrt}
%\bibliographystyle{abbrvnat}
%\setcitestyle{authoryear,open={((},close={))}}
\bibliography{bib} %%% Remove comment to use the external .bib file (using bibtex).

@article{simonyan2013deep,
  title={Deep inside convolutional networks: Visualising image classification models and saliency maps},
  author={Simonyan, Karen and Vedaldi, Andrea and Zisserman, Andrew},
  journal={arXiv preprint arXiv:1312.6034},
  year={2013}
}

@inproceedings{sundararajan2017axiomatic,
  title={Axiomatic attribution for deep networks},
  author={Sundararajan, Mukund and Taly, Ankur and Yan, Qiqi},
  booktitle={International conference on machine learning},
  pages={3319--3328},
  year={2017},
  organization={PMLR}
}

@inproceedings{selvaraju2017grad,
  title={Grad-cam: Visual explanations from deep networks via gradient-based localization},
  author={Selvaraju, Ramprasaath R and Cogswell, Michael and Das, Abhishek and Vedantam, Ramakrishna and Parikh, Devi and Batra, Dhruv},
  booktitle={Proceedings of the IEEE international conference on computer vision},
  pages={618--626},
  year={2017}
}

@inproceedings{abnar2020quantifying,
  title={Quantifying attention flow in transformers},
  author={Abnar, Samira and Zuidema, Willem},
  booktitle={Proceedings of the 58th annual meeting of the association for computational linguistics},
  pages={4190--4197},
  year={2020}
}

@inproceedings{chefer2021generic,
  title={Generic attention-model explainability for interpreting bi-modal and encoder-decoder transformers},
  author={Chefer, Hila and Gur, Shir and Wolf, Lior},
  booktitle={Proceedings of the IEEE/CVF international conference on computer vision},
  pages={397--406},
  year={2021}
}

@inproceedings{achtibat2024attnlrp,
  title={{{AttnLRP}}: Attention-Aware Layer-Wise Relevance Propagation for Transformers},
  author={Achtibat, Reduan and Hatefi, Sayed Mohammad Vakilzadeh and Dreyer, Maximilian and Jain, Aakriti and Wiegand, Thomas and Lapuschkin, Sebastian and Samek, Wojciech},
  booktitle={International Conference on Machine Learning},
  pages={135--168},
  year={2024},
  organization={PMLR}
}

@article{
bluecherdecoupling,
title={Decoupling Pixel Flipping and Occlusion Strategy for Consistent {XAI} Benchmarks},
author={Stefan Bluecher and Johanna Vielhaben and Nils Strodthoff},
journal={Transactions on Machine Learning Research},
issn={2835-8856},
year={2024},
url={https://openreview.net/forum?id=bIiLXdtUVM},
note={}
}

@inproceedings{li2025token,
  title={Token activation map to visually explain multimodal {{LLMs}}},
  author={Li, Yi and Wang, Hualiang and Ding, Xinpeng and Wang, Haonan and Li, Xiaomeng},
  booktitle={Proceedings of the IEEE/CVF International Conference on Computer Vision},
  pages={48--58},
  year={2025}
}

@inproceedings{zhang2025redundancy,
  title={From redundancy to relevance: Information flow in {{LVLMs}} across reasoning tasks},
  author={Zhang, Xiaofeng and Quan, Yihao and Shen, Chen and Yuan, Xiaosong and Yan, Shaotian and Xie, Liang and Wang, Wenxiao and Gu, Chaochen and Tang, Hao and Ye, Jieping},
  booktitle={Proceedings of the 2025 Conference of the Nations of the Americas Chapter of the Association for Computational Linguistics: Human Language Technologies (Volume 1: Long Papers)},
  pages={2289--2299},
  year={2025}
}

@inproceedings{petsiuk2018rise,
  author       = {Vitali Petsiuk and
                  Abir Das and
                  Kate Saenko},
  title        = {{RISE:} Randomized Input Sampling for Explanation of Black-box Models},
  booktitle    = {British Machine Vision Conference 2018, {BMVC} 2018, Newcastle, UK,
                  September 3-6, 2018},
  pages        = {151},
  publisher    = {{BMVA} Press},
  year         = {2018},
  url          = {http://bmvc2018.org/contents/papers/1064.pdf},
  bibsource    = {dblp computer science bibliography, https://dblp.org}
}

@article{grabisch1999axiomatic,
  title={An axiomatic approach to the concept of interaction among players in cooperative games},
  author={Grabisch, Michel and Roubens, Marc},
  journal={International Journal of game theory},
  volume={28},
  number={4},
  pages={547--565},
  year={1999},
  publisher={Springer}
}

@inproceedings{jacovi2020towards,
  title={Towards faithfully interpretable NLP systems: How should we define and evaluate faithfulness?},
  author={Jacovi, Alon and Goldberg, Yoav},
  booktitle={Proceedings of the 58th annual meeting of the association for computational linguistics},
  pages={4198--4205},
  year={2020}
}

@article{zhang2018top,
  title={Top-down neural attention by excitation backprop},
  author={Zhang, Jianming and Bargal, Sarah Adel and Lin, Zhe and Brandt, Jonathan and Shen, Xiaohui and Sclaroff, Stan},
  journal={International Journal of Computer Vision},
  volume={126},
  number={10},
  pages={1084--1102},
  year={2018},
  publisher={Springer}
}

@inproceedings{li2023evaluating,
  title={Evaluating object hallucination in large vision-language models},
  author={Li, Yifan and Du, Yifan and Zhou, Kun and Wang, Jinpeng and Zhao, Wayne Xin and Wen, Ji-Rong},
  booktitle={Proceedings of the 2023 conference on empirical methods in natural language processing},
  pages={292--305},
  year={2023}
}

@inproceedings{tong2024eyes,
  title={Eyes wide shut? exploring the visual shortcomings of multimodal llms},
  author={Tong, Shengbang and Liu, Zhuang and Zhai, Yuexiang and Ma, Yi and LeCun, Yann and Xie, Saining},
  booktitle={Proceedings of the IEEE/CVF conference on computer vision and pattern recognition},
  pages={9568--9578},
  year={2024}
}

@article{williams2010nonnegative,
  title={Nonnegative decomposition of multivariate information},
  author={Williams, Paul L and Beer, Randall D},
  journal={arXiv preprint arXiv:1004.2515},
  year={2010}
}

@inproceedings{wenderoth2025measuring,
  title={Measuring cross-modal interactions in multimodal models},
  author={Wenderoth, Laura and Hemker, Konstantin and Simidjievski, Nikola and Jamnik, Mateja},
  booktitle={Proceedings of the AAAI Conference on Artificial Intelligence},
  volume={39},
  number={20},
  pages={21501--21509},
  year={2025}
}

@article{banieckiexplaining,
  title={Explaining similarity in vision-language encoders with weighted Banzhaf interactions},
  author={Baniecki, Hubert and Muschalik, Maximilian and Fumagalli, Fabian and Hammer, Barbara and H{\"u}llermeier, Eyke and Biecek, Przemyslaw},
  journal={Advances in Neural Information Processing Systems},
  volume={38},
  pages={54529--54566},
  year={2026}
}

@article{roth1988introduction,
  title={Introduction to the Shapley value},
  author={Roth, Alvin E},
  journal={The Shapley value},
  volume={1},
  pages={3},
  year={1988},
  publisher={University of Cambridge Press, Cambridge}
}

@inproceedings{chen2025mllms,
    title={Where MLLMs Attend and What They Rely On: Explaining Autoregressive Token Generation},
    author={Chen, Ruoyu and Guo, Xiaoqing and Liu, Kangwei and Liang, Siyuan and Liu, Shiming and Zhang, Qunli and Zhang, Hua and Cao, Xiaochun},
    booktitle={Proceedings of the Computer Vision and Pattern Recognition Conference},
    year={2026}
}

@article{xing2025large,
  title={Where do large vision-language models look at when answering questions?},
  author={Xing, Xiaoying and Kuo, Chia-Wen and Fuxin, Li and Niu, Yulei and Chen, Fan and Li, Ming and Wu, Ying and Wen, Longyin and Zhu, Sijie},
  journal={arXiv preprint arXiv:2503.13891},
  year={2025}
}

@inproceedings{liu2024improved,
  title={Improved baselines with visual instruction tuning},
  author={Liu, Haotian and Li, Chunyuan and Li, Yuheng and Lee, Yong Jae},
  booktitle={Proceedings of the IEEE/CVF conference on computer vision and pattern recognition},
  pages={26296--26306},
  year={2024}
}

@article{bai2025qwen3,
  title={Qwen3-vl technical report},
  author={Bai, Shuai and Cai, Yuxuan and Chen, Ruizhe and Chen, Keqin and Chen, Xionghui and Cheng, Zesen and Deng, Lianghao and Ding, Wei and Gao, Chang and Ge, Chunjiang and others},
  journal={arXiv preprint arXiv:2511.21631},
  year={2025}
}

@article{wang2025internvl3,
  title={Internvl3. 5: Advancing open-source multimodal models in versatility, reasoning, and efficiency},
  author={Wang, Weiyun and Gao, Zhangwei and Gu, Lixin and Pu, Hengjun and Cui, Long and Wei, Xingguang and Liu, Zhaoyang and Jing, Linglin and Ye, Shenglong and Shao, Jie and others},
  journal={arXiv preprint arXiv:2508.18265},
  year={2025}
}

@article{wang2025multishap,
  title={Multishap: A shapley-based framework for explaining cross-modal interactions in multimodal ai models},
  author={Wang, Zhanliang and Wang, Kai},
  journal={arXiv preprint arXiv:2508.00576},
  year={2025}
}

@article{tong2024cambrian,
  title={Cambrian-1: A fully open, vision-centric exploration of multimodal llms},
  author={Tong, Shengbang and Brown, Ellis and Wu, Penghao and Woo, Sanghyun and Middepogu, Manoj and Akula, Sai C and Yang, Jihan and Yang, Shusheng and Iyer, Adithya and Pan, Xichen and others},
  journal={Advances in Neural Information Processing Systems},
  volume={37},
  pages={87310--87356},
  year={2024}
}

@article{chen2024we,
  title={Are we on the right way for evaluating large vision-language models?},
  author={Chen, Lin and Li, Jinsong and Dong, Xiaoyi and Zhang, Pan and Zang, Yuhang and Chen, Zehui and Duan, Haodong and Wang, Jiaqi and Qiao, Yu and Lin, Dahua and others},
  journal={Advances in Neural Information Processing Systems},
  volume={37},
  pages={27056--27087},
  year={2024}
}

@article{neuhaus2025repope,
  title={Repope: Impact of annotation errors on the pope benchmark},
  author={Neuhaus, Yannic and Hein, Matthias},
  journal={arXiv preprint arXiv:2504.15707},
  year={2025}
}

@inproceedings{lin2014microsoft,
  title={Microsoft coco: Common objects in context},
  author={Lin, Tsung-Yi and Maire, Michael and Belongie, Serge and Hays, James and Perona, Pietro and Ramanan, Deva and Doll{\'a}r, Piotr and Zitnick, C Lawrence},
  booktitle={European conference on computer vision},
  pages={740--755},
  year={2014},
  organization={Springer}
}

@article{chen2015microsoft,
  title={Microsoft coco captions: Data collection and evaluation server},
  author={Chen, Xinlei and Fang, Hao and Lin, Tsung-Yi and Vedantam, Ramakrishna and Gupta, Saurabh and Doll{\'a}r, Piotr and Zitnick, C Lawrence},
  journal={arXiv preprint arXiv:1504.00325},
  year={2015}
}

@article{agarwal2025rethinking,
author = {Agarwal, Chirag},
title = {Rethinking explainability in the era of multimodal AI},
year = {2026},
issue_date = {Fall 2026},
publisher = {John Wiley \& Sons, Inc.},
address = {USA},
volume = {47},
number = {3},
issn = {0738-4602},
url = {https://doi-org.proxy.bnl.lu/10.1002/aaai.70094},
doi = {10.1002/aaai.70094},
journal = {AI Mag.},
month = sep,
numpages = {11}
}

@article{xiao2025comprehensive,
  title={A comprehensive survey of large language models and multimodal large language models in medicine},
  author={Xiao, Hanguang and Zhou, Feizhong and Liu, Xingyue and Liu, Tianqi and Li, Zhipeng and Liu, Xin and Huang, Xiaoxuan},
  journal={Information Fusion},
  volume={117},
  pages={102888},
  year={2025},
  publisher={Elsevier}
}

@article{comanici2025gemini,
  title={Gemini 2.5: Pushing the frontier with advanced reasoning, multimodality, long context, and next generation agentic capabilities},
  author={Comanici, Gheorghe and Bieber, Eric and Schaekermann, Mike and Pasupat, Ice and Sachdeva, Noveen and Dhillon, Inderjit and Blistein, Marcel and Ram, Ori and Zhang, Dan and Rosen, Evan and others},
  journal={arXiv preprint arXiv:2507.06261},
  year={2025}
}

@article{team2026qwen3,
  title={Qwen3. 5-omni technical report},
  author={Team, Qwen},
  journal={arXiv preprint arXiv:2604.15804},
  year={2026}
}

@article{li2024mmro,
  title={Mmro: Are multimodal llms eligible as the brain for in-home robotics?},
  author={Li, Jinming and Zhu, Yichen and Xu, Zhiyuan and Gu, Jindong and Zhu, Minjie and Liu, Xin and Liu, Ning and Peng, Yaxin and Feng, Feifei and Tang, Jian},
  journal={arXiv preprint arXiv:2406.19693},
  year={2024}
}

@inproceedings{cui2024survey,
  title={A survey on multimodal large language models for autonomous driving},
  author={Cui, Can and Ma, Yunsheng and Cao, Xu and Ye, Wenqian and Zhou, Yang and Liang, Kaizhao and Chen, Jintai and Lu, Juanwu and Yang, Zichong and Liao, Kuei-Da and others},
  booktitle={Proceedings of the IEEE/CVF winter conference on applications of computer vision},
  pages={958--979},
  year={2024}
}

@article{li2024naturalbench,
  title={Naturalbench: Evaluating vision-language models on natural adversarial samples},
  author={Li, Baiqi and Lin, Zhiqiu and Peng, Wenxuan and Nyandwi, Jean de Dieu and Jiang, Daniel and Ma, Zixian and Khanuja, Simran and Krishna, Ranjay and Neubig, Graham and Ramanan, Deva},
  journal={Advances in Neural Information Processing Systems},
  volume={37},
  pages={17044--17068},
  year={2024}
}

@article{zheng2025mllms,
  title={Mllms are deeply affected by modality bias},
  author={Zheng, Xu and Liao, Chenfei and Fu, Yuqian and Lei, Kaiyu and Lyu, Yuanhuiyi and Jiang, Lutao and Ren, Bin and Chen, Jialei and Wang, Jiawen and Li, Chengxin and others},
  journal={arXiv preprint arXiv:2505.18657},
  year={2025}
}

@inproceedings{chen2024quantifying,
  title={Quantifying and mitigating unimodal biases in multimodal large language models: A causal perspective},
  author={Chen, Meiqi and Cao, Yixin and Zhang, Yan and Lu, Chaochao},
  booktitle={Findings of the Association for Computational Linguistics: EMNLP 2024},
  pages={16449--16469},
  year={2024}
}

@inproceedings{lee2025vlind,
  title={Vlind-bench: Measuring language priors in large vision-language models},
  author={Lee, Kang-il and Kim, Minbeom and Yoon, Seunghyun and Kim, Minsung and Lee, Dongryeol and Koh, Hyukhun and Jung, Kyomin},
  booktitle={Findings of the Association for Computational Linguistics: NAACL 2025},
  pages={4129--4144},
  year={2025}
}

@inproceedings{liu2024insight,
    title = "Insight Over Sight: Exploring the Vision-Knowledge Conflicts in Multimodal {LLM}s",
    author = "Liu, Xiaoyuan  and
      Wang, Wenxuan  and
      Yuan, Youliang  and
      Huang, Jen-tse  and
      Liu, Qiuzhi  and
      He, Pinjia  and
      Tu, Zhaopeng",
    editor = "Che, Wanxiang  and
      Nabende, Joyce  and
      Shutova, Ekaterina  and
      Pilehvar, Mohammad Taher",
    booktitle = "Proceedings of the 63rd Annual Meeting of the Association for Computational Linguistics (Volume 1: Long Papers)",
    month = jul,
    year = "2025",
    address = "Vienna, Austria",
    publisher = "Association for Computational Linguistics",
    url = "https://aclanthology.org/2025.acl-long.872/",
    doi = "10.18653/v1/2025.acl-long.872",
    pages = "17825--17846",
    ISBN = "979-8-89176-251-0",
}

@article{muschalik2024shapiq,
  title={shapiq: Shapley interactions for machine learning},
  author={Muschalik, Maximilian and Baniecki, Hubert and Fumagalli, Fabian and Kolpaczki, Patrick and Hammer, Barbara and H{\"u}llermeier, Eyke},
  journal={Advances in Neural Information Processing Systems},
  volume={37},
  pages={130324--130357},
  year={2024}
}

@article{mitchell2022sampling,
  title={Sampling permutations for shapley value estimation},
  author={Mitchell, Rory and Cooper, Joshua and Frank, Eibe and Holmes, Geoffrey},
  journal={Journal of Machine Learning Research},
  volume={23},
  number={43},
  pages={1--46},
  year={2022}
}

@inproceedings{covert2021improving,
  title={Improving kernelshap: Practical shapley value estimation using linear regression},
  author={Covert, Ian and Lee, Su-In},
  booktitle={International conference on artificial intelligence and statistics},
  pages={3457--3465},
  year={2021},
  organization={PMLR}
}

@inproceedings{belsare2026gridvqa,
  title={GridVQA-X: A Diagnostic Framework for Evaluating Multimodal Explainability Methods},
  author={Belsare, Sujay and Nikhil, Sudarshan and Kumar, Sushant and Kumaraguru, Ponnurangam and Agarwal, Chirag},
  booktitle={European Conference on Computer Vision},
  pages={398--429},
  year={2026},
  organization={Springer}
}

@article{hooker2019benchmark,
  title={A benchmark for interpretability methods in deep neural networks},
  author={Hooker, Sara and Erhan, Dumitru and Kindermans, Pieter-Jan and Kim, Been},
  journal={Advances in neural information processing systems},
  volume={32},
  year={2019}
}

@article{deng2024unifying,
  title={Unifying fourteen post-hoc attribution methods with taylor interactions},
  author={Deng, Huiqi and Zou, Na and Du, Mengnan and Chen, Weifu and Feng, Guocan and Yang, Ziwei and Li, Zheyang and Zhang, Quanshi},
  journal={IEEE Transactions on Pattern Analysis and Machine Intelligence},
  volume={46},
  number={7},
  pages={4625--4640},
  year={2024},
  publisher={IEEE}
}

%%%%%%%%%%%%%%%%%%%%%%%%%%%%%%%%%%%%%%%%%%%%%%%%%%%%%%%%%%%%
\newpage
\appendix

\section{Reproducibility and implementation details}\label{appendix_reproducibility}
\subsection{VLM Configurations and Generation Strategy}
\label{app:vlm_config}

\textbf{Model Checkpoints:} We evaluate all methods across three distinct Vision-Language Model architectures using their official Hugging Face checkpoints: \texttt{llava-hf/llava-1.5-7b-hf}, \texttt{Qwen/Qwen2.5-VL-3B-Instruct}, and \texttt{OpenGVLab/InternVL2-2B}. 

\textbf{Image Processing and Memory Constraints:} Due to the immense VRAM overhead required to compute attributions, particularly for gradient-based explainers that retain massive computation graphs during backpropagation, we applied necessary constraints to the dynamic resolution pipelines of the newer VLMs. This was critical to prevent Out-of-Memory (OOM) errors during evaluation. Specifically, for InternVL2-2B, we restricted the image processor by capping the number of tiles to \texttt{max\_patches = 2}. Similarly, for Qwen2.5-VL, we explicitly bounded the \texttt{max\_pixels=313600} parameter to restrict unbounded visual tile generation. These constraints ensured stable explainer execution while maintaining sufficient visual fidelity for accurate VQA reasoning.

\textbf{Decoding Parameters:} To ensure deterministic evaluation of the underlying reasoning pathways, we utilized default generation parameters for all models with greedy decoding and bounded the sequence length to $max\_new\_tokens = 32$. 

\textbf{Prompting and Target Extraction:} To preserve the natural distribution of the models' outputs, we minimized prompt engineering and allowed the models to generate freely where possible. For binary Yes/No tasks (e.g., RePOPE), questions were passed to the models without any formatting pre-prompts. For multiple-choice Visual Question Answering (CVBench and MMStar), we appended the following directive to standardize the output format: \textit{"\textbackslash nAnswer directly with only the letter inside parentheses, and nothing else. \textbackslash n"} (Note: For LLaVA-1.5 specifically, we appended only \textit{"Answer:"} at the end of the original prompt, as empirical testing revealed this significantly stabilizes its VQA performance). Following generation, we utilized regex-based parsing to isolate the specific target token (e.g., "Yes", "No", or the choice letter "A", "B", etc.). All explainer attributions were explicitly computed with respect to these isolated target tokens.

\subsection{Explainer Implementations and Autoregressive Adaptation}
\label{app:explainer_config}
Standard XAI methods are historically designed for discriminative models (single forward pass). To apply them to modern VLMs, we adapted all explainers to support autoregressive decoding. Specifically, the explanation for a generated token $t$ is computed using the full context of the prompt and all previously generated tokens $\{0, \dots, t-1\}$. 

With the exception of TAM \cite{li2025token} (which utilized its official repository\footnote{\url{https://github.com/xmed-lab/TAM/tree/main}}), all explainers were custom-reimplemented or heavily adapted to ensure standardized extraction across the three diverse VLM architectures. Specific implementation details include:

\begin{itemize}
    \item \textbf{AttnLRP (LxT):} We utilized the official \texttt{lxt}\footnote{\url{https://lxt.readthedocs.io/en/latest/index.html}} library but extended it to support the specific Hugging Face architectures of our evaluated VLMs. Following standard LRP protocols, we applied \texttt{lxt.efficient.patches} to modify the computational graph. Specifically, we patched the attention functions (e.g., \texttt{ALL\_ATTENTION\_FUNCTIONS}) to apply the uniform rule to matrix multiplications, modified the RMS-Norm layers to halt gradient flow through variance computation, and patched the MLP blocks to apply the identity rule to non-linear activations. 
    
    \textit{Note on Gradient Shattering:} The AttnLRP authors recommend applying the Gamma rule to Vision Transformer layers to mitigate gradient shattering and denoise heatmaps. However, applying this rule to the dense, high-resolution visual encoders in our models resulted in intractable numerical instabilities (NaN values) that could not be resolved via hyperparameter tuning. Consequently, we omitted the Gamma rule for visual layers, relying exclusively on the core LRP propagation rules. We note that this architectural constraint likely contributes to the specific pixel-attribution distributions observed for AttnLRP in our benchmark (see Appendix \ref{appendix_quality}).

    \item \textbf{LLaVA-CAM and GradCAM:} LLaVA-CAM was implemented leveraging the codebase from EAGLE \cite{chen2025mllms}\footnote{\url{https://github.com/RuoyuChen10/EAGLE/blob/master/baselines/llavacam.py}}. As recommended by the authors of LLaVA-CAM \cite{zhang2025redundancy} to prevent information dilution in the deepest layers, we extracted gradients from the early \texttt{post\_attention\_layer\_norm} (specifically layer 7). GradCAM was implemented identically to LLaVA-CAM, but with the explicit removal of the specialized smoothing operation.
    
    \item \textbf{Integrated Gradients and InputxGradients:} We extended the implementation in Captum \footnote{\url{https://captum.ai/}} The integration path of Integrated Gradients was approximated using $N=5$ Riemann steps. The visual "zero-state" baseline was defined as a zero-tensor (pure black image), and the textual baseline consisted of standard padding tokens. 
    
    \item \textbf{Attention Rollout and GradRollout:} To extract the necessary attention maps and gradients across deep layers, we implemented custom PyTorch forward hooks. For specific architectures where attention weights are not natively retained in the computational graph (e.g., Qwen2.5-VL), we directly modified the forward pass of the attention modules to cache the weights during inference, enabling full gradient tracking for rollout methods.
\end{itemize}

\subsection{Faithfulness metric implementation details}
\label{app:metric_implementation}

To execute the perturbation sweeps required for both standard unimodal evaluation and our proposed metric ($\mathcal{F}_{syn}$), we standardized the masking strategies and confidence scoring functions across all instances.

\textbf{Modality masking and zero-state definitions:}
XAI algorithms produce attribution maps matching the specific tensor dimensions of the input modalities (e.g., the sequence length of \texttt{input\_ids} for text, and the patch grid of \texttt{pixel\_values} for images). 
\begin{itemize}
    \item \textbf{Textual zero-state:} When performing textual perturbation, we first filter the \texttt{input\_ids} sequence to isolate the specific semantic tokens corresponding to the user's query, explicitly ignoring instruction templates, system prompts, and encoded visual tokens. During deletion, the targeted semantic tokens are masked by replacing them with the model's specific \texttt{pad\_token\_id}.
    \item \textbf{Visual zero-state:} Rather than replacing visual patches with a pure zero-tensor (which can push the VLM heavily out-of-distribution and cause erratic activations), we define the visual zero-state using a severe low-pass filter. Specifically, the original image is subjected to a Gaussian blur with a radius of $r=30$ before being processed by the vision encoder into \texttt{pixel\_values}. During visual perturbation, masked patches are replaced by their corresponding blurred patches, effectively destroying high-frequency semantic features while preserving the general color palette and global illumination.
\end{itemize}

\textbf{VLM confidence scoring} ($f(I, T)$):
Shapley Interaction Index and faithfulness metrics strictly require the scoring function $f(I, T)$ to yield a bounded scalar, typically in the range $[0, 1]$. Because modern VLMs generate answers autoregressively rather than outputting a single classification logit, we implemented a custom, vectorized scoring function to extract the joint probability of the generated target answer. For a given instance, let $Y = \{y_1, y_2, \dots, y_m\}$ represent the sequence of target tokens generated by the VLM (e.g., the tokens comprising "Yes", "No", or "A"). During each perturbation step, we feed the partially masked inputs $(I_k, T_k)$ alongside the full context and previously generated tokens into the VLM. We extract the raw logits for the target positions, apply a \texttt{log\_softmax} to convert them into log-probabilities, and gather the specific log-probabilities corresponding to the ground-truth target tokens. To obtain the final confidence score $f(I_k, T_k)$ representing the model's certainty in generating the exact target answer under the perturbed state, we sum the log-probabilities over the target sequence length and project the result back into the linear probability space via an exponential function:
\begin{equation}
    f(I_k, T_k) = \exp \left( \sum_{i=1}^{m} \log P(y_i \mid I_k, T_k, y_{<i}) \right)
\end{equation}

This operation effectively computes the joint probability of the entire target sequence. Bounding the confidence score between $[0, 1]$ ensures the Harsanyi dividend remains mathematically stable and directly interpretable as a percentage of predictive confidence during the integration of $\mathcal{F}_{syn}$.

\textbf{Metric Hyperparameters and Environment:} For the calculation of the Synergistic Faithfulness metric ($\mathcal{F}_{syn}$), the continuous perturbation trajectory was approximated using $K=11$ discrete Riemann steps (e.g., $k \in \{0.0, 0.1, \dots, 1.0\}$). All XAI algorithms and autoregressive generation loops were implemented using PyTorch and the Hugging Face \texttt{transformers} library (version 4.57.6). The \texttt{pad\_token\_id} was dynamically extracted from the respective VLM's processor to ensure proper semantic masking during the text perturbation phase.

\subsection{Hardware and Computational Budget}
\label{app:hardware}
All experiments, including autoregressive VLM inference and the extensive perturbation sweeps required for evaluating the $\mathcal{F}_{syn}$ metric and baseline XAI methods, were executed on a dedicated high-performance computing node equipped with 4 $\times$ NVIDIA L40S GPUs (48GB VRAM per GPU). The evaluation pipeline was implemented in Python 3.12, heavily utilizing PyTorch and the Hugging Face \texttt{transformers} library for model distribution and inference. We estimate the total computational budget required to execute the complete benchmark, spanning three VLM architectures, nine explainers, and the extensive dataset splits to be approximately 300 GPU hours.

\subsection{Dataset Evaluation}
\label{app:datasets_hyperparams}

\textbf{Dataset Scale and Sampling:} To rigorously validate the $\mathcal{F}_{syn}$ metric without the statistical variance introduced by random sub-sampling, we evaluated all explainers on the \textit{entirety} of the respective dataset evaluation splits. This comprises the complete RePOPE, CVBench, and MMStar datasets.

\section{Hyperparameter Sensitivity and Extended Analysis}

\subsection{Hyperparameter $K$ Stability}
\label{app:hyperparameter_k}

Our metric approximates the continuous perturbation trajectory via a Riemann sum over $K$ discrete steps. To ensure that the explainer rankings are not sensitive to this hyperparameter while maintaining computational efficiency, we evaluated both the ranking stability and the computational cost across $K \in \{3, 5, 7, 11, 15, 20\}$ on the MMStar dataset using InternVL-2B. We compare the explainer hierarchies at each step against the high-resolution $K=20$ baseline using Kendall's rank correlation ($\tau$).

\begin{figure}[htbp]
\centering
\begin{minipage}[b]{0.48\linewidth}
\centering
\includegraphics[width=\linewidth]{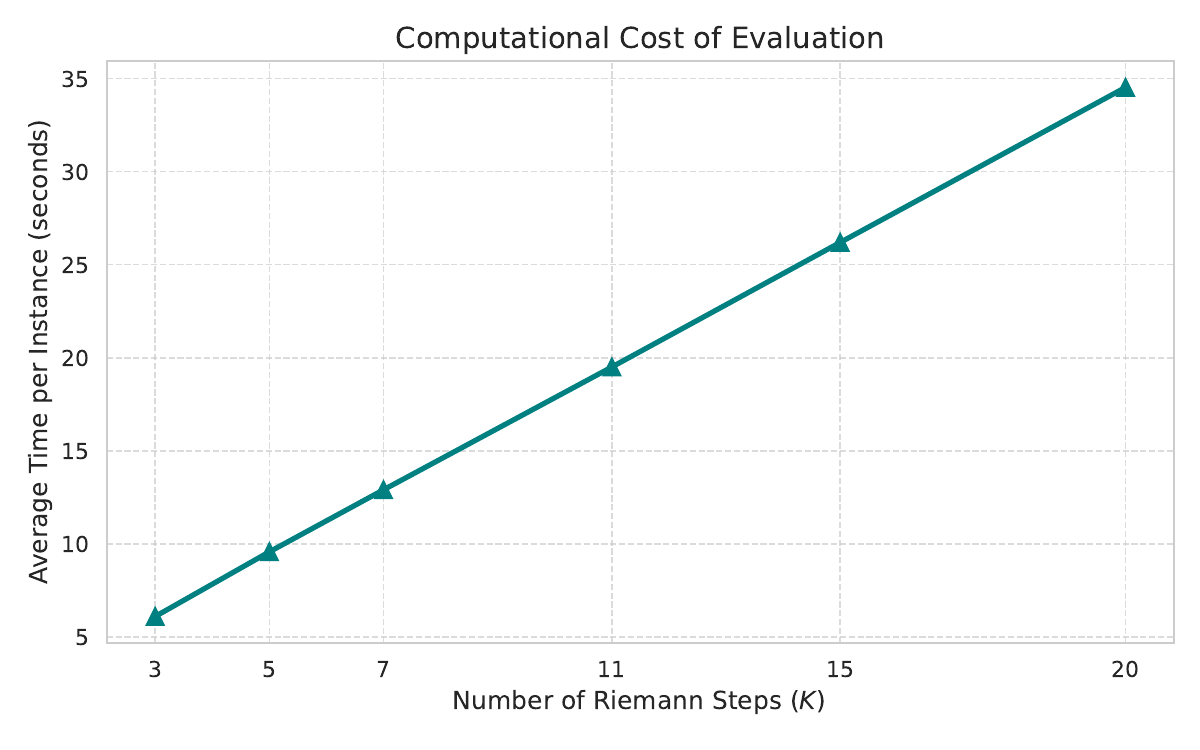}
\end{minipage}
\hfill
\begin{minipage}[b]{0.48\linewidth}
\centering
\includegraphics[width=\linewidth]{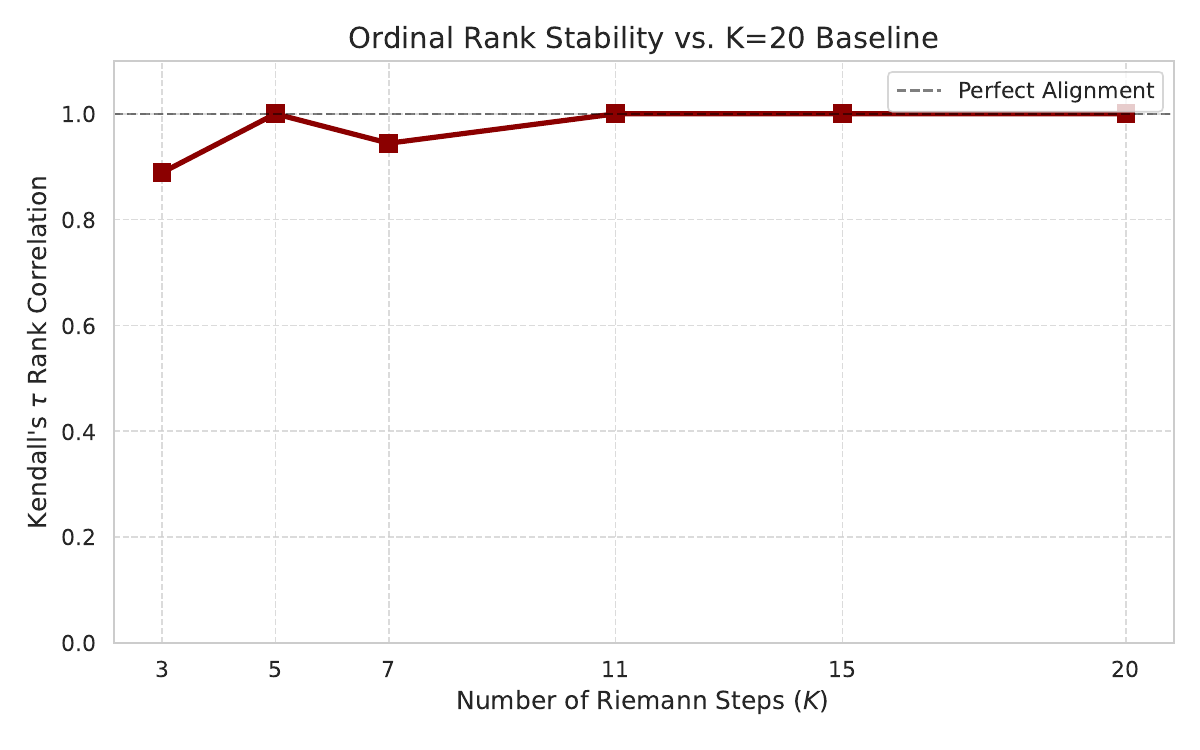}
\end{minipage}
\caption{Trade-off between computational cost and rank stability across integration steps ($K$). \textbf{Left:} The average evaluation time per instance scales linearly with the number of Riemann steps ($K$). \textbf{Right:} The explainer hierarchy exhibits estimation variance at lower steps but permanently converges to perfect ordinal alignment ($\tau = 1.0$) at $K=11$.}
\label{fig:k_stability}
\end{figure}

As visualized in Figure \ref{fig:k_stability} (right), lower values ($K \le 7$) exhibit estimation variance, leading to slight rank shuffling. At $K=11$, the mathematical integration stabilizes, achieving a perfect and permanent rank alignment ($\tau = 1.0$) with the $K=20$ baseline. Simultaneously, Figure \ref{fig:k_stability} (left) demonstrates that the computational cost of evaluation increases linearly with the number of steps. Consequently, we select $K=11$ for our main experiments to optimize the necessary trade-off between strict rank stability and computational efficiency.

\takeaway{Using $K=11$ Riemann steps to compute $\mathcal{F}_{syn}$ provides the optimal operational balance, achieving stable ordinal rank convergence ($\tau=1.0$) with the high-resolution baseline while minimizing computational cost.} 

\subsection{Zero-State Baseline Sensitivity}
\label{app:zero_state}

In cooperative game theory, the choice of the zero-state defines the mathematical "absence" of a feature when computing marginal contributions. For visual tokens, our default baseline utilizes a Gaussian blur, which removes semantic edges while preserving the macroscopic color distribution, thereby minimizing out-of-distribution (OOD) shock to the model.

To evaluate the sensitivity of $\mathcal{F}_{syn}$ to this choice, we conducted an ablation comparing the Gaussian blur against a strict zero-padding baseline (all masked pixels set to $0.0$). Substituting the baseline with zero-padding introduced an absolute score shift averaging $+7.61\%$ across all methods. This absolute shift is an expected OOD artifact \cite{hooker2019benchmark}, as replacing features with pure black pixels destroys the natural data manifold and artificially drops the model's confidence. 

However, the relative hierarchy of the explainers remained highly robust. We measured a Kendall rank correlation of $\tau = 0.833$ and a Spearman rank correlation of $\rho = 0.933$ between the explainer rankings derived under the blur and zero-pad conditions. This confirms that while the absolute $\mathcal{F}_{syn}$ scores are inherently tied to the chosen baseline, the comparative evaluation of the explainers is stable. For textual features, we consistently substitute masked tokens with standard \texttt{[PAD]} tokens to preserve the sequence length and maintain the transformer's causal positional logic without introducing structural syntax errors.

\takeaway{While substituting the Gaussian blur zero-state with strict zero-padding induces an expected absolute score shift (+7.61\%) due to out-of-distribution artifacts, the relative explainer hierarchy remains highly robust ($\rho=0.933$), confirming that $\mathcal{F}_{syn}$ rankings are stable across baseline definitions.}

\section{Comprehensive statistical analysis}\label{appendix_statistical}

\subsection{Linear Mixed-effects Model (LMM) Formulation}
\label{app:lmm_formulation}
A critical challenge in multimodal benchmarking is variance entanglement: the raw faithfulness score of an explainer is heavily confounded by the inherent reasoning capacity of the chosen VLM and the specific modality biases of the dataset. Furthermore, standard statistical tests (e.g., ANOVA) assume data independence, which is violated in our benchmark because the exact same image-question pairs are repeatedly evaluated across different explainers and models. To rigorously isolate the true algorithmic superiority of the XAI methods, we fit a Linear Mixed-Effects Model (LMM) to our evaluation data. The LMM allows us to treat the \textit{Explainer} as a fixed effect, while modeling the \textit{dataset}, \textit{model architecture}, and \textit{instance} $(I, T)$ as random effects to absorb contextual variance. The global model is formulated as:

\begin{equation}
    \mathcal{F}_{syn}^{(i, j, k, l)} = \beta_0 + \beta_i^{Explainer_i} + u_j + v_k + w_l + \epsilon_{ijkl} 
\end{equation}

where:
\begin{itemize}
    \item $\mathcal{F}_{syn}^{(i, j, k, l)} $ is the faithfulness score (e.g., $\mathcal{F}_{syn}$) for explainer $i$, instance $j$, model $k$, and dataset $m$.
    \item $\beta_0$ is the global intercept (defined as the random baseline explainer).
    \item $\beta_i^{E_i}$ is the fixed effect of the explainer type $i$ (the core algorithmic contribution we aim to measure).
    \item $u_j \sim \mathcal{N}(0, \sigma_u^2)$ is the random intercept for the specific multimodal instance.
    \item $v_k \sim \mathcal{N}(0, \sigma_v^2)$ is the random intercept for the VLM architecture.
    \item $w_l \sim \mathcal{N}(0, \sigma_w^2)$ is the random intercept for the dataset.
    \item $\epsilon_{ijkl} \sim \mathcal{N}(0, \sigma_\epsilon^2)$ is the residual error.
\end{itemize}

By extracting the estimated marginal means from this LMM, the resulting explainer rankings reflect pure algorithmic performance, statistically sanitized of the dataset biases and model-specific idiosyncrasies that typically distort traditional leaderboards.

\subsection{Dataset-Specific LMM results}
\label{app:fsyn_lmm_datasets}

\begin{figure}[t]
  \centering
  \begin{minipage}[b]{0.32\linewidth}
    \centering
    \includegraphics[width=\linewidth]{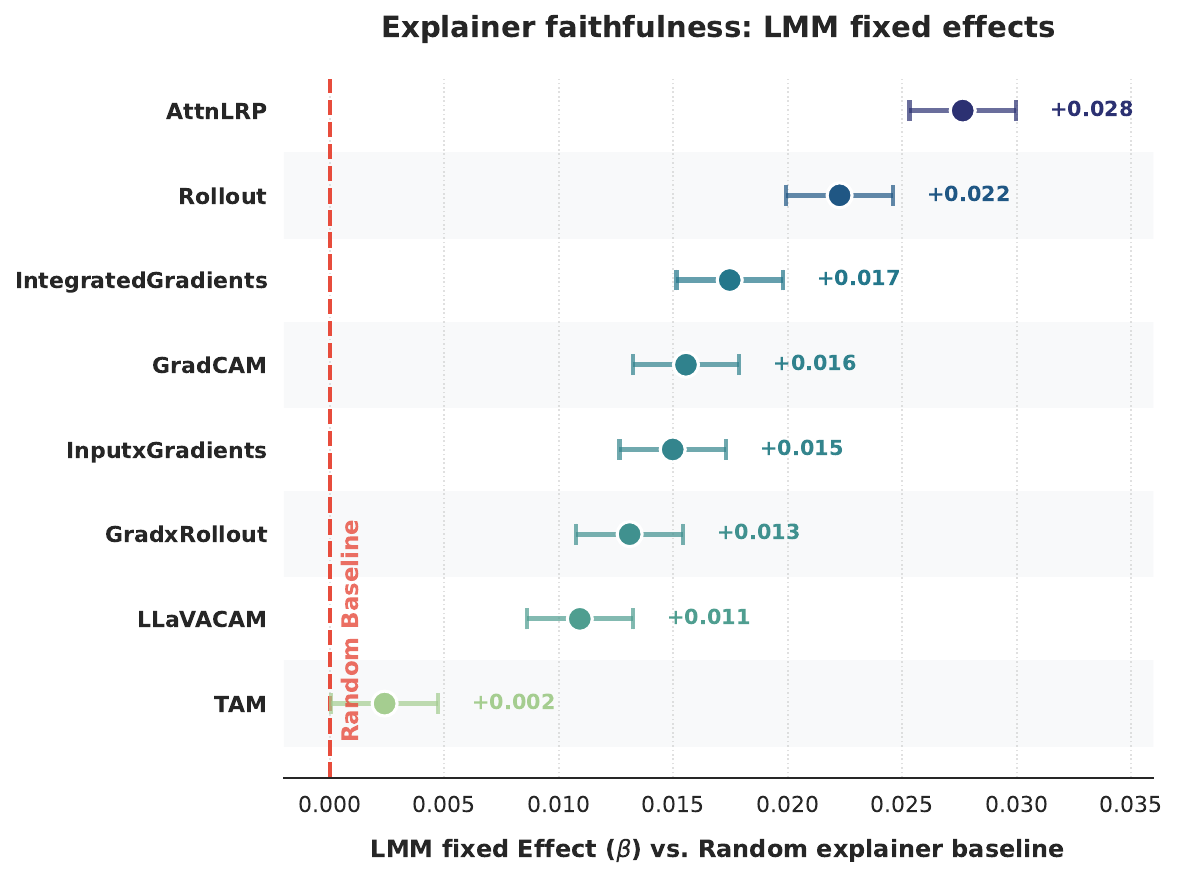}
    \centerline{(a) RePOPE}
  \end{minipage}\hfill
  \begin{minipage}[b]{0.32\linewidth}
    \centering
    \includegraphics[width=\linewidth]{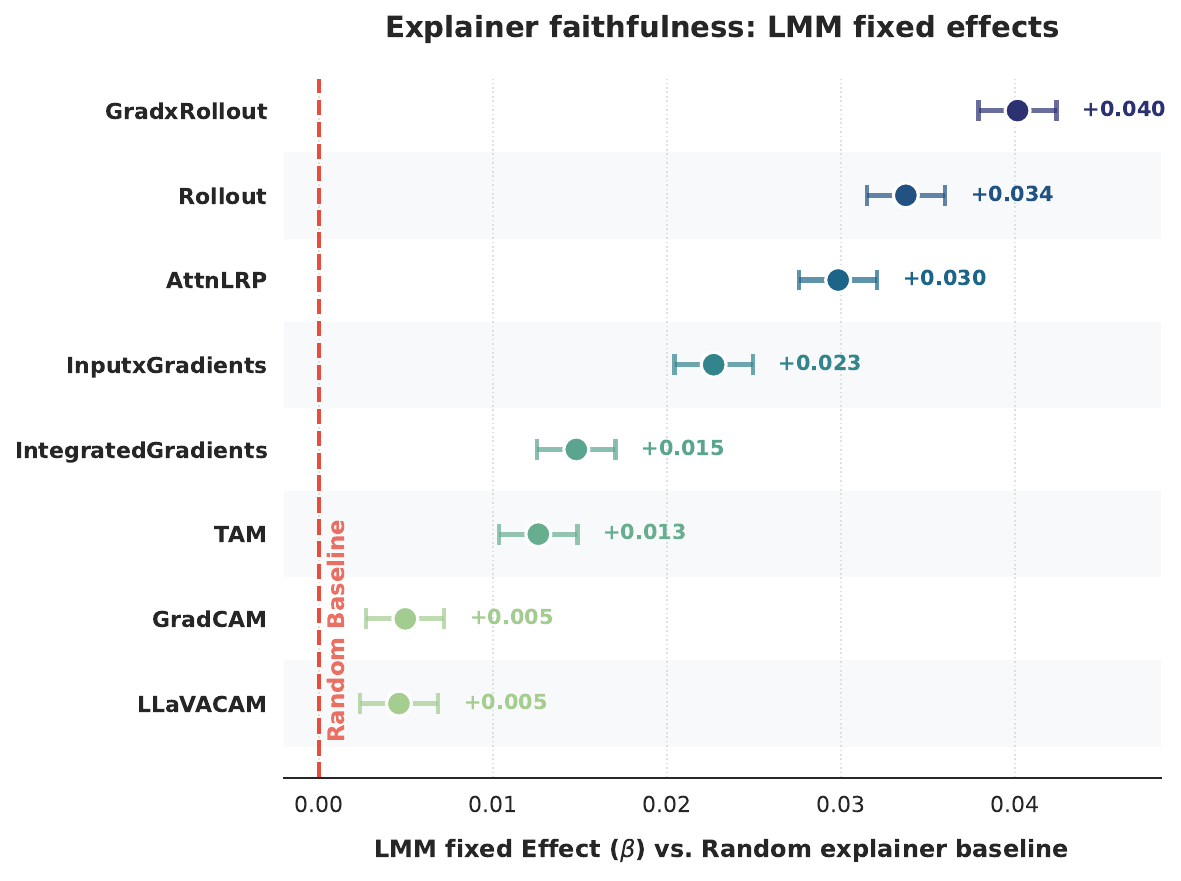}
    \centerline{(b) CVBench}
  \end{minipage}\hfill
  \begin{minipage}[b]{0.32\linewidth}
    \centering
    \includegraphics[width=\linewidth]{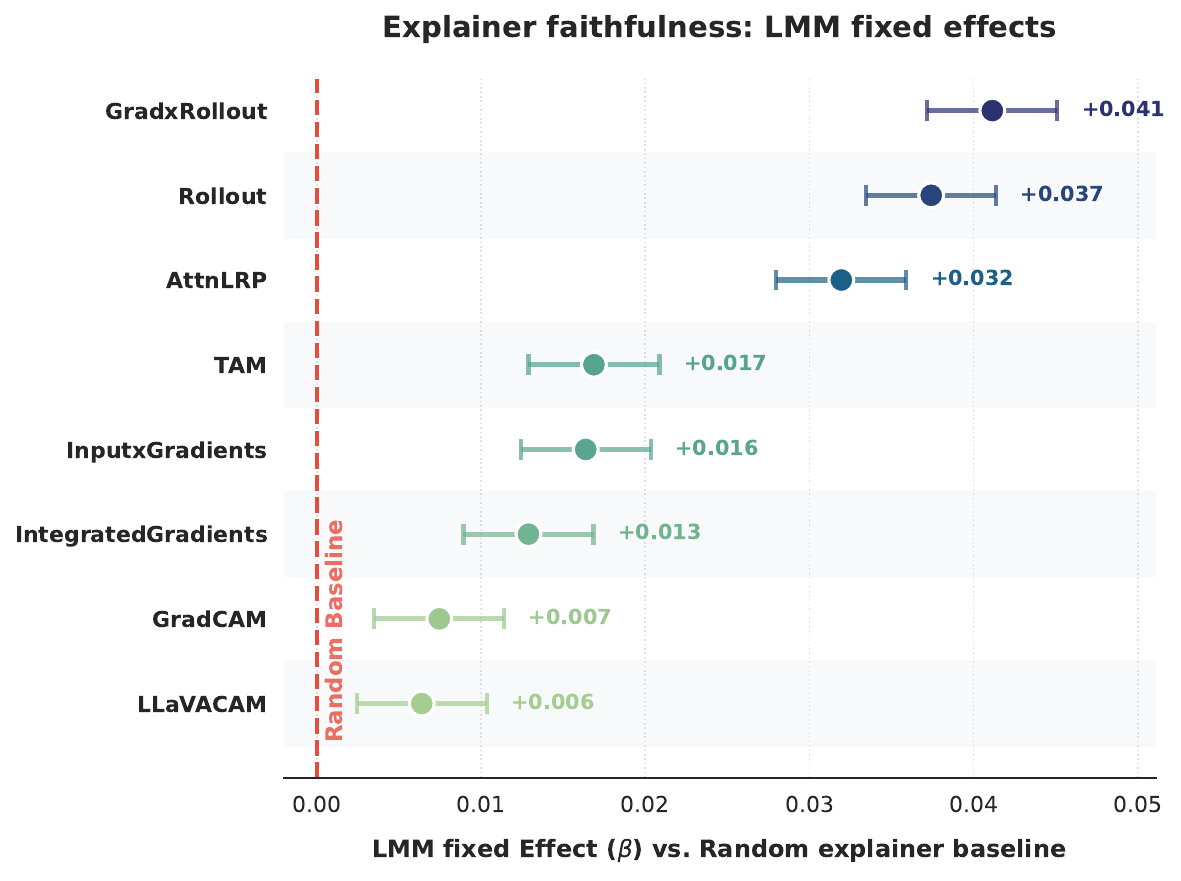}
    \centerline{(c) MMStar}
  \end{minipage}
  \caption{Forest plots of the dataset-specific LMM fixed effects ($\beta$) for Synergistic Faithfulness ($\mathcal{F}_{syn}$). The dots represent the estimated algorithmic contribution of each explainer relative to the Random baseline ($\beta=0$), flanked by 95\% confidence intervals.}
  \label{fig:lmm_plots_datasets}
\end{figure}

To investigate how dataset-specific modality biases influence the actual cross-modal synergy of explainers, we decomposed our global LMM and fitted independent mixed-effects models for the $\mathcal{F}_{syn}$ scores on RePOPE, CVBench, and MMStar. For these dataset-specific models, the dataset random intercept ($w_m$) was removed, retaining VLM architecture and Instance ID as random effects to absorb contextual variance.

Figure \ref{fig:lmm_plots_datasets} presents the estimated marginal means (LMM coefficients) and 95\% Confidence Intervals for each explainer across the three datasets. To provide comprehensive statistical rigor, Table \ref{tab:lmm_pvalues} reports the exact $\beta$ coefficients, Standard Errors (SE), and $p$-values for each explainer.

\begin{table}[h]
\centering
\caption{Dataset-Specific LMM Statistics for $\mathcal{F}_{syn}$. All values are calculated relative to the Random Attribution baseline. Significance thresholds: $^* p < 0.05$, $^{**} p < 0.01$, $^{***} p < 0.001$.}
\label{tab:lmm_pvalues}
\resizebox{\textwidth}{!}{%
\begin{tabular}{l | r r r | r r r | r r r}
\toprule
\textbf{} & \multicolumn{3}{c|}{\textbf{RePOPE}} & \multicolumn{3}{c|}{\textbf{CVBench}} & \multicolumn{3}{c}{\textbf{MMStar}} \\
\cmidrule(lr){2-4} \cmidrule(lr){5-7} \cmidrule(lr){8-10}
\textbf{Explainer} & $\beta$ Coef & SE & $p$-value & $\beta$ Coef & SE & $p$-value & $\beta$ Coef & SE & $p$-value \\
\midrule
AttnLRP & \bf 0.028 & 0.001 & $<0.001^{***}$ & 0.030 & 0.001 & $<0.001^{***}$ & 0.032 & 0.002 & $<0.001^{***}$ \\
Grad$\times$Rollout & 0.013 & 0.001 & $<0.001^{***}$ & \bf 0.040 & 0.001 & $<0.001^{***}$ & \bf 0.041 & 0.002 & $<0.001^{***}$ \\
Rollout & \underline{0.022} & 0.001 & $<0.001^{***}$ & \underline{0.034} & 0.001 & $<0.001^{***}$ & \underline{0.037} & 0.002 & $<0.001^{***}$ \\
\midrule
TAM & 0.002 & 0.001 & $0.044^{*}$ & 0.013 & 0.001 & $<0.001^{***}$ & 0.017 & 0.002 & $<0.001^{***}$ \\
LLaVA-CAM & 0.011 & 0.001 & $<0.001^{***}$ & 0.005 & 0.001 & $<0.001^{***}$ & 0.006 & 0.002 & $0.002^{**}$ \\
\midrule
GradCAM & 0.016 & 0.001 & $<0.001^{***}$ & 0.005 & 0.001 & $<0.001^{***}$ & 0.007 & 0.002 & $<0.001^{***}$ \\
Input$\times$Grad & 0.015 & 0.001 & $<0.001^{***}$ & 0.023 & 0.001 & $<0.001^{***}$ & 0.016 & 0.002 & $<0.001^{***}$ \\
Integ. Gradients & 0.017 & 0.001 & $<0.001^{***}$ & 0.015 & 0.001 & $<0.001^{***}$ & 0.013 & 0.002 & $<0.001^{***}$ \\
\bottomrule
\end{tabular}%
}
\end{table}

This granular statistical view reveals a critical trend regarding algorithmic scalability. As the evaluation shifts to datasets that strictly enforce non-bypassable cross-modal reasoning (MMStar), the performance gap widens dramatically. These results confirm the findings in Section \ref{experiments}: current VLM-native explainers fail to capture genuine cross-modal interactions.

\subsection{Unimodal Rank Instability (Kendall's $\tau$)}
\label{app:unimodal_instability}

\begin{figure}[t]
  \centering
  \begin{minipage}[b]{0.32\linewidth}
    \centering
    \includegraphics[width=\linewidth]{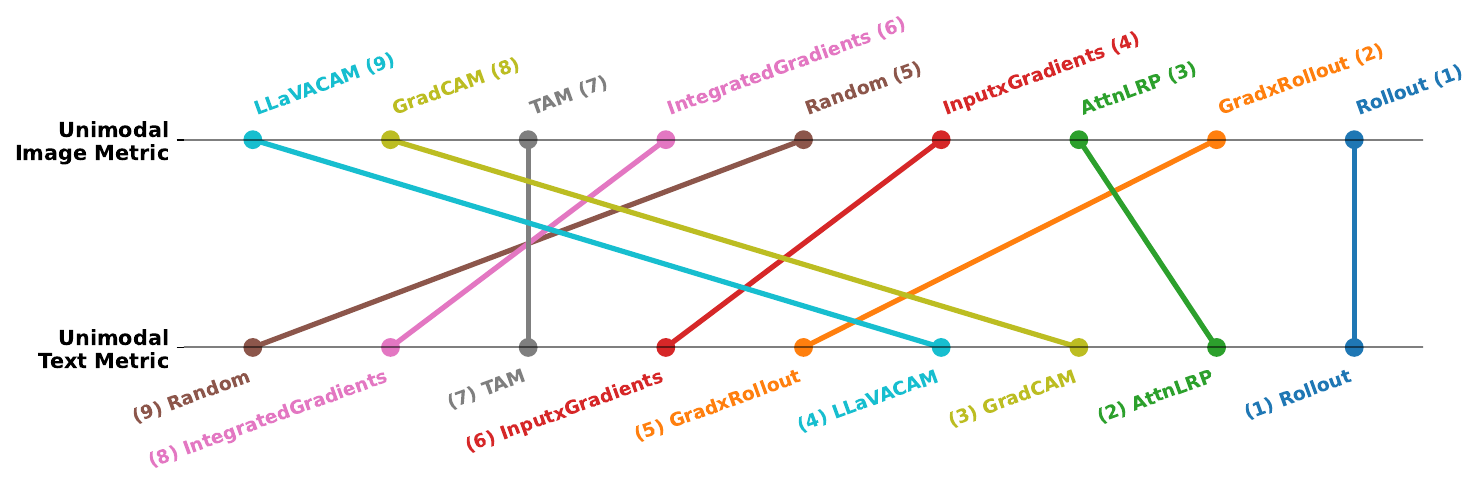}
    \centerline{(a) RePOPE}
  \end{minipage}\hfill
  \begin{minipage}[b]{0.32\linewidth}
    \centering
    \includegraphics[width=\linewidth]{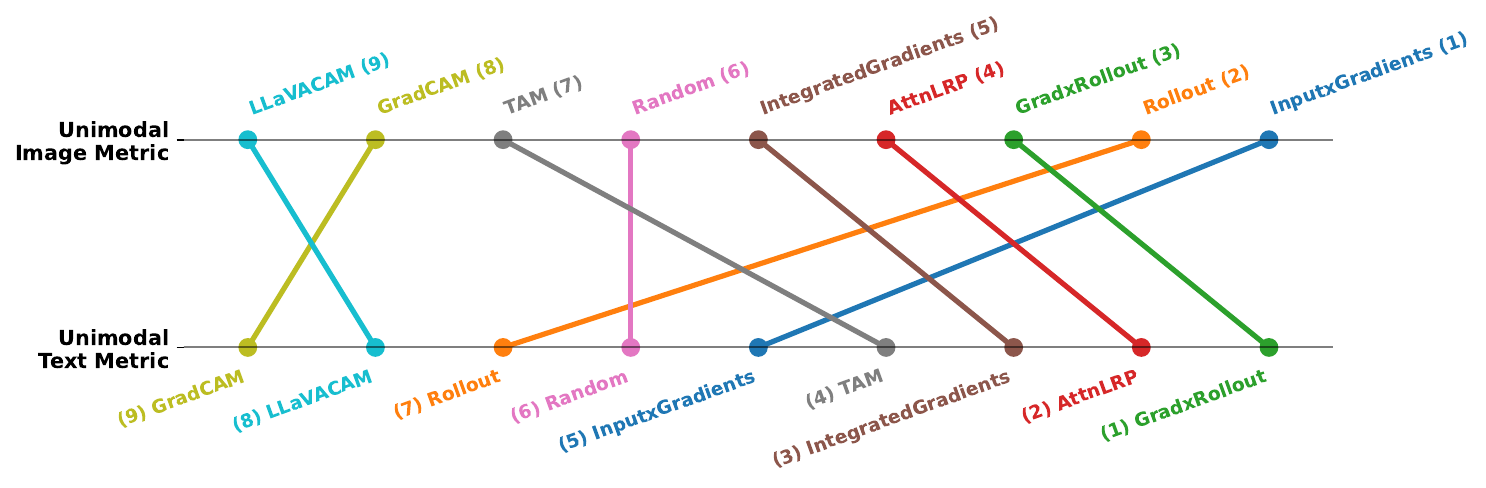}
    \centerline{(b) CVBench}
  \end{minipage}\hfill
  \begin{minipage}[b]{0.32\linewidth}
    \centering
    \includegraphics[width=\linewidth]{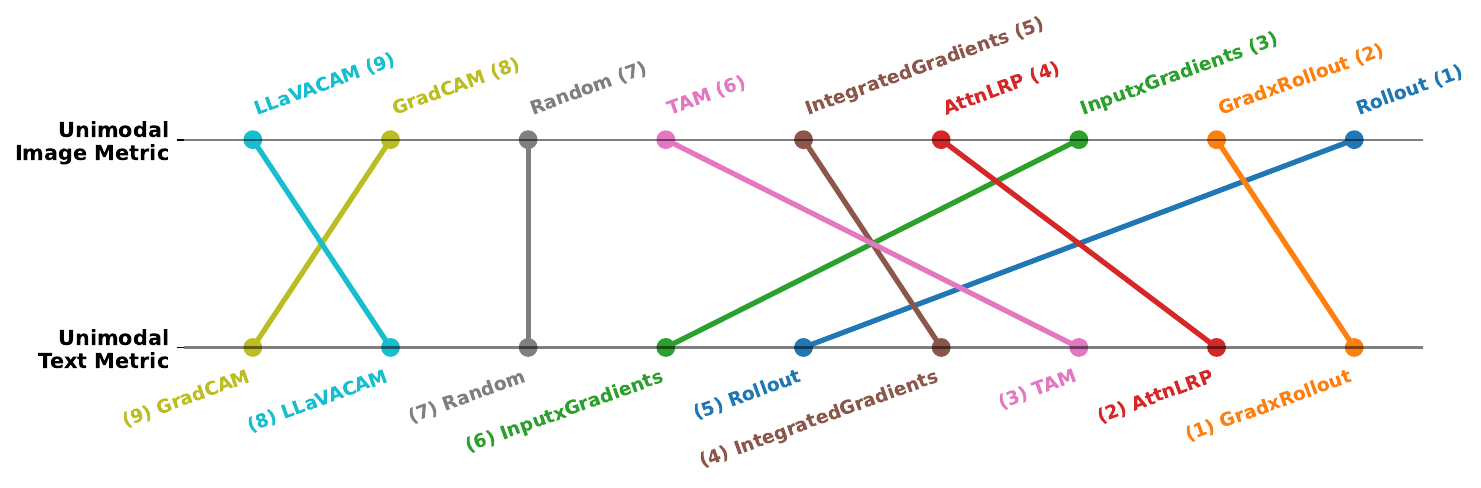}
    \centerline{(c) MMStar}
  \end{minipage}
  \caption{Unimodal rank instability across datasets. The lines track how each explainer's ranking shifts when evaluated by visual faithfulness ($\mu_I$) versus textual faithfulness ($\mu_T$). The severe rank crossing visually demonstrates the evaluation collapse triggered by cross-modal redundancy.}
  \label{fig:supp_rank_shifts}
\end{figure}

As demonstrated in Section \ref{sec:limitations}, unimodal metrics suffer from an evaluation collapse. We complete the empirical experiments and analysis presented in Section \ref{sec:unimodal_rank_results}, by providing the rank shifts per datasets of the evaluated explainers between the visual faithfulness rankings ($\mu_{I}^{srg}$) and the textual faithfulness rankings ($\mu_{T}^{srg}$) in Figure \ref{fig:supp_rank_shifts}. 

% Globally across all evaluated instances, the rank correlation between modalities is statistically indistinguishable from random noise ($\tau = 0.167$, $p = 0.612$). When decomposed by dataset, the Kendall's $\tau$ coefficients demonstrate catastrophic rank crossing on highly redundant datasets like RePOPE ($\tau = 0.222$, $p = 0.477$). On this dataset, explainers scoring in the top-3 for visual fidelity routinely fall to the bottom quartile for textual fidelity. CVBench shows slight improvement but remains statistically insignificant ($\tau = 0.389$, $p = 0.180$). 

% Conversely, on MMStar, which explicitly filters out unimodal textual shortcuts and demands strict joint reasoning, the rankings begin to stabilize and exhibit a moderate correlation ($\tau = 0.500$, $p = 0.075$). This task-dependent fluctuation visually and mathematically confirms our core hypothesis: unimodal metric reliability is a fragile artifact of dataset redundancy rather than a reflection of true XAI capability.

\subsection{Dataset-Specific Unimodal Rank Correlations}
\label{app:dataset_breakdown}

Figure \ref{fig:rank_instability} in the main text illustrates the macro-level instability of unimodal metrics by presenting the global average correlation across all evaluated datasets. To provide a granular view, we break down this correlation by dataset in Table \ref{tab:dataset_breakdown}.

\begin{table}[h]
\centering
\caption{Kendall rank correlation ($\tau$) between image-only and text-only evaluation metrics across individual datasets. The lack of statistical significance ($p > 0.05$) highlights that unimodal metrics fail to produce a consistent explainer hierarchy.}
\label{tab:dataset_breakdown}
\begin{tabular}{lcc}
\toprule
\textbf{Dataset} & \textbf{Kendall Correlation ($\tau$)} & \textbf{$p$-value} \\
\midrule
RePOPE & 0.222 & 0.477 \\
CV-Bench & 0.389 & 0.180 \\
MMStar & 0.500 & 0.075 \\
\midrule
\textbf{Global} & \textbf{0.167} & \textbf{0.612} \\
\bottomrule
\end{tabular}
\end{table}

The results confirm that across every individual dataset, as well as globally, the correlation between rankings derived from isolated textual and visual metrics lacks statistical significance. Evaluating the same explainer yields entirely different hierarchies depending on which modality is isolated, necessitating the joint evaluation provided by $\mathcal{F}_{syn}$.

\takeaway{The granular statistical breakdown confirms that across all individual datasets, the correlation between isolated visual and textual explainer rankings entirely lacks statistical significance ($p > 0.05$), thus necessitating the joint synergistic evaluation provided by $\mathcal{F}_{syn}$.}

\section{Comprehensive Benchmark Results}
\label{appendix_benchmark}
This appendix details the complete empirical results of our benchmark, expanding on the core findings with granular dataset-level breakdowns, cross-architecture metric distributions, and computational efficiency trade-offs.

\subsection{Average Explainer Rankings per Dataset and Metric}
\label{app:average_rankings}

To ensure that the global ordinal superiority of attention-based methods is not disproportionately driven by a single dataset, and to further visualize the ranking instability of unimodal metrics, we decompose the average explainer rankings across our three evaluation benchmarks. 

\begin{figure}[htbp]
  \centering
  % --- ROW 1: F_syn (Synergistic) ---
  \begin{minipage}[b]{0.32\linewidth}
    \centering
    \includegraphics[width=\linewidth]{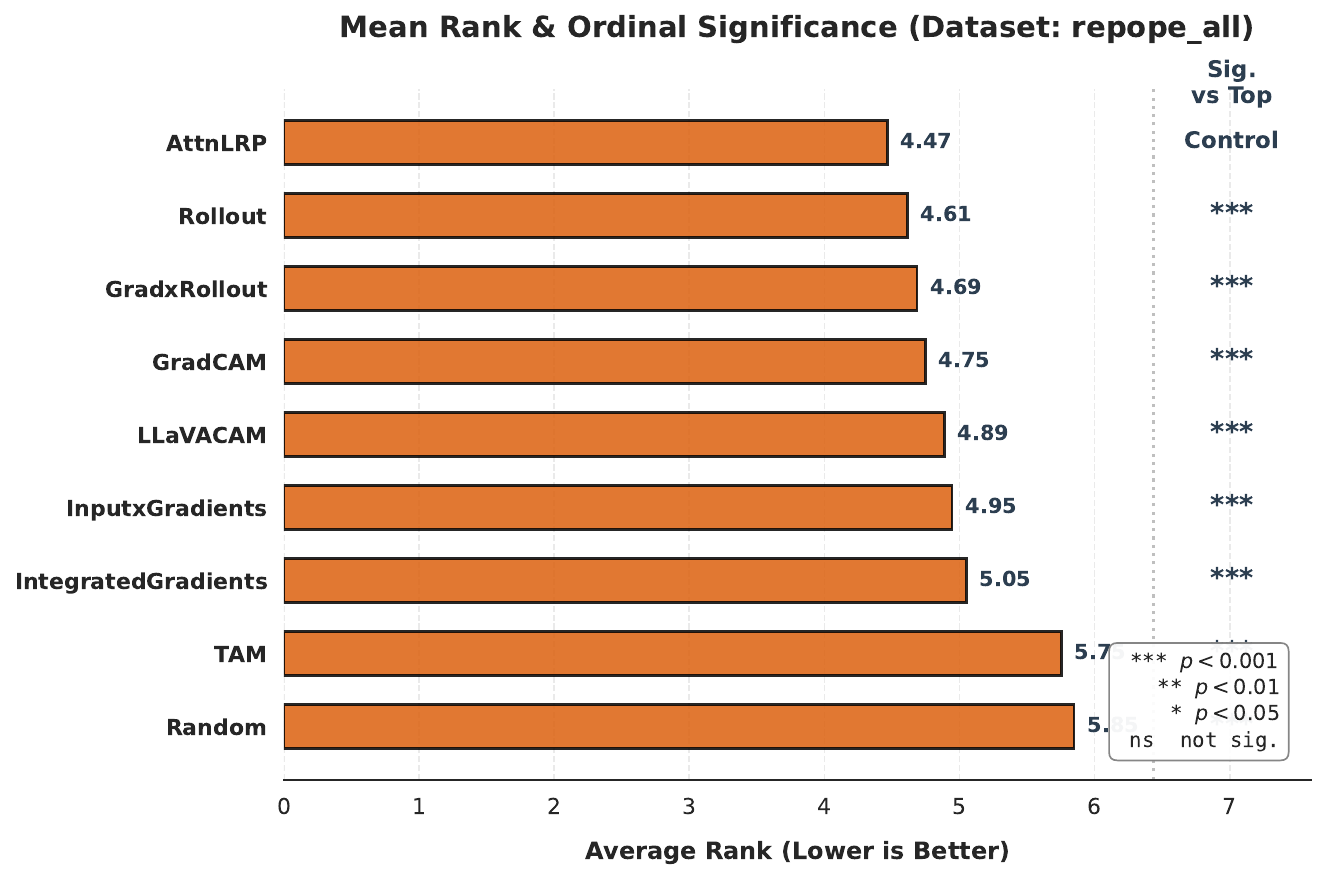} 
    \centerline{(a) $\mathcal{F}_{syn}$ - RePOPE}
  \end{minipage}\hfill
  \begin{minipage}[b]{0.32\linewidth}
    \centering
    \includegraphics[width=\linewidth]{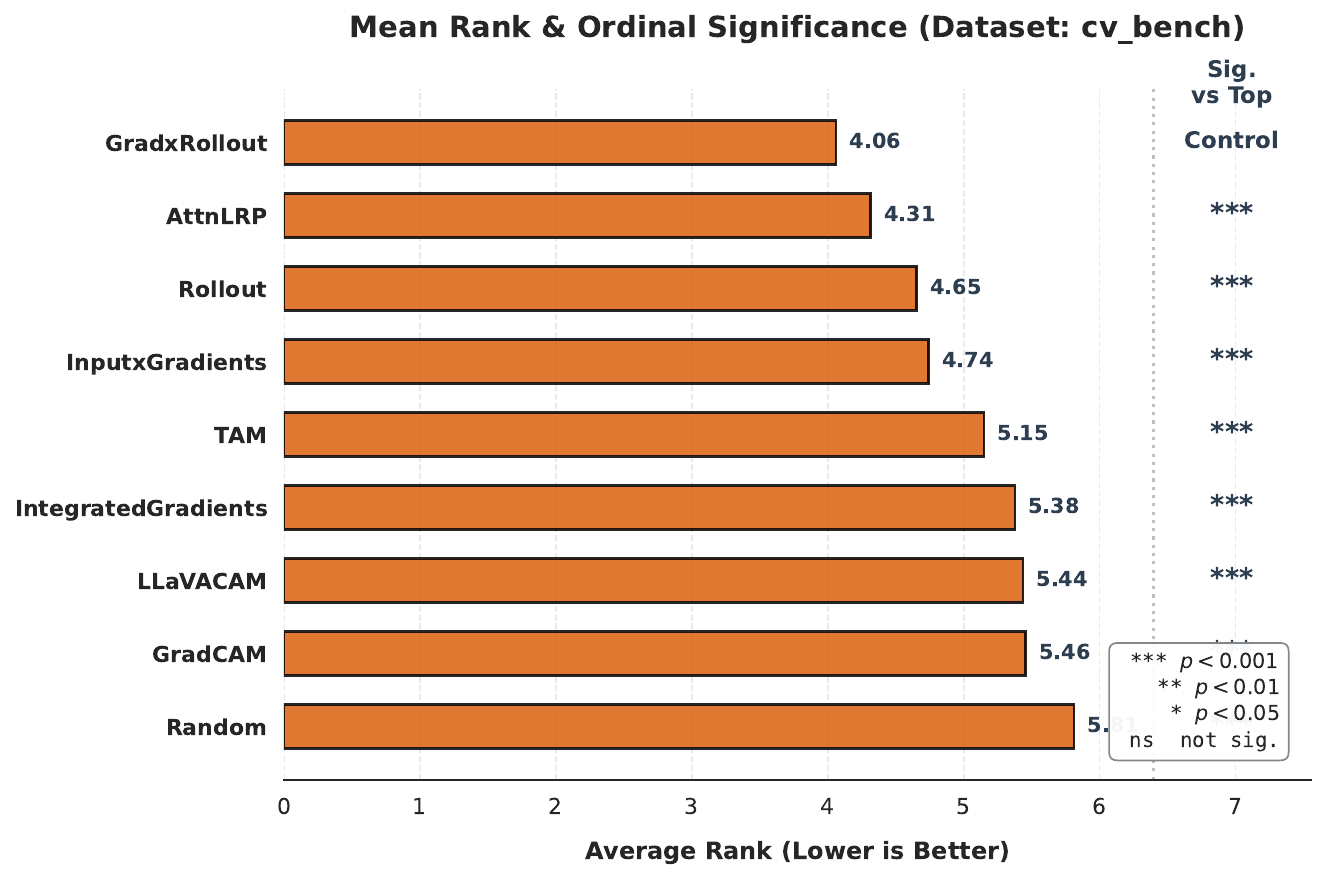} 
    \centerline{(b) $\mathcal{F}_{syn}$ - CVBench}
  \end{minipage}\hfill
  \begin{minipage}[b]{0.32\linewidth}
    \centering
    \includegraphics[width=\linewidth]{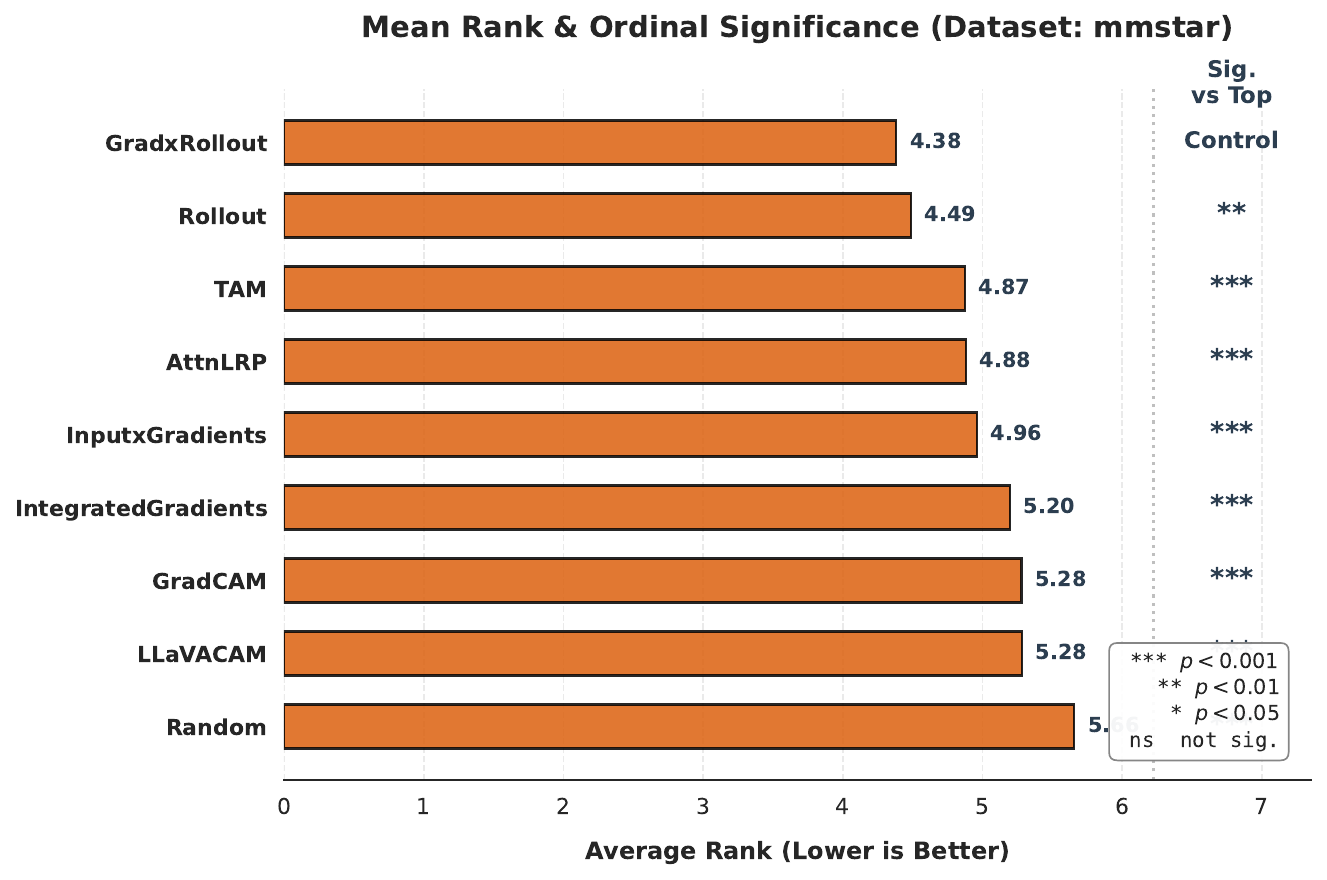} 
    \centerline{(c) $\mathcal{F}_{syn}$ - MMStar}
  \end{minipage}

  \vspace{0.4cm} % Vertical space between rows

  % --- ROW 2: Image SRG (Unimodal Visual) ---
  \begin{minipage}[b]{0.32\linewidth}
    \centering
    \includegraphics[width=\linewidth]{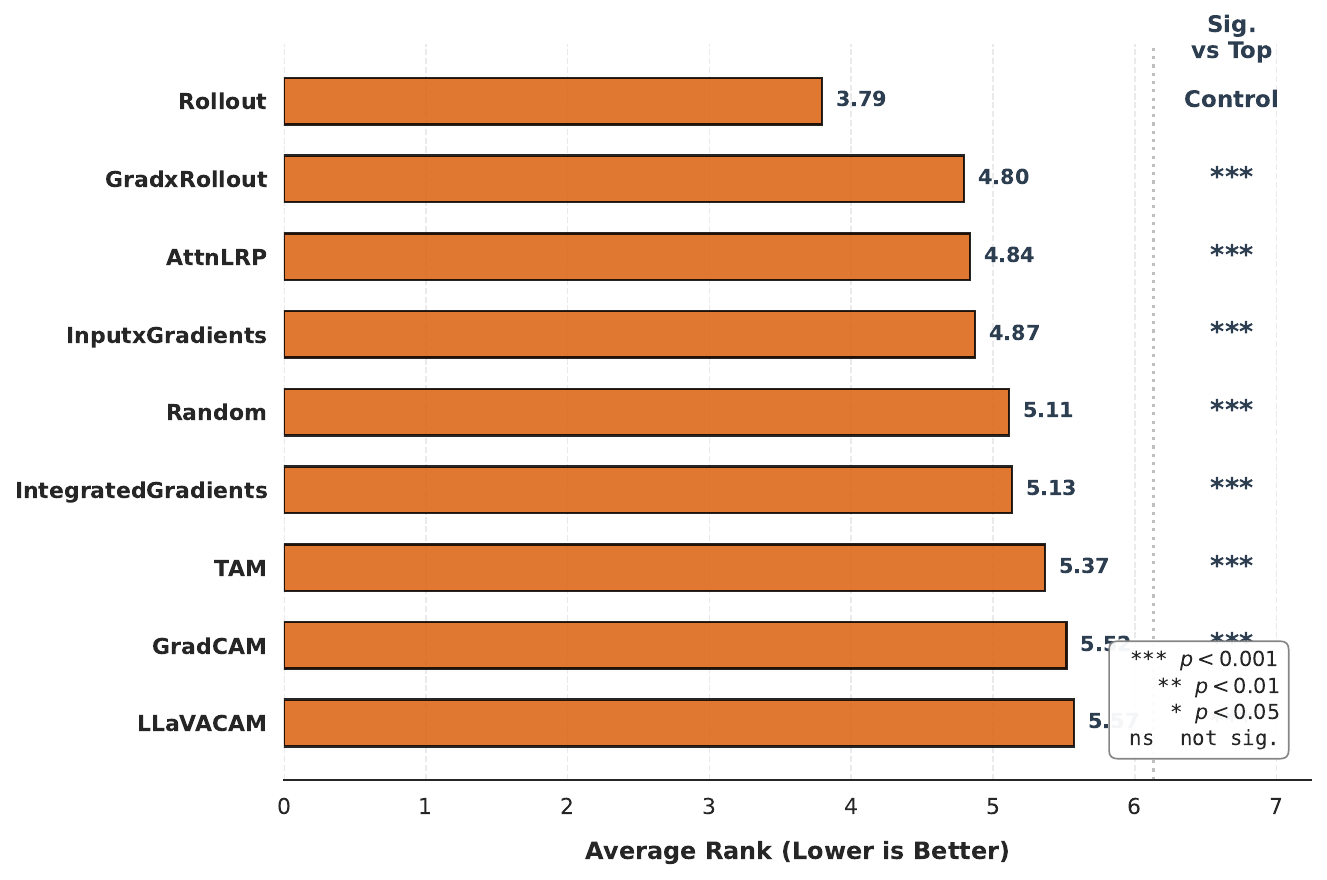} 
    \centerline{(d) $\mu_I^{srg}$ - RePOPE}
  \end{minipage}\hfill
  \begin{minipage}[b]{0.32\linewidth}
    \centering
    \includegraphics[width=\linewidth]{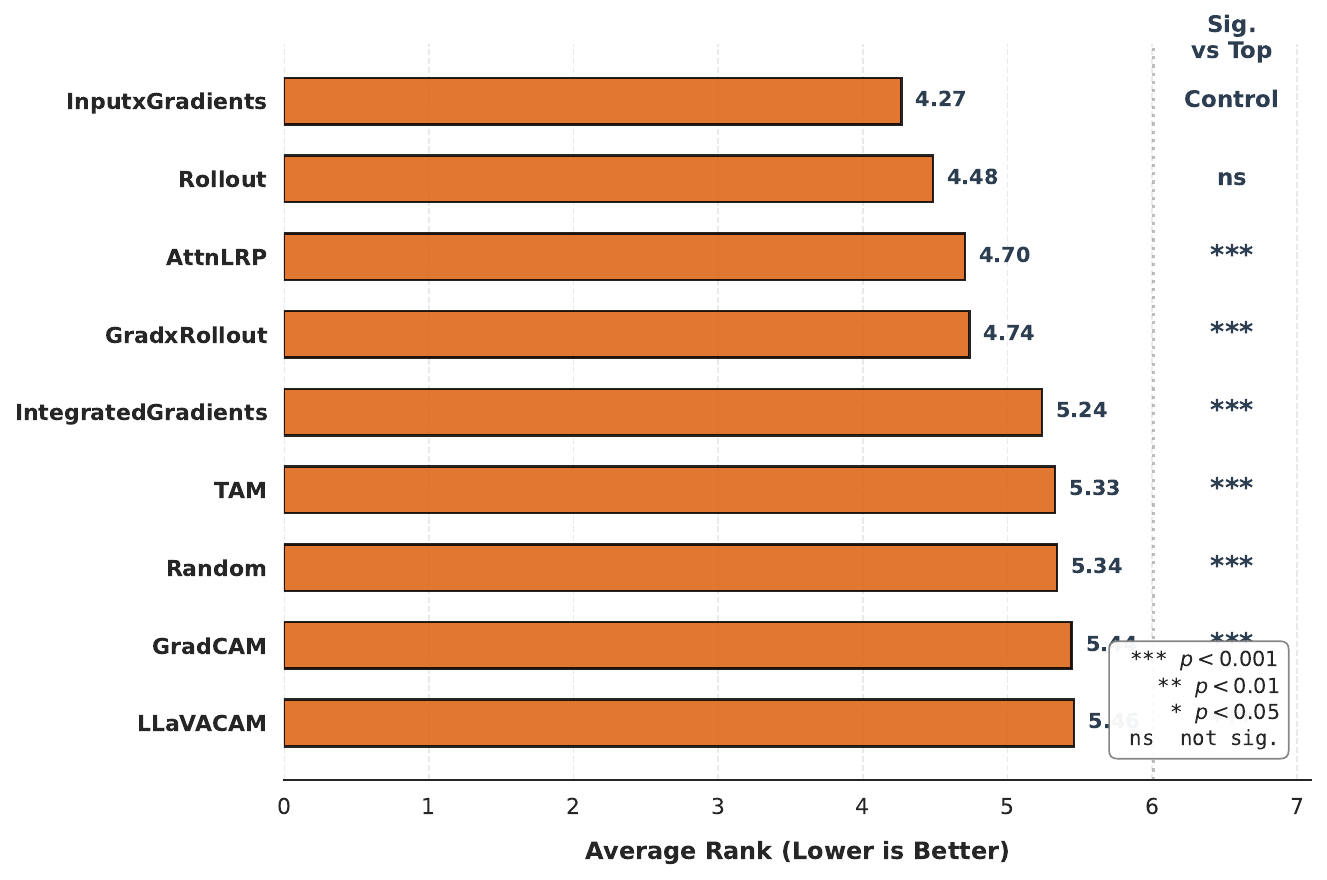} 
    \centerline{(e) $\mu_I^{srg}$ - CVBench}
  \end{minipage}\hfill
  \begin{minipage}[b]{0.32\linewidth}
    \centering
    \includegraphics[width=\linewidth]{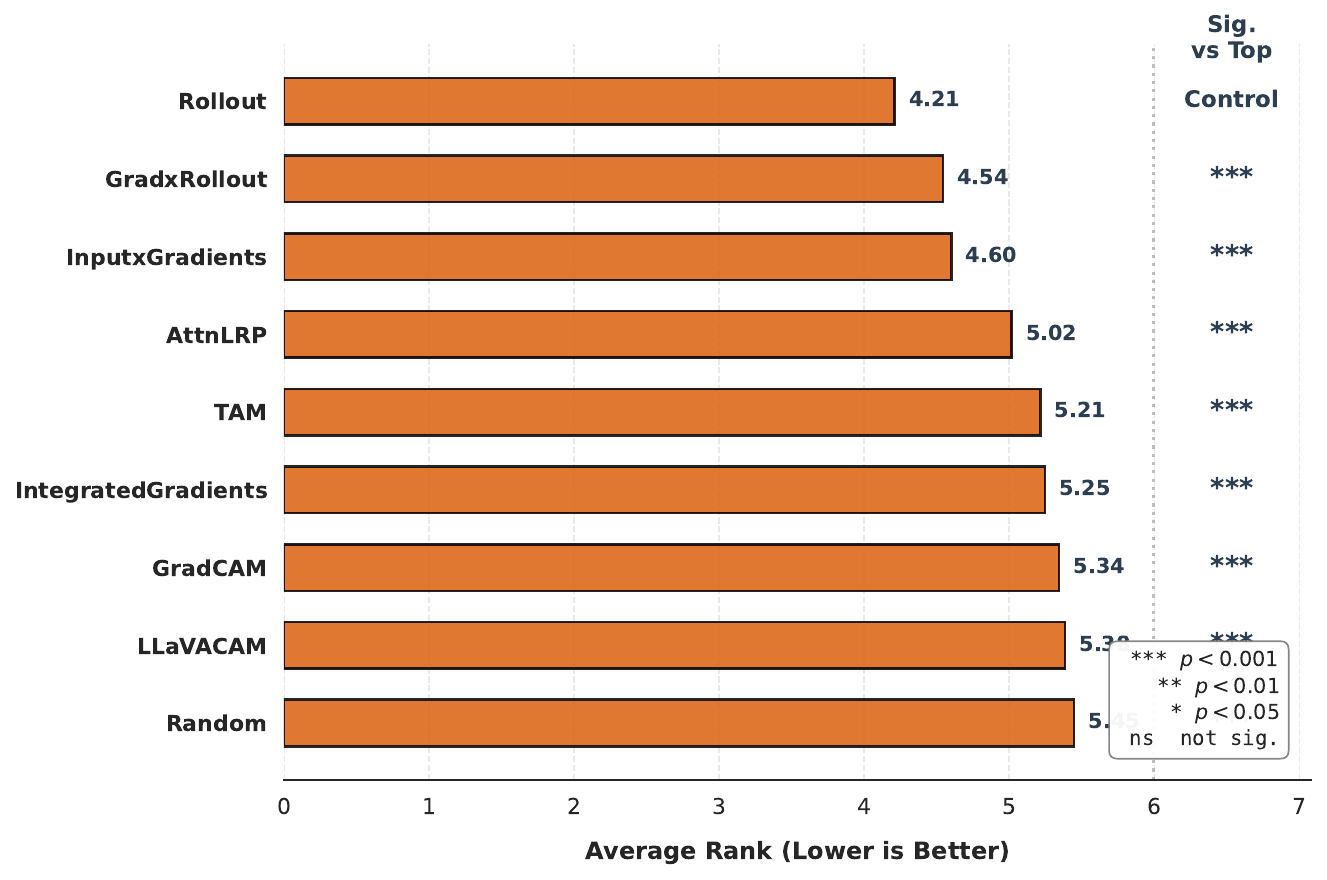} 
    \centerline{(f) $\mu_I^{srg}$ - MMStar}
  \end{minipage}

  \vspace{0.4cm} % Vertical space between rows

  % --- ROW 3: Token SRG (Unimodal Textual) ---
  \begin{minipage}[b]{0.32\linewidth}
    \centering
    \includegraphics[width=\linewidth]{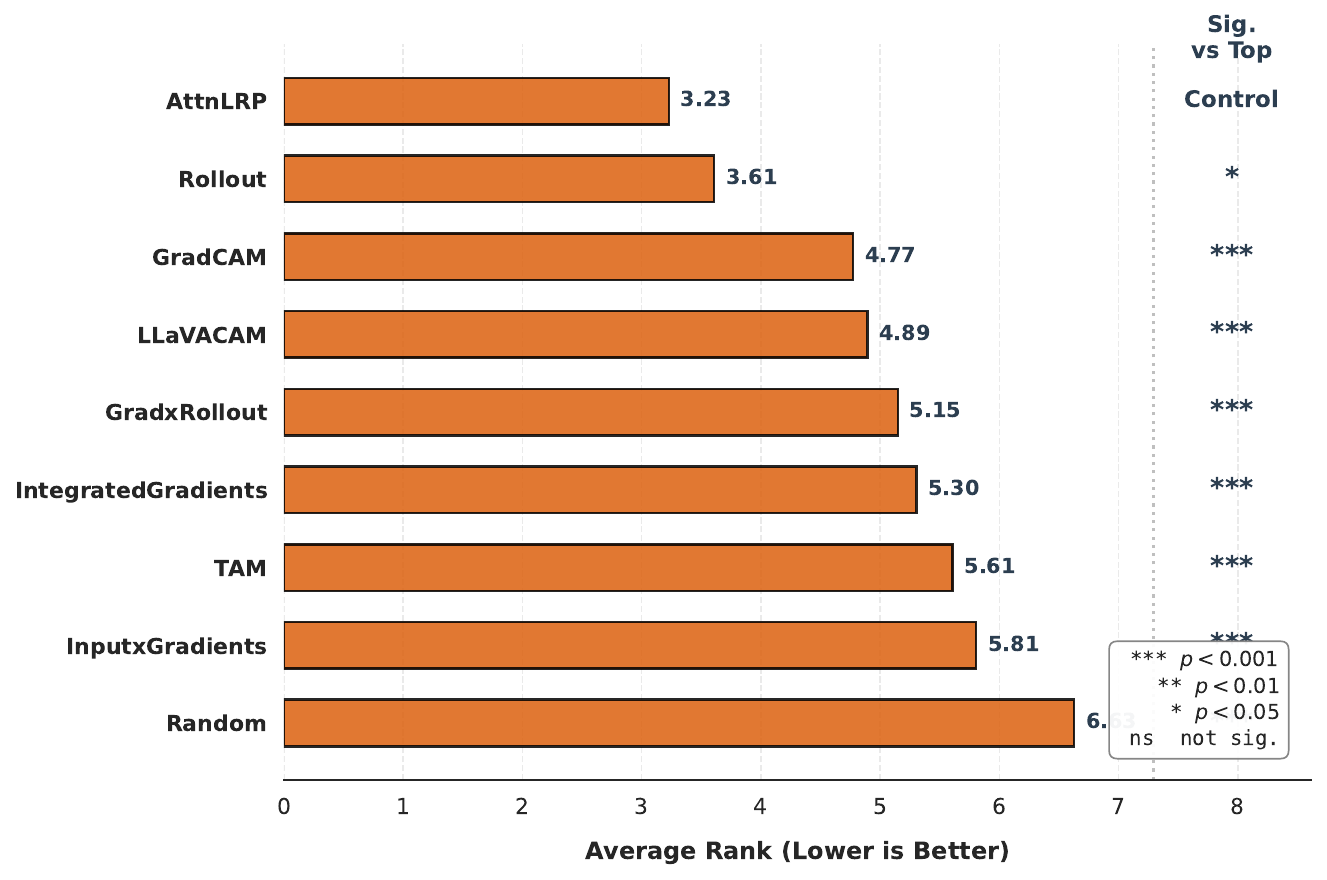} 
    \centerline{(g) $\mu_T^{srg}$ - RePOPE}
  \end{minipage}\hfill
  \begin{minipage}[b]{0.32\linewidth}
    \centering
    \includegraphics[width=\linewidth]{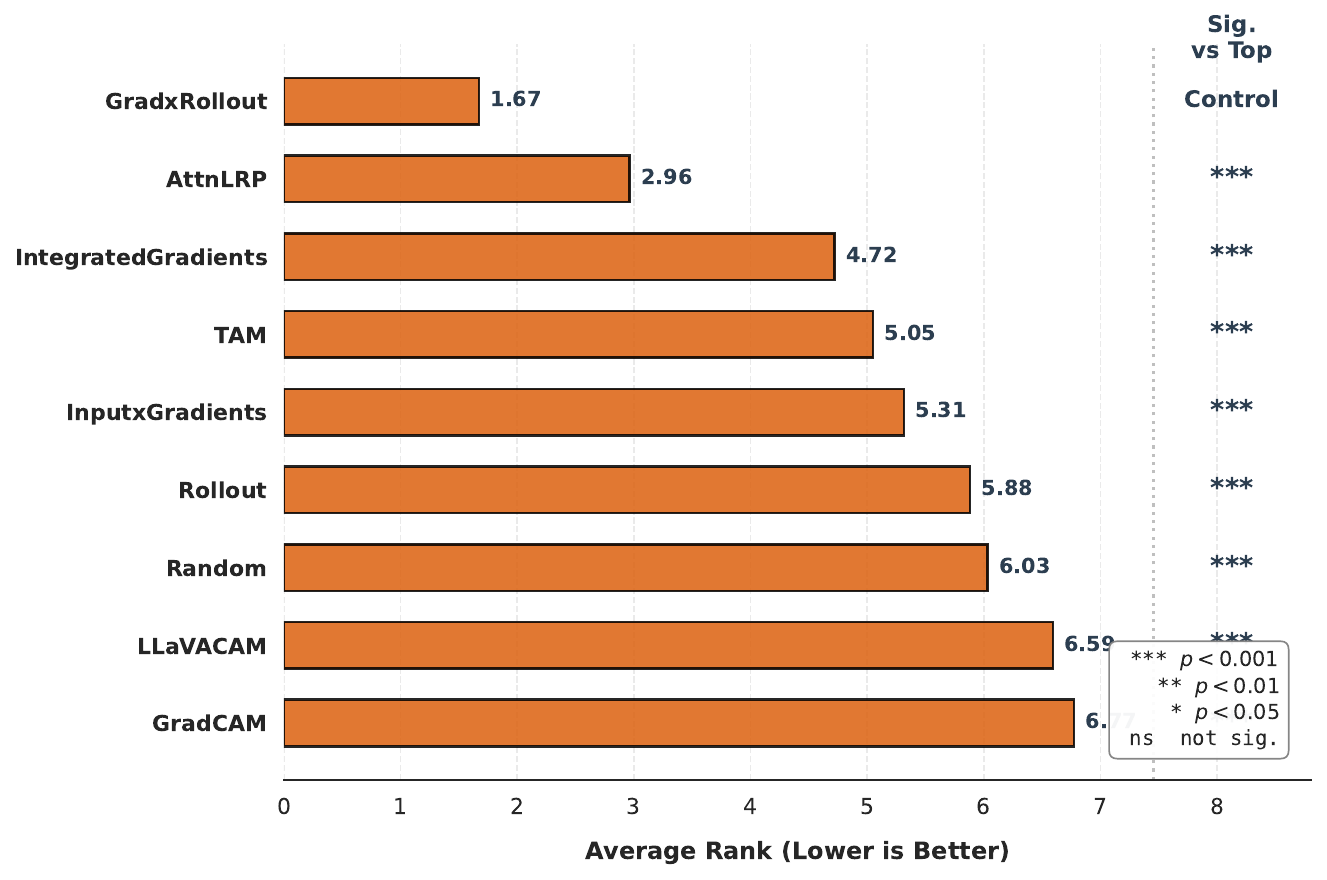} 
    \centerline{(h) $\mu_T^{srg}$ - CVBench}
  \end{minipage}\hfill
  \begin{minipage}[b]{0.32\linewidth}
    \centering
    \includegraphics[width=\linewidth]{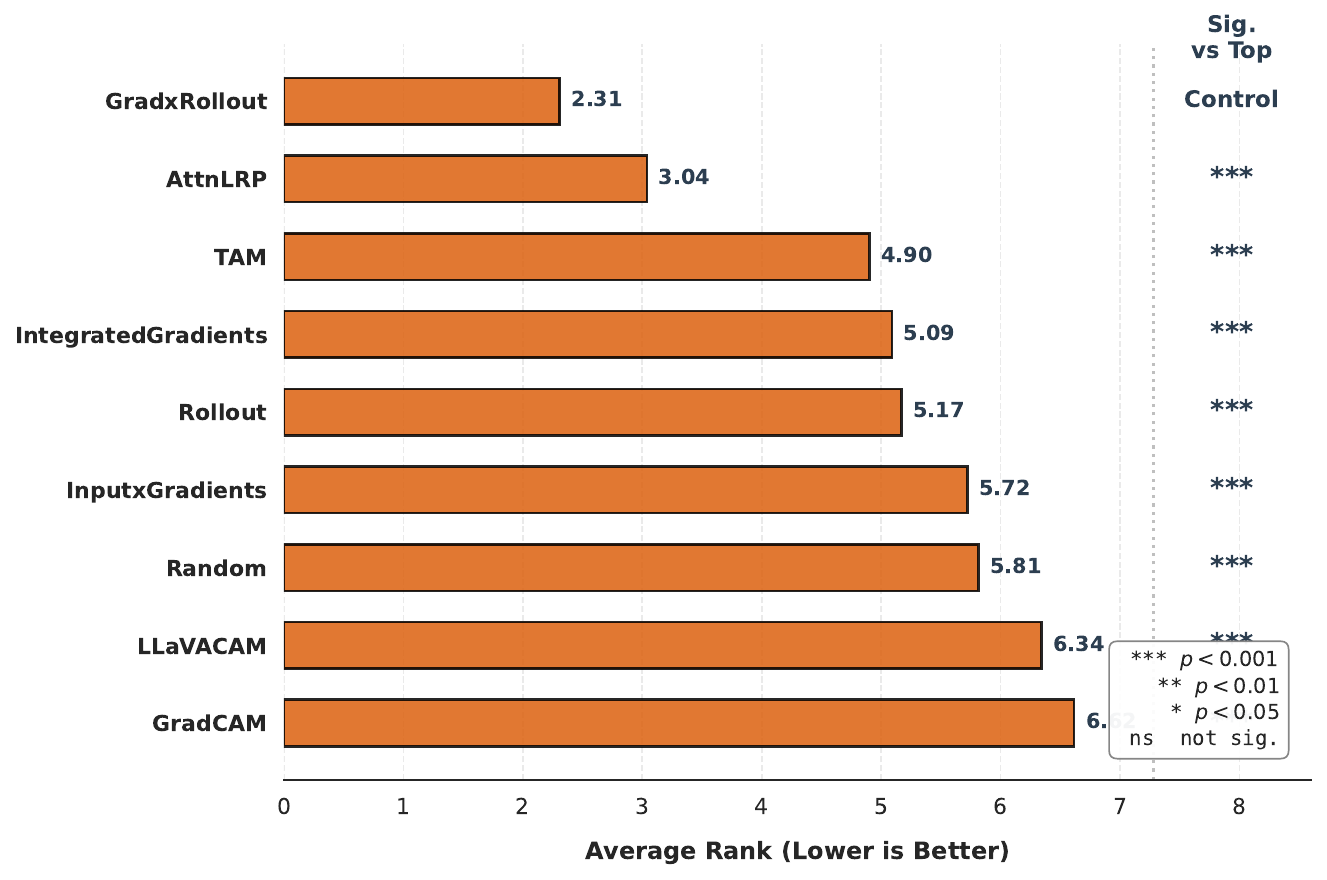} 
    \centerline{(i) $\mu_T^{srg}$ - MMStar}
  \end{minipage}
  
  \caption{Decomposed ordinal rankings of explainers across metrics and datasets. The top row demonstrates that attention-based methods remain robustly superior under true multimodal evaluation ($\mathcal{F}_{syn}$). Conversely, the middle and bottom rows expose how unimodal evaluations ($\mu_I^{srg}$ and $\mu_T^{srg}$) yield erratic, contradictory leaderboards that fluctuate heavily depending on the dataset's specific cross-modal demands.}
  \label{fig:app_3x3_rankings}
\end{figure}

Figure \ref{fig:app_3x3_rankings} breaks down the mean explainer rankings across RePOPE, CVBench, and MMStar, comparing Synergistic Faithfulness ($\mathcal{F}_{syn}$) against isolated visual ($\mu_I^{srg}$) and textual ($\mu_T^{srg}$) faithfulness evaluations.

The top row demonstrates that attention-based methods consistently secure the top rankings under the synergistic faithfulness metric ($\mathcal{F}_{syn}$), independent of the evaluated dataset. Conversely, the unimodal rankings presented in the middle and bottom rows exhibit severe instability. Under isolated $\mu_I^{srg}$ or $\mu_T^{srg}$ faithfulness metrics, specific explainers occasionally underperform even the random baseline explainer. This volatility confirms that unimodal metrics fail to generalize, producing contradictory leaderboards shaped by dataset-specific modal biases rather than true explainer quality.

Consistent with the Linear Mixed-effects Model (LMM) analysis in Section \ref{experiments}, the decomposed rankings in the top row reveal that attention-based methods maintain a robust lead when evaluated for cross-modal synergy, regardless of the dataset. Conversely, the second and third rows empirically validate the structural vulnerabilities of unimodal metrics: VLM-native explainers often artificially inflate their ranks on $\mu_I^{srg}$ by exploiting visual salience, only to collapse when subjected to textual perturbation $\mu_T^{srg}$.

\takeaway{Decomposing ordinal rankings by dataset confirms that while unimodal metrics generate contradictory leaderboards driven by dataset-specific modality biases, $\mathcal{F}_{syn}$ provides a stable, consistent measure of true multimodal groundings.} 

\subsection{Exhaustive Tabular Results}
\label{app:full_tables}

Tables \ref{tab:full_repope_all}, \ref{tab:full_mmstar}, and \ref{tab:full_cv_bench} provide the complete, mean scores and standard deviations for all metric-explainer-model configurations. For each setup, we report the visual metric ($\mu_I^{srg}$), textual metric ($\mu_T^{srg}$), and the unified Synergistic Faithfulness ($\mathcal{F}_{syn}$).

\takeaway{Exhaustive tabular results across all evaluated models and datasets validate the global LMM trends, highlighting both the variance of unimodal metrics and the stability of attention-based methods to identify cross-modal synergy.}

\begin{table}[ht]
\centering
\caption{Comprehensive Benchmark Results on the \textbf{RePOPE} Dataset. Values represent the unadjusted mean scores $\pm$ standard deviation across all dataset instances.}
\label{tab:full_repope_all}
\resizebox{\textwidth}{!}{%
\begin{tabular}{l | ccc | ccc | ccc}
\toprule
& \multicolumn{3}{c|}{\textbf{Qwen2.5-VL}} & \multicolumn{3}{c|}{\textbf{LLaVA-1.5}} & \multicolumn{3}{c}{\textbf{InternVL-3.5 }} \\
\cmidrule(lr){2-4} \cmidrule(lr){5-7} \cmidrule(lr){8-10}
\textbf{Explainer} & $\mu_I^{srg}$ & $\mu_T^{srg}$ & $\mathcal{F}_{syn}$ & $\mu_I^{srg}$ & $\mu_T^{srg}$ & $\mathcal{F}_{syn}$ & $\mu_I^{srg}$ & $\mu_T^{srg}$ & $\mathcal{F}_{syn}$ \\
\midrule
AttnLRP & $-0.001_{ \pm 0.075 }$ & $0.263_{ \pm 0.388 }$ & $0.077_{ \pm 0.131 }$ & $0.013_{ \pm 0.075 }$ & $0.321_{ \pm 0.243 }$ & $0.032_{ \pm 0.050 }$ & $0.033_{ \pm 0.131 }$ & $0.239_{ \pm 0.288 }$ & $0.178_{ \pm 0.163 }$ \\
Grad$\times$Rollout & $-0.003_{ \pm 0.086 }$ & $0.050_{ \pm 0.207 }$ & $0.063_{ \pm 0.117 }$ & $0.110_{ \pm 0.264 }$ & $0.129_{ \pm 0.184 }$ & $0.032_{ \pm 0.049 }$ & $0.018_{ \pm 0.236 }$ & $0.085_{ \pm 0.157 }$ & $0.147_{ \pm 0.146 }$ \\
Rollout & $0.002_{ \pm 0.094 }$ & $0.236_{ \pm 0.102 }$ & $0.058_{ \pm 0.118 }$ & $0.141_{ \pm 0.270 }$ & $0.531_{ \pm 0.232 }$ & $0.036_{ \pm 0.050 }$ & $0.211_{ \pm 0.228 }$ & $0.124_{ \pm 0.123 }$ & $0.177_{ \pm 0.177 }$ \\
\midrule
TAM & $0.002_{ \pm 0.088 }$ & $0.192_{ \pm 0.285 }$ & $0.064_{ \pm 0.117 }$ & $-0.076_{ \pm 0.269 }$ & $-0.454_{ \pm 0.321 }$ & $0.025_{ \pm 0.038 }$ & $-0.033_{ \pm 0.264 }$ & $0.026_{ \pm 0.209 }$ & $0.122_{ \pm 0.133 }$ \\
LLaVA-CAM & $0.007_{ \pm 0.094 }$ & $0.189_{ \pm 0.306 }$ & $0.068_{ \pm 0.108 }$ & $-0.133_{ \pm 0.365 }$ & $0.279_{ \pm 0.370 }$ & $0.029_{ \pm 0.043 }$ & $-0.062_{ \pm 0.256 }$ & $0.021_{ \pm 0.193 }$ & $0.140_{ \pm 0.138 }$ \\
\midrule
GradCAM & $0.002_{ \pm 0.086 }$ & $0.218_{ \pm 0.298 }$ & $0.079_{ \pm 0.121 }$ & $-0.108_{ \pm 0.337 }$ & $0.288_{ \pm 0.345 }$ & $0.029_{ \pm 0.043 }$ & $-0.044_{ \pm 0.247 }$ & $0.043_{ \pm 0.184 }$ & $0.142_{ \pm 0.141 }$ \\
Input$\times$Grad & $0.001_{ \pm 0.074 }$ & $-0.028_{ \pm 0.262 }$ & $0.062_{ \pm 0.104 }$ & $0.013_{ \pm 0.087 }$ & $-0.060_{ \pm 0.674 }$ & $0.028_{ \pm 0.041 }$ & $0.021_{ \pm 0.155 }$ & $0.030_{ \pm 0.244 }$ & $0.158_{ \pm 0.156 }$ \\
Integ. Gradients & $-0.000_{ \pm 0.076 }$ & $0.107_{ \pm 0.381 }$ & $0.071_{ \pm 0.120 }$ & $0.001_{ \pm 0.078 }$ & $-0.641_{ \pm 0.299 }$ & $0.023_{ \pm 0.036 }$ & $-0.013_{ \pm 0.127 }$ & $0.216_{ \pm 0.288 }$ & $0.162_{ \pm 0.157 }$ \\
\midrule
Random (Baseline) & $-0.002_{ \pm 0.073 }$ & $0.024_{ \pm 0.209 }$ & $0.055_{ \pm 0.112 }$ & $0.003_{ \pm 0.060 }$ & $-0.633_{ \pm 0.308 }$ & $0.022_{ \pm 0.036 }$ & $-0.004_{ \pm 0.075 }$ & $0.039_{ \pm 0.283 }$ & $0.126_{ \pm 0.124 }$ \\
\bottomrule
\end{tabular}%
}
\end{table}

\begin{table}[ht]
\centering
\caption{Comprehensive Benchmark Results on the \textbf{MMStar} Dataset. Values represent the unadjusted mean scores $\pm$ standard deviation across all dataset instances.}
\label{tab:full_mmstar}
\resizebox{\textwidth}{!}{%
\begin{tabular}{l | ccc | ccc | ccc}
\toprule
& \multicolumn{3}{c|}{\textbf{Qwen2.5-VL}} & \multicolumn{3}{c|}{\textbf{LLaVA-1.5}} & \multicolumn{3}{c}{\textbf{InternVL-3.5 }} \\
\cmidrule(lr){2-4} \cmidrule(lr){5-7} \cmidrule(lr){8-10}
\textbf{Explainer} & $\mu_I^{srg}$ & $\mu_T^{srg}$ & $\mathcal{F}_{syn}$ & $\mu_I^{srg}$ & $\mu_T^{srg}$ & $\mathcal{F}_{syn}$ & $\mu_I^{srg}$ & $\mu_T^{srg}$ & $\mathcal{F}_{syn}$ \\
\midrule
AttnLRP & $-0.002_{ \pm 0.096 }$ & $0.157_{ \pm 0.331 }$ & $0.079_{ \pm 0.098 }$ & $0.020_{ \pm 0.089 }$ & $0.372_{ \pm 0.265 }$ & $0.057_{ \pm 0.079 }$ & $0.068_{ \pm 0.137 }$ & $0.308_{ \pm 0.267 }$ & $0.246_{ \pm 0.158 }$ \\
Grad$\times$Rollout & $0.005_{ \pm 0.106 }$ & $0.390_{ \pm 0.181 }$ & $0.068_{ \pm 0.097 }$ & $0.106_{ \pm 0.277 }$ & $0.333_{ \pm 0.198 }$ & $0.066_{ \pm 0.094 }$ & $0.159_{ \pm 0.279 }$ & $0.347_{ \pm 0.184 }$ & $0.276_{ \pm 0.177 }$ \\
Rollout & $-0.002_{ \pm 0.102 }$ & $0.045_{ \pm 0.186 }$ & $0.078_{ \pm 0.097 }$ & $0.150_{ \pm 0.284 }$ & $0.118_{ \pm 0.282 }$ & $0.072_{ \pm 0.092 }$ & $0.212_{ \pm 0.287 }$ & $0.046_{ \pm 0.196 }$ & $0.249_{ \pm 0.168 }$ \\
\midrule
TAM & $0.000_{ \pm 0.102 }$ & $0.224_{ \pm 0.183 }$ & $0.068_{ \pm 0.088 }$ & $-0.051_{ \pm 0.296 }$ & $0.031_{ \pm 0.230 }$ & $0.048_{ \pm 0.065 }$ & $0.067_{ \pm 0.250 }$ & $0.042_{ \pm 0.234 }$ & $0.221_{ \pm 0.150 }$ \\
LLaVA-CAM & $-0.003_{ \pm 0.101 }$ & $-0.094_{ \pm 0.265 }$ & $0.067_{ \pm 0.086 }$ & $0.026_{ \pm 0.307 }$ & $-0.040_{ \pm 0.275 }$ & $0.056_{ \pm 0.075 }$ & $-0.038_{ \pm 0.249 }$ & $-0.031_{ \pm 0.212 }$ & $0.183_{ \pm 0.131 }$ \\
\midrule
GradCAM & $-0.005_{ \pm 0.107 }$ & $-0.151_{ \pm 0.255 }$ & $0.070_{ \pm 0.090 }$ & $0.010_{ \pm 0.303 }$ & $-0.044_{ \pm 0.274 }$ & $0.055_{ \pm 0.071 }$ & $-0.016_{ \pm 0.236 }$ & $-0.057_{ \pm 0.222 }$ & $0.183_{ \pm 0.128 }$ \\
Input$\times$Grad & $-0.001_{ \pm 0.095 }$ & $0.093_{ \pm 0.215 }$ & $0.065_{ \pm 0.079 }$ & $0.052_{ \pm 0.109 }$ & $0.003_{ \pm 0.206 }$ & $0.051_{ \pm 0.068 }$ & $0.127_{ \pm 0.177 }$ & $-0.036_{ \pm 0.191 }$ & $0.219_{ \pm 0.141 }$ \\
Integ. Gradients & $0.003_{ \pm 0.100 }$ & $0.070_{ \pm 0.290 }$ & $0.072_{ \pm 0.085 }$ & $0.011_{ \pm 0.088 }$ & $0.042_{ \pm 0.206 }$ & $0.049_{ \pm 0.067 }$ & $0.022_{ \pm 0.124 }$ & $0.103_{ \pm 0.214 }$ & $0.204_{ \pm 0.136 }$ \\
\midrule
Random (Baseline) & $0.003_{ \pm 0.094 }$ & $0.029_{ \pm 0.207 }$ & $0.068_{ \pm 0.083 }$ & $0.003_{ \pm 0.082 }$ & $0.019_{ \pm 0.169 }$ & $0.046_{ \pm 0.063 }$ & $-0.005_{ \pm 0.084 }$ & $0.003_{ \pm 0.187 }$ & $0.172_{ \pm 0.119 }$ \\
\bottomrule
\end{tabular}%
}
\end{table}

\begin{table}[ht]
\centering
\caption{Comprehensive Benchmark Results on the \textbf{CVBench} Dataset. Values represent the unadjusted mean scores $\pm$ standard deviation across all dataset instances.}
\label{tab:full_cv_bench}
\resizebox{\textwidth}{!}{%
\begin{tabular}{l | ccc | ccc | ccc}
\toprule
& \multicolumn{3}{c|}{\textbf{Qwen2.5-VL}} & \multicolumn{3}{c|}{\textbf{LLaVA-1.5}} & \multicolumn{3}{c}{\textbf{InternVL-3.5 }} \\
\cmidrule(lr){2-4} \cmidrule(lr){5-7} \cmidrule(lr){8-10}
\textbf{Explainer} & $\mu_I^{srg}$ & $\mu_T^{srg}$ & $\mathcal{F}_{syn}$ & $\mu_I^{srg}$ & $\mu_T^{srg}$ & $\mathcal{F}_{syn}$ & $\mu_I^{srg}$ & $\mu_T^{srg}$ & $\mathcal{F}_{syn}$ \\
\midrule
AttnLRP & $-0.001_{ \pm 0.088 }$ & $0.440_{ \pm 0.297 }$ & $0.049_{ \pm 0.056 }$ & $0.022_{ \pm 0.104 }$ & $0.366_{ \pm 0.310 }$ & $0.080_{ \pm 0.106 }$ & $0.133_{ \pm 0.189 }$ & $0.219_{ \pm 0.206 }$ & $0.167_{ \pm 0.127 }$ \\
Grad$\times$Rollout & $-0.001_{ \pm 0.101 }$ & $0.610_{ \pm 0.129 }$ & $0.035_{ \pm 0.056 }$ & $0.032_{ \pm 0.274 }$ & $0.511_{ \pm 0.189 }$ & $0.088_{ \pm 0.108 }$ & $0.161_{ \pm 0.267 }$ & $0.353_{ \pm 0.146 }$ & $0.206_{ \pm 0.136 }$ \\
Rollout & $-0.001_{ \pm 0.104 }$ & $-0.049_{ \pm 0.186 }$ & $0.039_{ \pm 0.053 }$ & $0.065_{ \pm 0.238 }$ & $-0.224_{ \pm 0.237 }$ & $0.079_{ \pm 0.109 }$ & $0.158_{ \pm 0.229 }$ & $0.077_{ \pm 0.174 }$ & $0.189_{ \pm 0.141 }$ \\
\midrule
TAM & $0.003_{ \pm 0.099 }$ & $0.013_{ \pm 0.154 }$ & $0.040_{ \pm 0.050 }$ & $0.022_{ \pm 0.260 }$ & $0.178_{ \pm 0.221 }$ & $0.071_{ \pm 0.091 }$ & $-0.054_{ \pm 0.244 }$ & $0.041_{ \pm 0.151 }$ & $0.133_{ \pm 0.099 }$ \\
LLaVA-CAM & $0.008_{ \pm 0.100 }$ & $-0.278_{ \pm 0.252 }$ & $0.036_{ \pm 0.046 }$ & $-0.026_{ \pm 0.280 }$ & $-0.045_{ \pm 0.342 }$ & $0.059_{ \pm 0.075 }$ & $-0.048_{ \pm 0.247 }$ & $-0.007_{ \pm 0.175 }$ & $0.126_{ \pm 0.095 }$ \\
\midrule
GradCAM & $0.002_{ \pm 0.103 }$ & $-0.306_{ \pm 0.235 }$ & $0.036_{ \pm 0.045 }$ & $-0.018_{ \pm 0.260 }$ & $-0.087_{ \pm 0.364 }$ & $0.061_{ \pm 0.076 }$ & $-0.032_{ \pm 0.244 }$ & $-0.009_{ \pm 0.172 }$ & $0.125_{ \pm 0.096 }$ \\
Input$\times$Grad & $0.004_{ \pm 0.089 }$ & $0.204_{ \pm 0.224 }$ & $0.038_{ \pm 0.050 }$ & $0.069_{ \pm 0.135 }$ & $-0.013_{ \pm 0.198 }$ & $0.070_{ \pm 0.087 }$ & $0.171_{ \pm 0.198 }$ & $-0.014_{ \pm 0.172 }$ & $0.166_{ \pm 0.113 }$ \\
Integ. Gradients & $0.003_{ \pm 0.090 }$ & $0.220_{ \pm 0.329 }$ & $0.041_{ \pm 0.044 }$ & $0.012_{ \pm 0.107 }$ & $0.019_{ \pm 0.218 }$ & $0.067_{ \pm 0.082 }$ & $0.012_{ \pm 0.152 }$ & $0.112_{ \pm 0.177 }$ & $0.143_{ \pm 0.105 }$ \\
\midrule
Random (Baseline) & $0.002_{ \pm 0.093 }$ & $0.014_{ \pm 0.226 }$ & $0.034_{ \pm 0.047 }$ & $0.002_{ \pm 0.087 }$ & $-0.003_{ \pm 0.197 }$ & $0.060_{ \pm 0.075 }$ & $0.003_{ \pm 0.082 }$ & $-0.044_{ \pm 0.118 }$ & $0.113_{ \pm 0.085 }$ \\
\bottomrule
\end{tabular}%
}
\end{table}

\subsection{Visualization of faithfulness metrics distributions}
\label{app:boxplots}

\begin{figure}[htbp]
  \centering
  \begin{minipage}[b]{\linewidth}
    \centering
    \includegraphics[width=.9\linewidth]{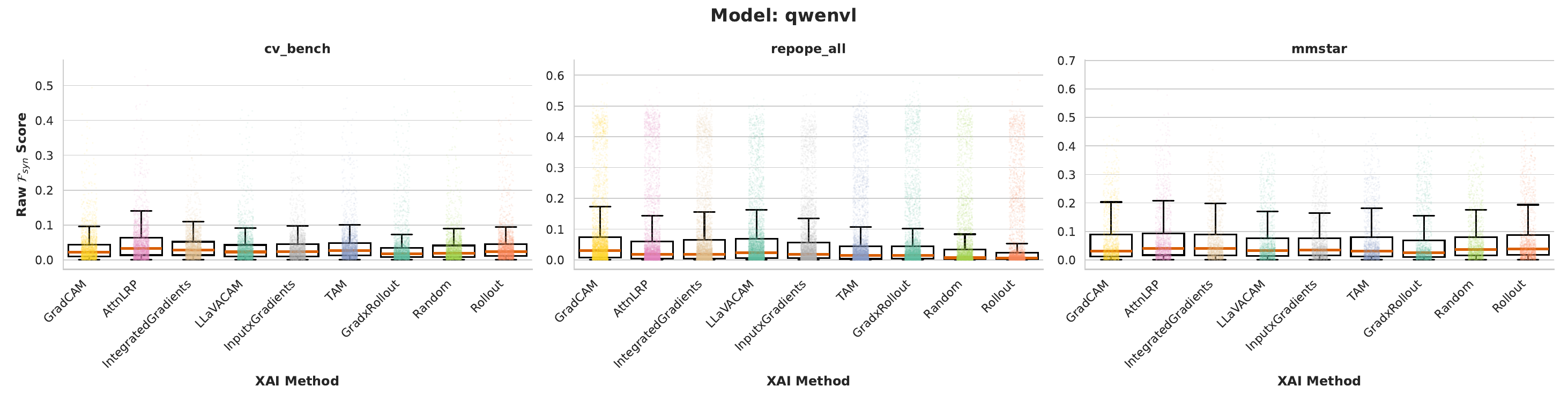}
    \centerline{(a) Qwen2.5-VL}
  \end{minipage}
  
  \vspace{1em}
  
  \begin{minipage}[b]{\linewidth}
    \centering
    \includegraphics[width=.9\linewidth]{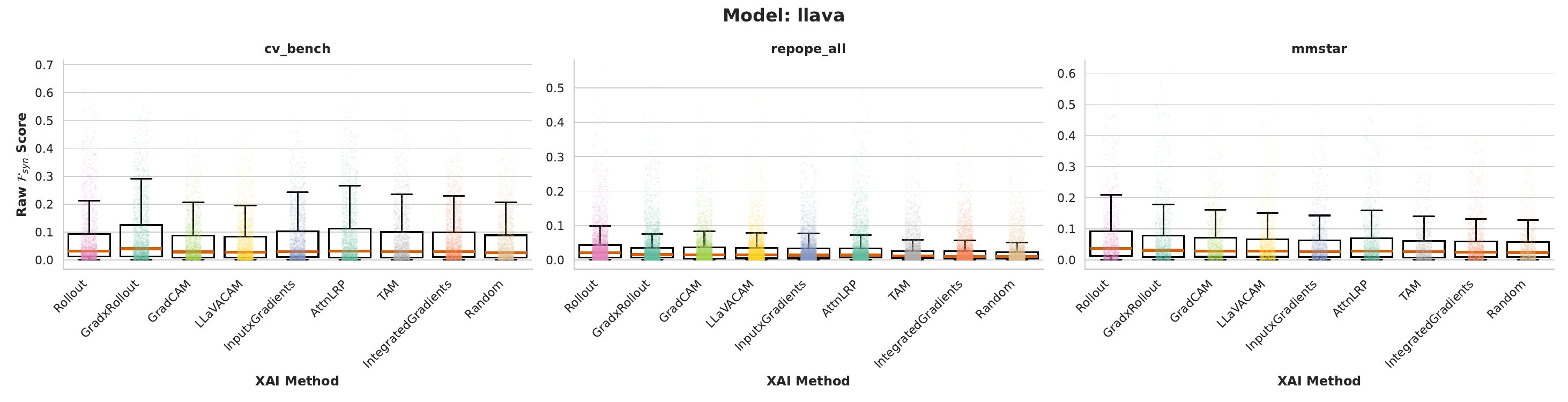}
    \centerline{(b) LLaVA-1.5}
  \end{minipage}
  
  \vspace{1em}
  
  \begin{minipage}[b]{\linewidth}
    \centering
    \includegraphics[width=.9\linewidth]{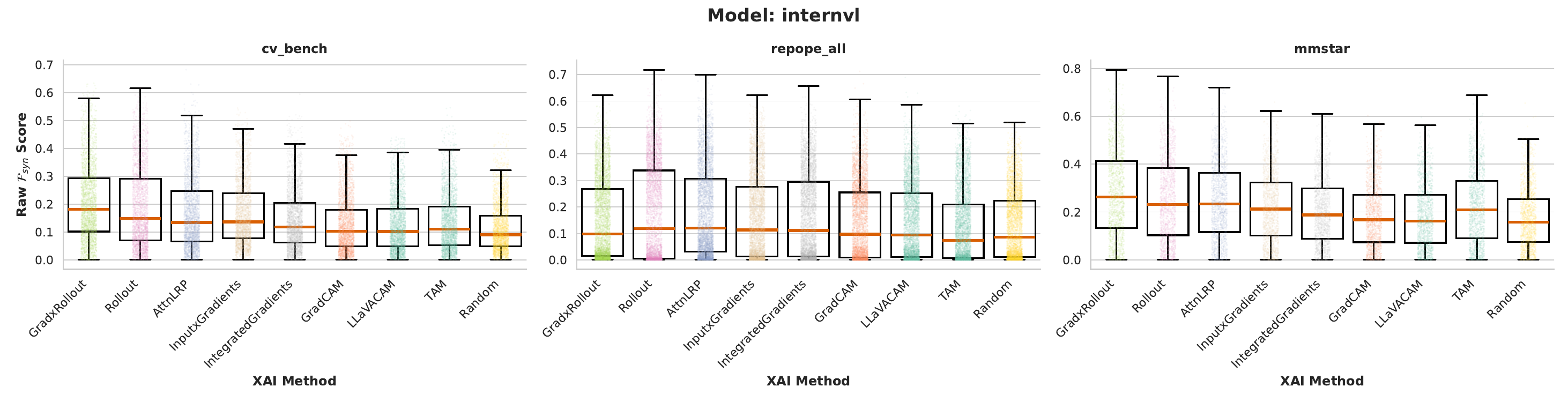}
    \centerline{(c) InternVL-3.5}
  \end{minipage}
  \caption{Raw Synergistic Faithfulness ($\mathcal{F}_{syn}$) score distributions per explainer, evaluated across individual datasets and broken down by VLM architecture: (a) Qwen2.5-VL, (b) LLaVA-1.5, and (c) InternVL-3.5. Low-opacity scatter points visualize the dense instance-level variance, while high-contrast overlaid boxplots denote the median and interquartile ranges. While the absolute magnitudes of the scores naturally fluctuate based on model capacity and dataset-specific difficulty, the relative ordinal hierarchy of the explainers remains robustly consistent across all architectures.}
  \label{fig:app_boxplots_syn}
\end{figure}

\begin{figure}[htbp]
  \centering
  \begin{minipage}[b]{\linewidth}
    \centering
    \includegraphics[width=.9\linewidth]{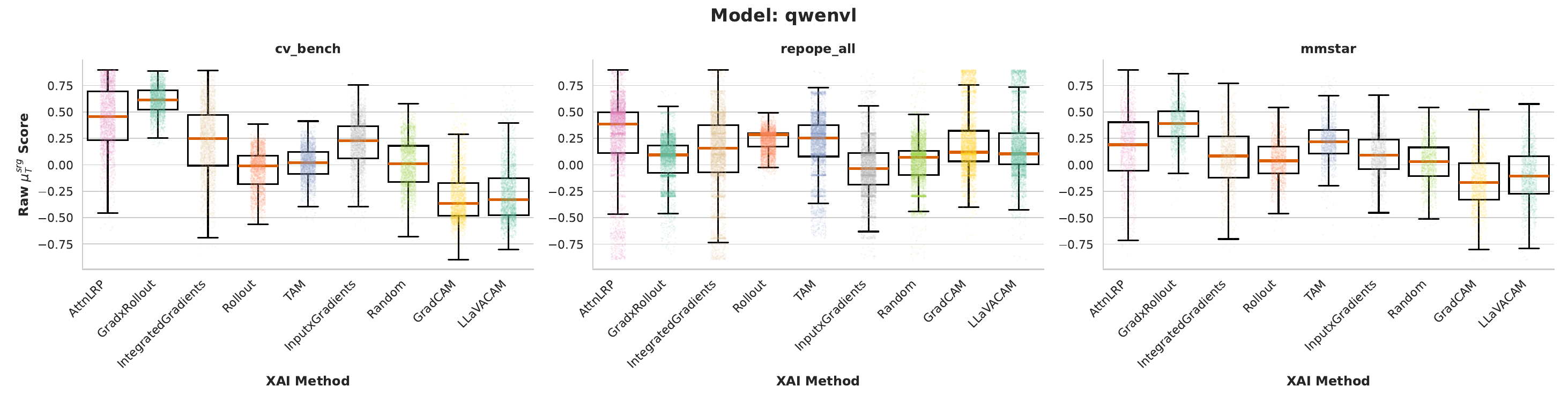}
    \centerline{(a) Qwen2.5-VL}
  \end{minipage}
  \vspace{1em}
  
  \begin{minipage}[b]{\linewidth}
    \centering
    \includegraphics[width=.9\linewidth]{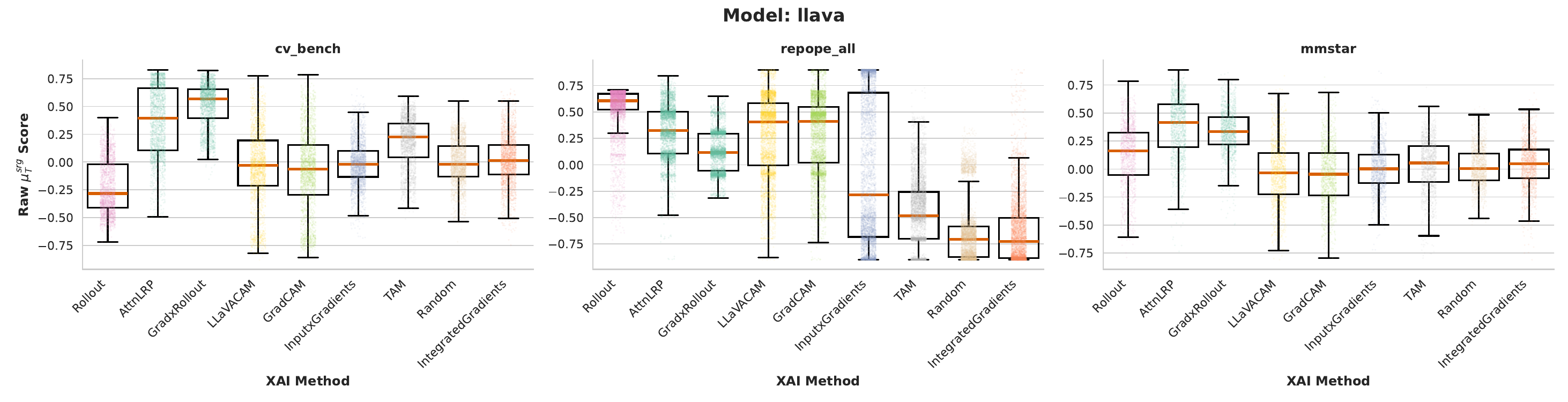}
    \centerline{(b) LLaVA-1.5}
  \end{minipage}
  \vspace{1em}
  
  \begin{minipage}[b]{\linewidth}
    \centering
    \includegraphics[width=.9\linewidth]{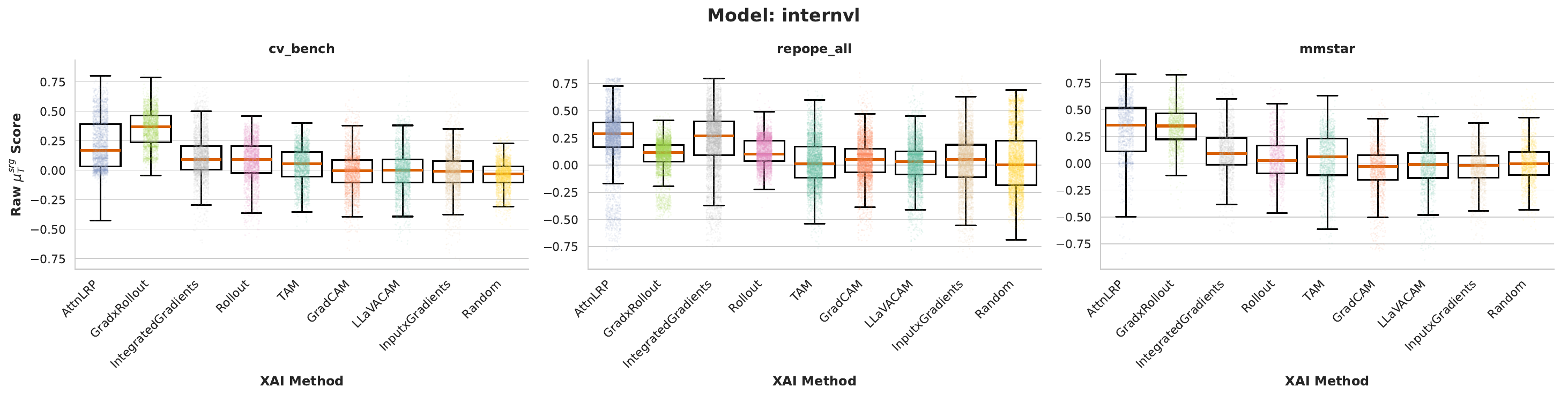}
    \centerline{(c) InternVL-3.5}
  \end{minipage}
  \caption{Raw textual Faithfulness ($\mu_{T}^{srg}$) score distributions per explainer, evaluated across individual datasets and broken down by VLM architecture: (a) Qwen2.5-VL, (b) LLaVA-1.5, and (c) InternVL-3.5. Low-opacity scatter points visualize the dense instance-level variance, while high-contrast overlaid boxplots denote the median and interquartile ranges.}
  \label{fig:app_boxplots_tok}
\end{figure}

\begin{figure}[htbp]
  \centering
  \begin{minipage}[b]{\linewidth}
    \centering
    \includegraphics[width=.9\linewidth]{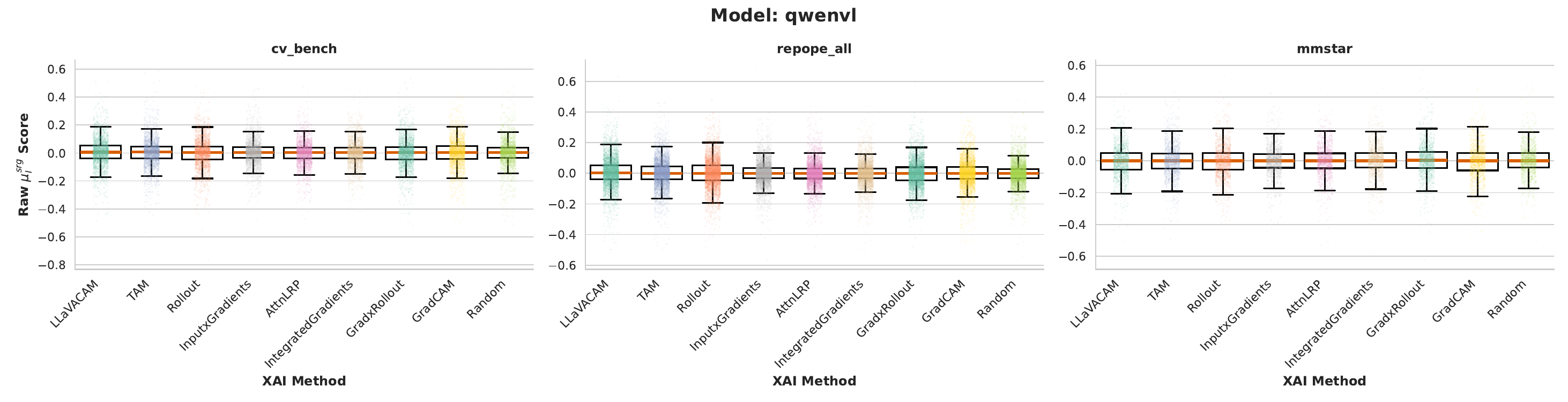}
    \centerline{(a) Qwen2.5-VL}
  \end{minipage}
  \vspace{1em}
  
  \begin{minipage}[b]{\linewidth}
    \centering
    \includegraphics[width=.9\linewidth]{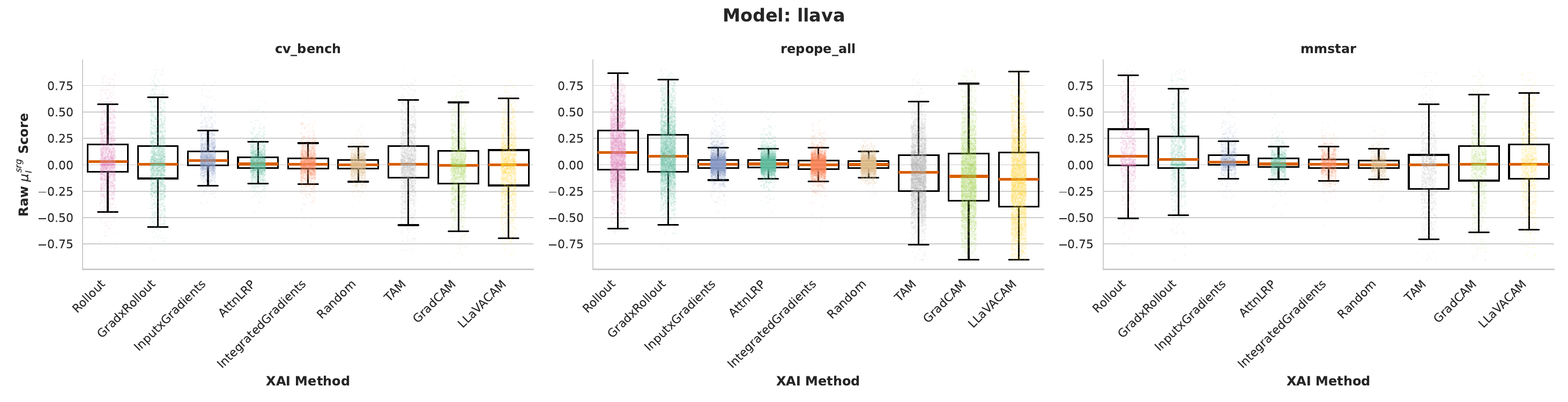}
    \centerline{(b) LLaVA-1.5}
  \end{minipage}
  \vspace{1em}
  
  \begin{minipage}[b]{\linewidth}
    \centering
    \includegraphics[width=.9\linewidth]{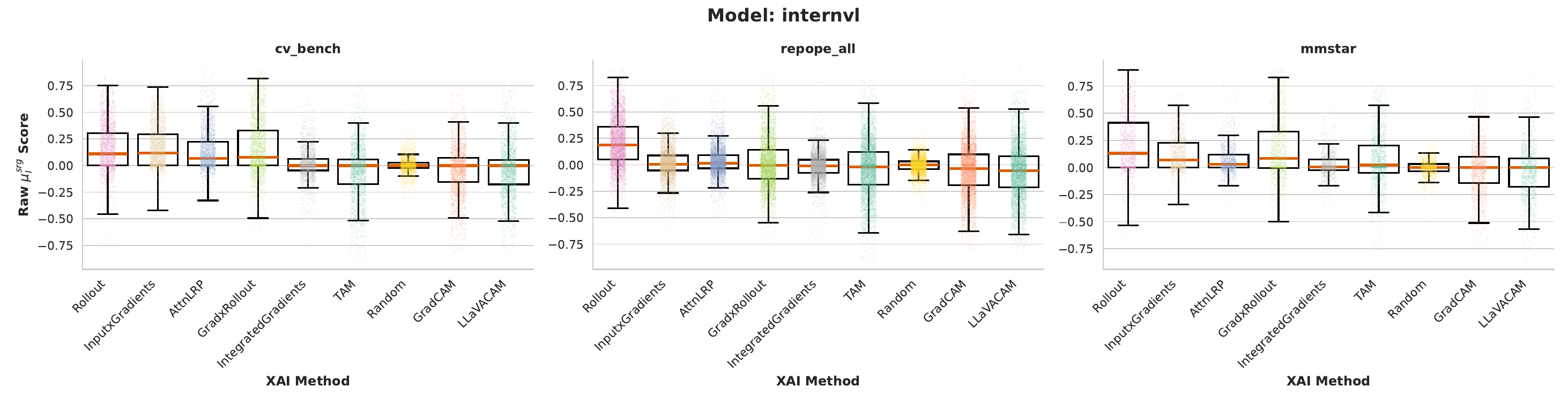}
    \centerline{(c) InternVL-3.5}
  \end{minipage}
  \caption{Raw visual faithfulness ($\mu_{I}^{srg}$) score distributions per explainer, evaluated across individual datasets and broken down by VLM architecture: (a) Qwen2.5-VL, (b) LLaVA-1.5, and (c) InternVL-3.5. Low-opacity scatter points visualize the dense instance-level variance, while high-contrast overlaid boxplots denote the median and interquartile ranges.}
  \label{fig:app_boxplots_img}
\end{figure}

We complete the results presented in the previous Tables with visualization plots to understand how explainer performance fluctuates across underlying model architectures and dataset using faithfulness metrics. Figures \ref{fig:app_boxplots_syn}, \ref{fig:app_boxplots_tok}, and \ref{fig:app_boxplots_img} present the distributions of synergistic ($\mathcal{F}_{syn}$), textual ($\mu_{T}^{srg}$), and visual ($\mu_{I}^{srg}$) faithfulness scores for Qwen2.5-VL, LLaVA-1.5, and InternVL-3.5, per dataset.

\subsection{The Efficiency-Faithfulness Trade-off}
\label{app:time_vs_perf}

\begin{figure}[htbp]
  \centering
  % --- ROW 1 ---
  \begin{minipage}[b]{0.48\linewidth}
    \centering
    \includegraphics[width=\linewidth]{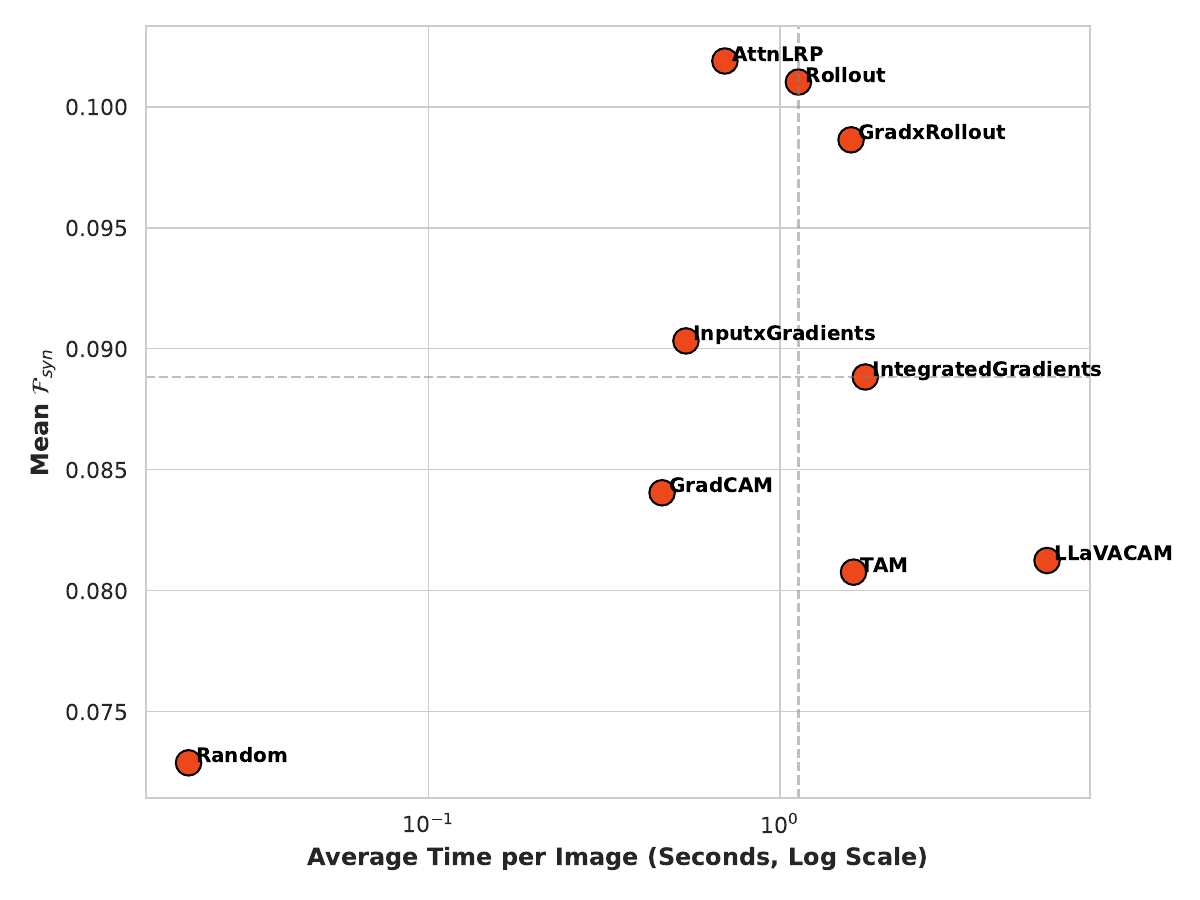} % REPLACE
    \centerline{(a) Global Trade-off}
  \end{minipage}\hfill
  \begin{minipage}[b]{0.48\linewidth}
    \centering
    \includegraphics[width=\linewidth]{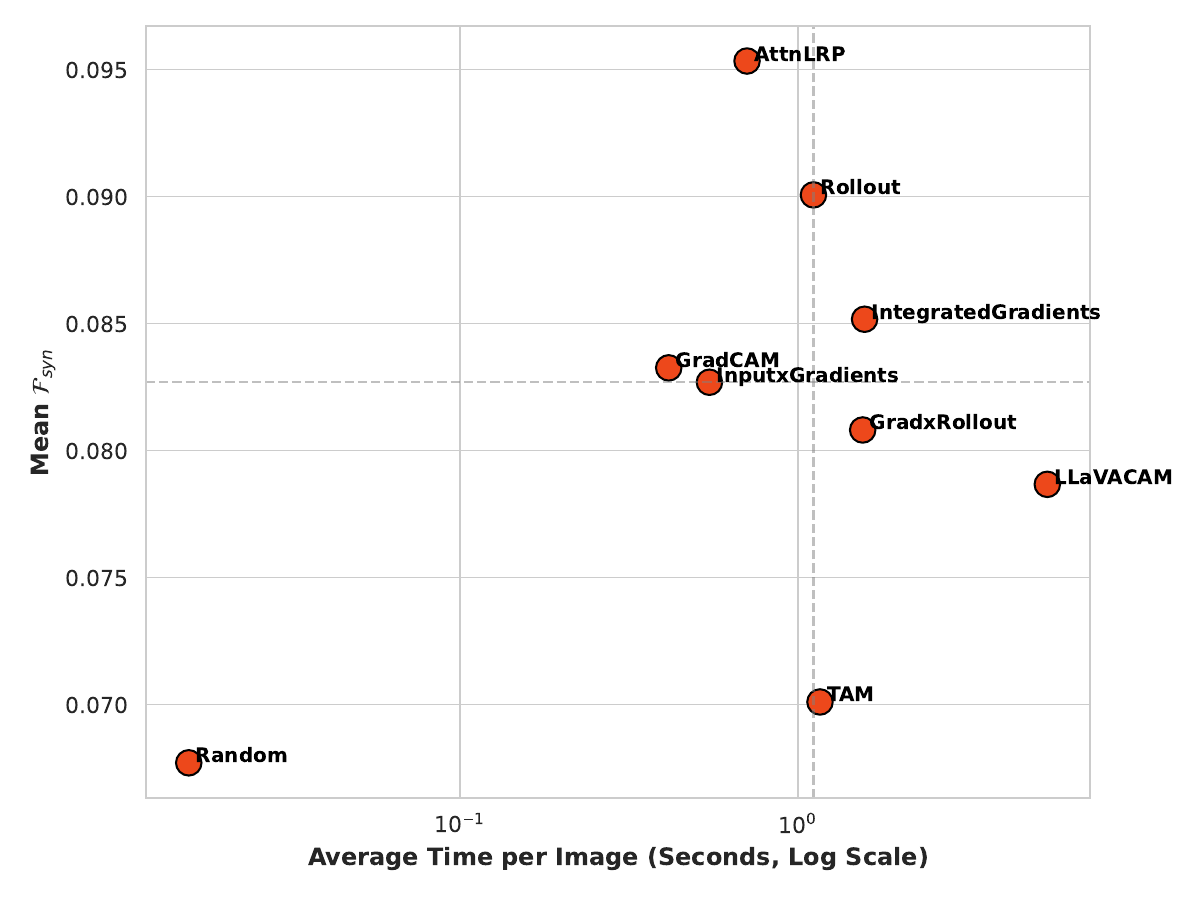} % REPLACE
    \centerline{(b) RePOPE}
  \end{minipage}

  \vspace{0.5cm} % Vertical space between rows

  % --- ROW 2 ---
  \begin{minipage}[b]{0.48\linewidth}
    \centering
    \includegraphics[width=\linewidth]{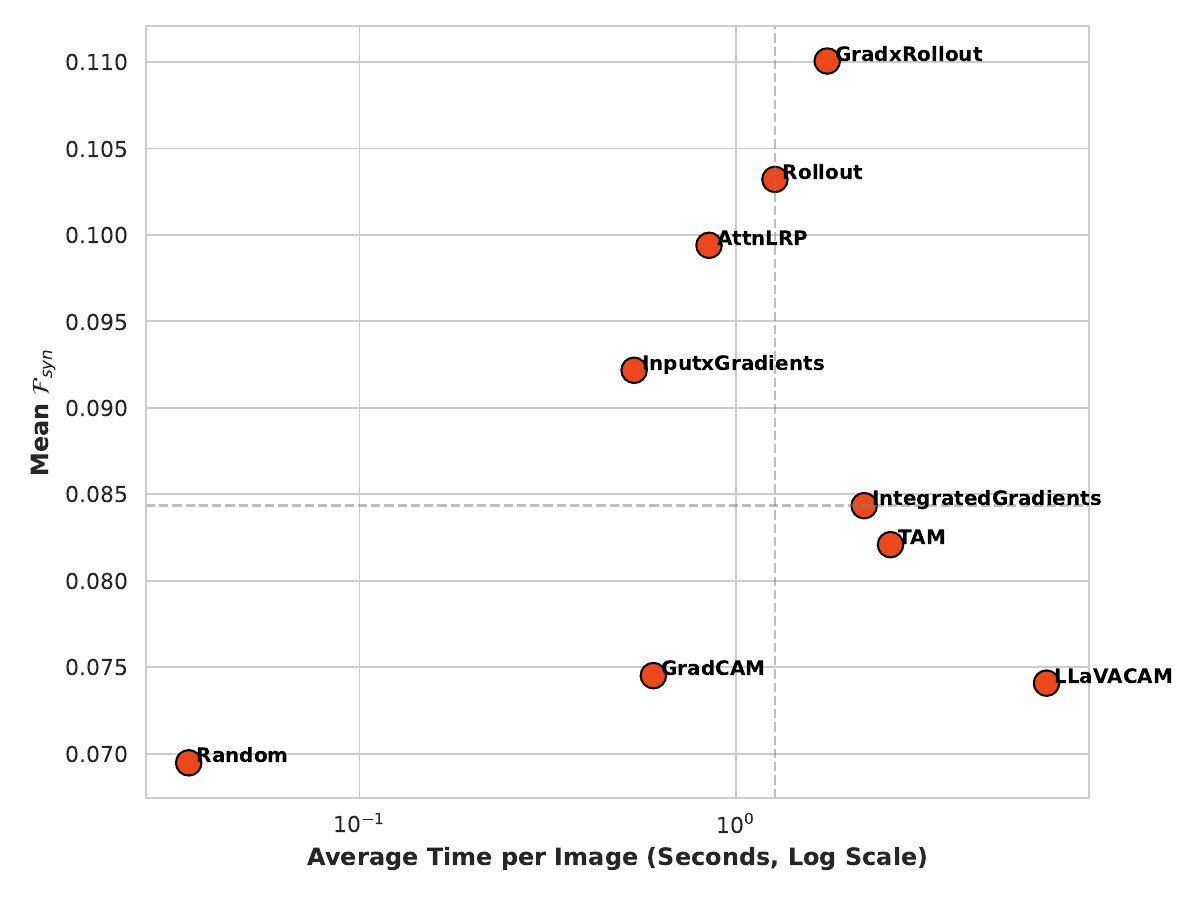} % REPLACE
    \centerline{(c) CVBench}
  \end{minipage}\hfill
  \begin{minipage}[b]{0.48\linewidth}
    \centering
    \includegraphics[width=\linewidth]{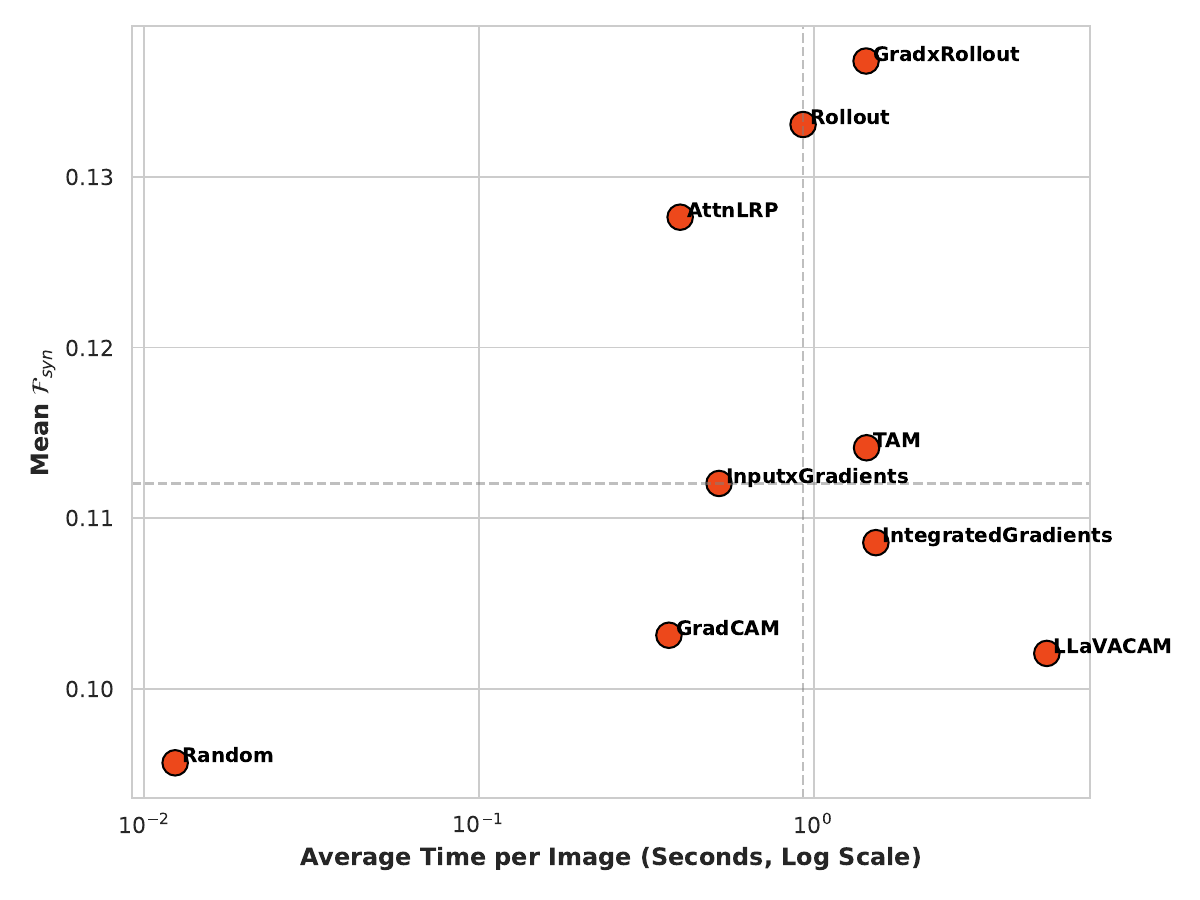} % REPLACE
    \centerline{(d) MMStar}
  \end{minipage}
  
  \caption{Pareto frontier of explainer efficiency vs. synergistic faithfulness ($\mathcal{F}_{syn}$). The ideal explainer occupies the top-left quadrant (high faithfulness, low execution time). Attention-based methods consistently dominate this frontier, offering superior multimodal grounding without the prohibitive computational costs of gradient-heavy or perturbation-heavy alternatives.}
  \label{fig:app_pareto}
\end{figure}

Evaluating practical utility requires weighing an explainer's cross-modal grounding capability against its computational overhead. Figure \ref{fig:app_pareto} maps the empirical Pareto frontier, charting average execution time against Synergistic Faithfulness ($\mathcal{F}_{syn}$). The global and dataset-specific plots reveal that while attention-based explainers achieve the highest faithfulness, they incur a significant execution cost, occupying the top-right quadrant.

Similarly, methods like Integrated Gradients (which requires multiple iterations to approximate the integration path via Riemann steps) and LLaVA-CAM (which performs smoothing over numerous samples) demand substantial computational resources. However, unlike attention-based methods, their severe computational overhead does not translate to improved cross-modal faithfulness. Ultimately, these figures expose a gap in the current landscape: the optimal "high faithfulness, low execution time" quadrant remains largely vacant, highlighting a critical direction for future research.

\takeaway{While attention-based methods deliver superior faithfulness, the execution costs of the explainers evaluated in this benchmark, demonstrate an urgent need for efficient, high-fidelity multimodal explainers.}

\subsection{Scalability to Larger VLM Architectures}
\label{app:larger_vlms}

\subsubsection{Explainers benchmark Results}
\label{app:large_synergy}

To ensure that the explainer hierarchies observed in our primary benchmark generalize to higher-capacity architectures, we extended our evaluation on the MMStar dataset to include larger parameter-count VLMs. 

\begin{table}[h]
\centering
\caption{Synergistic Faithfulness ($\mathcal{F}_{syn}$) scores on the MMStar dataset utilizing larger VLM architectures. The relative explainer hierarchy remains stable at scale.}
\label{tab:larger_vlms}
\resizebox{0.9\linewidth}{!}{
\begin{tabular}{lcccc}
\toprule
\textbf{Explainer} & \textbf{InternVL-3.5 (14B)} & \textbf{LLaVA-1.5 (13B)} & \textbf{Qwen-3.5 (27B)} & \textbf{Overall Mean} \\
\midrule
Grad$\times$Rollout & \bf 0.282 $_{(\pm0.18)}$ & \bf 0.107 $_{(\pm0.13)}$ & \bf 0.270 $_{(\pm0.17)}$ & \bf 0.204 $_{(\pm0.18)}$ \\
Rollout & \underline{0.227 $_{(\pm0.16)}$} & \underline{0.100 $_{(\pm0.12)}$} & \underline{0.260 $_{(\pm0.17)}$} & \underline{0.178 $_{(\pm0.16)}$} \\
AttnLRP & 0.231 $_{(\pm0.18)}$ & 0.097 $_{(\pm0.12)}$ & 0.232 $_{(\pm0.16)}$ & 0.174 $_{(\pm0.17)}$ \\
TAM & 0.200 $_{(\pm0.15)}$ & 0.084 $_{(\pm0.11)}$ & 0.253 $_{(\pm0.16)}$ & 0.161 $_{(\pm0.15)}$ \\
Input$\times$Gradients & 0.208 $_{(\pm0.15)}$ & 0.086 $_{(\pm0.11)}$ & 0.221 $_{(\pm0.15)}$ & 0.158 $_{(\pm0.15)}$ \\
IntegratedGradients & 0.197 $_{(\pm0.15)}$ & 0.085 $_{(\pm0.10)}$ & 0.226 $_{(\pm0.15)}$ & 0.153 $_{(\pm0.14)}$ \\
LLaVACAM & 0.200 $_{(\pm0.15)}$ & 0.076 $_{(\pm0.10)}$ & 0.230 $_{(\pm0.15)}$ & 0.152 $_{(\pm0.15)}$ \\
GradCAM & 0.193 $_{(\pm0.15)}$ & 0.077 $_{(\pm0.10)}$ & 0.243 $_{(\pm0.16)}$ & 0.151 $_{(\pm0.15)}$ \\
Random & 0.171 $_{(\pm0.13)}$ & 0.075 $_{(\pm0.09)}$ & 0.214 $_{(\pm0.14)}$ & 0.137 $_{(\pm0.13)}$ \\
\bottomrule
\end{tabular}
}
\end{table}

As detailed in Table \ref{tab:larger_vlms}, while the absolute $\mathcal{F}_{syn}$ scores fluctuate based on the underlying model's inherent reasoning capacity, the relative faithfulness rankings of the explainers remain strictly stable at scale.

\subsubsection{Linear Mixed Model Statistical Analysis}
\label{app:large_lmm}

To rigorously confirm that the explainer hierarchy established on our primary benchmark (which utilized small-size VLMs) generalizes to higher-capacity architectures, we applied a Linear Mixed Model (LMM) to the evaluation results of the larger models. The LMM isolates the fixed effect ($\beta$) of each explainer on the final $\mathcal{F}_{syn}$ score while controlling for the random effects introduced by the varying high-capacity VLM backbones.

\begin{figure}[htbp]
  \centering
  % Rename the image file in your directory to avoid spaces and special characters for LaTeX safety, e.g., 'lmm_fixed_effects.png'
  \includegraphics[width=0.85\linewidth]{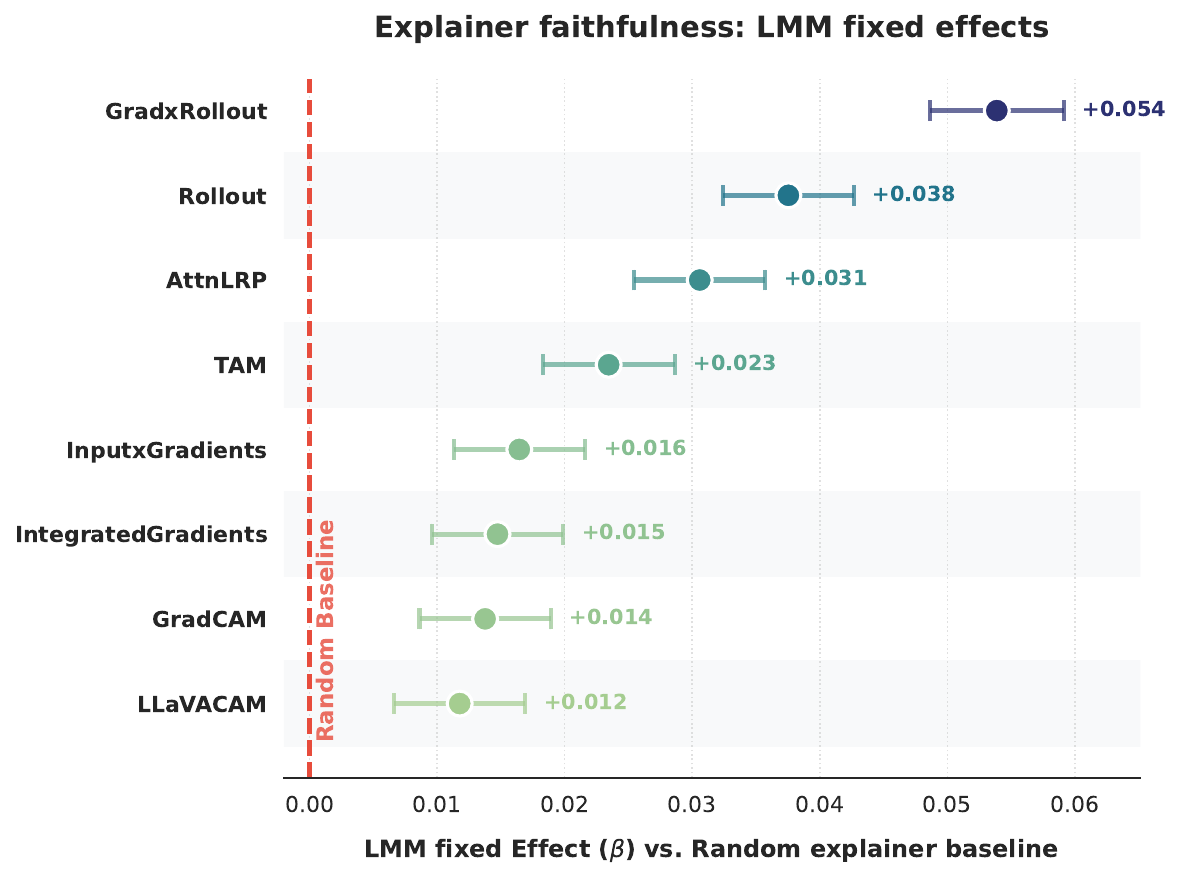}
  \caption{Linear Mixed Model (LMM) fixed effects ($\beta$) of each explainer compared to the Random attribution baseline. Higher values indicate a stronger, statistically significant positive impact on synergistic faithfulness.}
  \label{fig:lmm_fixed_effects}
\end{figure}

\begin{table}[h]
\centering
\caption{Global LMM Statistics for $\mathcal{F}_{syn}$. All values are calculated relative to the Random baseline. Significance thresholds: $^* p < 0.05$, $^{**} p < 0.01$, $^{***} p < 0.001$.}
\label{tab:large_lmm_pvalues}
\begin{tabular}{lccc}
\toprule
\textbf{Explainer} & \textbf{$\beta$ Coefficient} & \textbf{Standard Error (SE)} & \textbf{$p$-value} \\
\midrule
Grad$\times$Rollout & \bf +0.054 & 0.003 & $<0.001^{***}$ \\
Rollout & \underline{+0.038} & 0.003 & $<0.001^{***}$ \\
AttnLRP & +0.031 & 0.003 & $<0.001^{***}$ \\
\midrule
TAM & +0.023 & 0.003 & $<0.001^{***}$ \\
LLaVACAM & +0.012 & 0.003 & $<0.001^{***}$ \\
\midrule
Input$\times$Gradients & +0.016 & 0.003 & $<0.001^{***}$ \\
IntegratedGradients & +0.015 & 0.003 & $<0.001^{***}$ \\
GradCAM & +0.014 & 0.003 & $<0.001^{***}$ \\
\bottomrule
\end{tabular}
\end{table}

As illustrated in Figure \ref{fig:lmm_fixed_effects} and detailed in Table \ref{tab:lmm_pvalues}, all evaluated explainers maintain a statistically significant improvement over random attribution ($p < 0.001$) on these larger models. Crucially, the $\beta$ coefficients extracted from the larger VLMs perfectly mirror the hierarchy observed in the small-model benchmark: attention-based methods (e.g., Grad$\times$Rollout, $\beta = +0.054$) continue to significantly outperform VLM-native methods (e.g., LLaVACAM, $\beta = +0.012$). This confirms that the underperformance of native explainers is not an artifact of limited model capacity, but a consistent architectural trend that persists at scale.

\takeaway{We prove that our experimental results in Section \ref{sec:benchmark_results} are not an artifact of the size of the model. The LMM analysis confirms that the attention-based methods remains statistically significant and ordinally stable even as VLM architectures scale up to 27B parameters.}

\section{Game-Theoretic Foundations and Interaction Estimation}\label{appendix_gametheory}
This section details the exact implementation of the cooperative game-theoretic baseline introduced in Section \ref{method}, specifically outlining how we bridged the intractable theoretical formulation of the Shapley Interaction Index (SII) to a computable, zero-variance ground truth.

\subsection{Intractability of the Multimodal Micro-Game}
In the main text, we formulated multimodal inference as a cooperative game over the micro-feature set $\mathcal{M} = I \cup T$. While the exact Shapley Interaction Index $\Phi_{ij}(\nu)$ provides the theoretical gold standard for measuring the cross-modal Harsanyi dividend $\Delta_{i,j}\nu(S)$ between a specific patch $p_i$ and token $t_j$, computing it is fundamentally a \textit{\#P-hard} problem. Because a modern VLM processes thousands of visual patches and text tokens ($|\mathcal{M}| \approx 2000$), calculating the exact interaction requires evaluating the characteristic function $\nu(S)$ over $2^{|\mathcal{M}|}$ unique coalitions. This combinatorial explosion renders the exact micro-game computationally impossible, necessitating a rigorous dimensionality reduction to establish a valid ground truth.

\subsection{Estimation via Coupled Macro-Coalitions}
\label{app:macro_coalitions}
To establish the ground-truth baseline used to validate our $\mathcal{F}_{syn}$ metric (Section 4.3), we constructed a macro-coalitional game \cite{muschalik2024shapiq}. Rather than treating every individual background patch and token as an independent player, we aggregated the atomic features into explicit \textit{cross-modal coalitions} (macro-players). This drastically reduces the player set $|\mathcal{M}|$ while perfectly preserving the complex multimodal interaction dynamics required to compute the baseline.

For a given instance and explainer at a specific continuous perturbation threshold $k$, we map the micro-features to an 8-player macro-game:
\begin{enumerate}
    \item \textbf{Unimodal target players (Players 1 \& 2):} Player 1 is defined as the subset of top-attributed visual patches ($I_k \subseteq I$). Player 2 is defined as the subset of top-attributed textual tokens ($T_k \subseteq T$).
    \item \textbf{Cross-modal background coalitions (Players 3-8):} The remaining background image patches ($I \setminus I_k$) and background text tokens ($T \setminus T_k$) are independently shuffled and partitioned into $C=6$ equal subsets. We form 6 bimodal macro-players by explicitly coupling these modalities. Thus, each background Player $c$ is inherently multimodal: activating it simultaneously reveals the $c$-th subset of background text \textit{and} the $c$-th subset of background image.
\end{enumerate}

This dimensionality reduction condenses the intractable micro-game into exactly 8 players, reducing the total state space from approximately $2^{2000}$ to just $2^8 = 256$ possible coalitions. By utilizing the \texttt{shapiq} framework \cite{muschalik2024shapiq} with an evaluation budget of 400 forward passes, our allocated budget strictly exceeded the total state space. Consequently, this allowed us to compute the \textit{exact, deterministic} SII for the macro-game, completely eliminating sampling variance. This process yields the highly accurate ground-truth measurement of cross-modal synergy used in Figure \ref{fig:correlation_subplots}, mathematically validating that our proposed $\mathcal{F}_{syn}$ approximation preserves near-perfect ordinal alignment ($\rho=0.92$) while reducing the computational cost per sample to just $6K+2$ forward passes.

\subsection{Robustness of the Macro-Coalition Baseline (Varying $C$)}
\label{app:varying_c}

Our at-scale validation in Section \ref{sec:macro_correlations} utilizes a macro-coalition Shapley approximation to create a computationally tractable ground truth for standard VLM resolutions. To ensure that this baseline is mathematically robust and not artificially biased by the hyperparameter $C$ (the number of background macro-coalitions), we conducted a stability ablation. We evaluated both the instance-level correlation between $\mathcal{F}_{syn}$ and the macro-coalition SII, as well as the computational cost required to generate the baseline across varying values of $C$. 

\begin{figure}[htbp]
  \centering
  \begin{minipage}[t]{0.48\linewidth}
    \vspace{0pt}
    \centering
    \includegraphics[width=\linewidth]{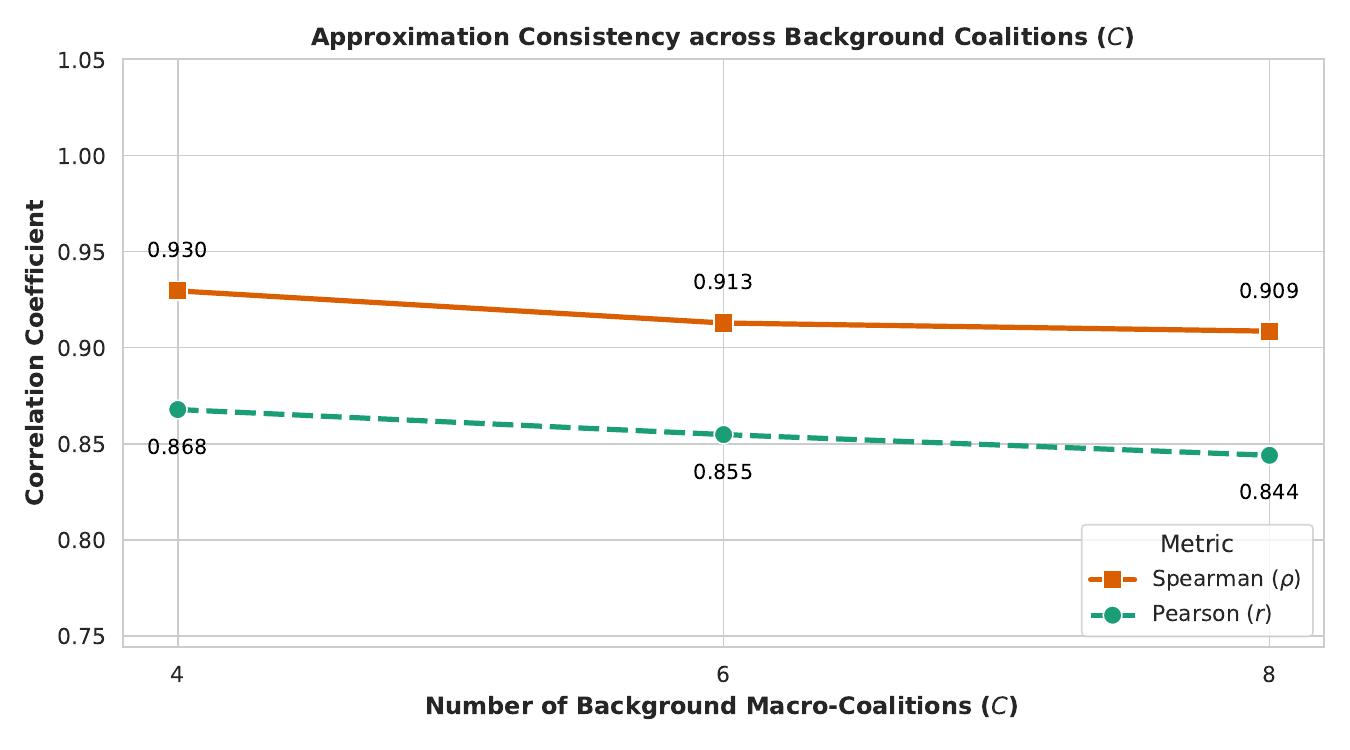}
  \end{minipage}
  \hfill
  \begin{minipage}[t]{0.48\linewidth}
    \vspace{0pt}
    \centering
    \includegraphics[width=\linewidth]{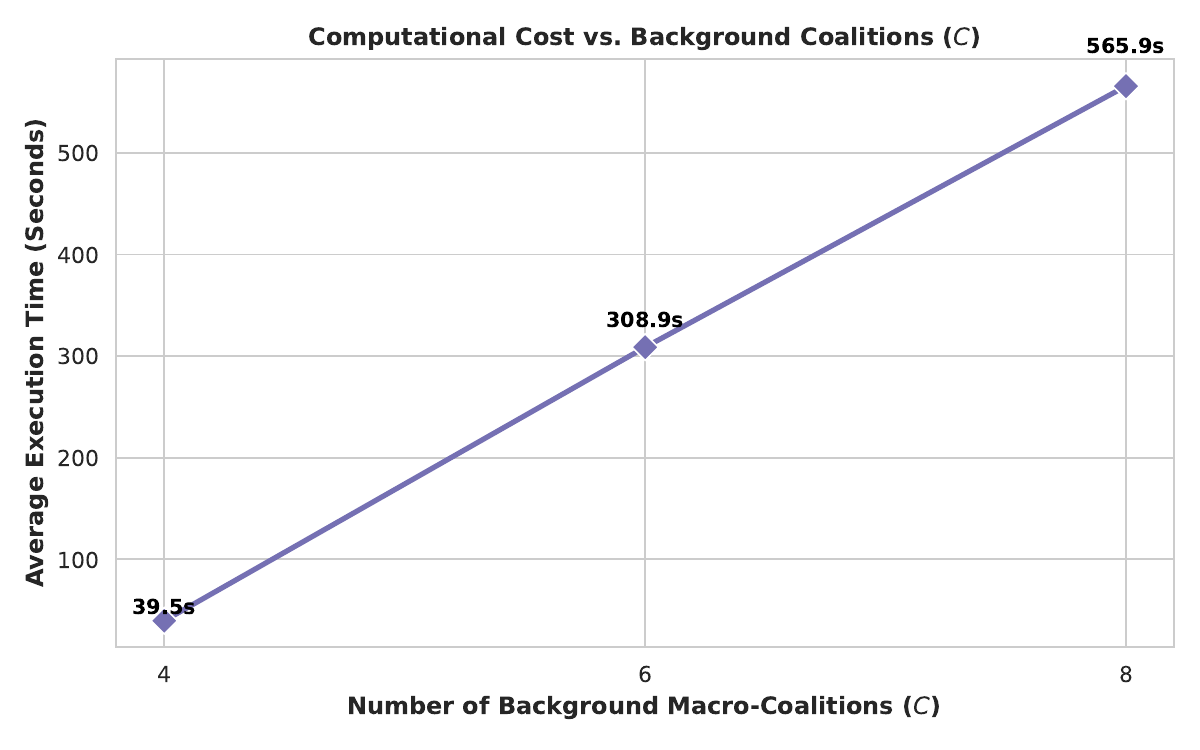}
  \end{minipage}
  \caption{Impact of varying the number of background macro-coalitions ($C$). \textbf{Left:} The Spearman ($\rho$) and Pearson ($r$) correlations between $\mathcal{F}_{syn}$ and the SII baseline remain highly stable regardless of background clustering granularity. \textbf{Right:} The average execution time per instance for the macro-coalition SII scales severely as $C$ increases.}
  \label{fig:c_stability}
\end{figure}

As shown in Figure \ref{fig:c_stability} (left), the correlation remains highly consistent across different $C$ values, maintaining a Spearman correlation $\rho > 0.90$ across all tested granularities[cite: 5]. This confirms that our validation baseline is structurally sound and that $\mathcal{F}_{syn}$ accurately tracks true interaction effects independently of how the background state space is partitioned. 

However, as demonstrated in Figure \ref{fig:c_stability} (right), the execution time for $\mathcal{F}_{syn}$ scales linearly with $C$, jumping from 39.5 seconds at $C=4$ to 565.9 seconds at $C=8$. This computational cost underscores the importance of the choice of $C$ to make $\mathcal{F}_{syn}$ efficient for at-scale evaluation.

\takeaway{While the macro-coalition approximation maintains strict ordinal stability across varying background granularities ($\rho > 0.90$), its computational cost scales linearly with $C$.}

\subsection{Exact Atomic Shapley Validation}
\label{app:exact_sii_validation}

In Section \ref{sec:exact_correlations} of the main text, we presented a dual-stage validation of our synergistic faithfulness metric ($\mathcal{F}_{syn}$). While the main text visualizes the macro-coalition baseline to demonstrate architectural generalization at scale across multiple models, this section provides the detailed empirical results for the exact atomic Shapley Interaction Index (SII) validation.

Because computing the exact, un-approximated SII is mathematically intractable for standard resolutions, this experiment was uniquely constrained to the QwenVL-2.5 architecture \cite{bai2025qwen3}. By dynamically downsampling the input images to $32 \times 32$ pixels, we strictly bounded the game-theoretic state space to $N \le 17$ atomic players. This constraint allowed us to compute the exact interaction index without any spatial approximation, providing a pure 1:1 ground-truth comparison.

% \begin{figure}[htbp]
%   \centering
%   \includegraphics[width=0.85\linewidth]{figs/full_sii_correlation_scatter.pdf}
%   \caption{Correlation between the exact atomic Shapley Interaction Index (SII) and the proposed $\mathcal{F}_{syn}$ metric on QwenVL-2.5. The global correlation reaches $\rho=0.93$ and $r=0.91$, with strong alignment consistently preserved across fundamentally different explainers: Random ($\rho=0.94$), Input$\times$Gradients ($\rho=0.91$), Rollout ($\rho=0.93$), and TAM ($\rho=0.92$)[cite: 3].}
%   \label{fig:exact_sii_correlation}
% \end{figure}

\begin{figure}[t]
  \centering
  \begin{minipage}[b]{0.48\linewidth}
    \centering
    \includegraphics[width=\linewidth]{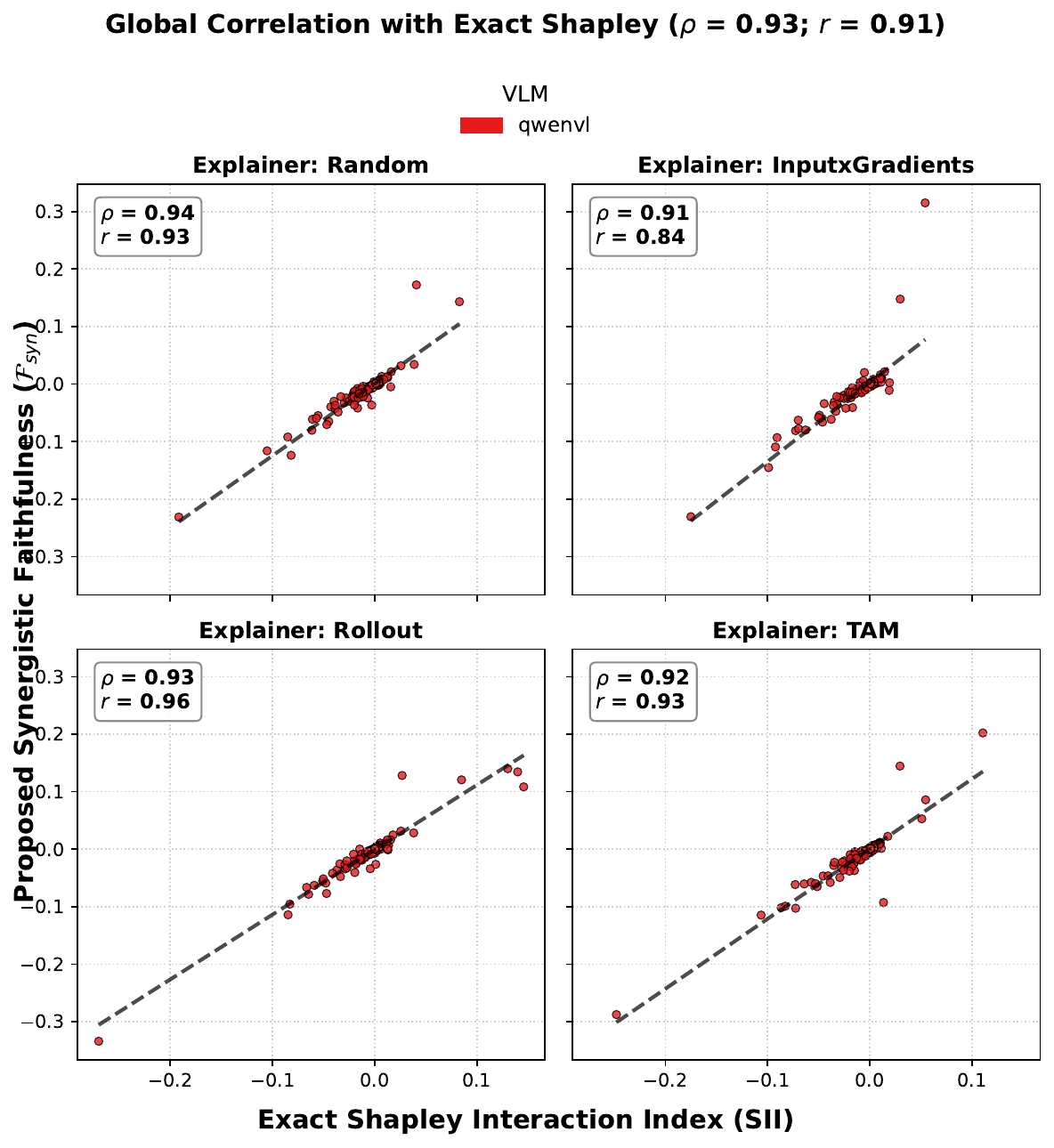}
    \caption{Correlation between the exact atomic Shapley Interaction Index (SII) and the proposed $\mathcal{F}_{syn}$ metric on QwenVL-2.5.}
    \label{fig:exact_sii_correlation}
  \end{minipage}
  \hfill
  \begin{minipage}[b]{0.48\linewidth}
    \centering
    \includegraphics[width=\linewidth]{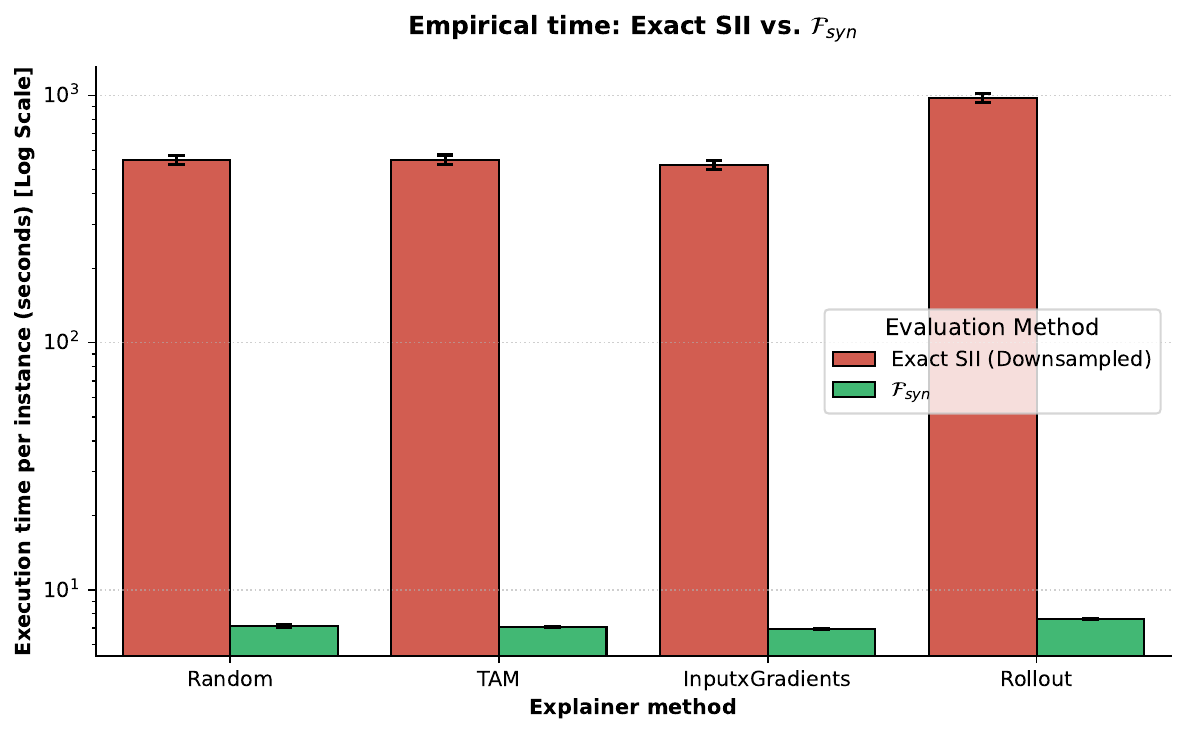}
    \caption{Empirical execution time per instance (log scale) for the exact atomic SII versus $\mathcal{F}_{syn}$. The proposed metric achieves a 88.7x speedup over the exact Shapley across all evaluated explainers.}
    \label{fig:exact_sii_runtime}
  \end{minipage}
\end{figure}

As illustrated in Figure \ref{fig:exact_sii_correlation}, $\mathcal{F}_{syn}$ maintains a highly stable rank alignment with the exact atomic index across all evaluated explainers. This confirms that our continuous approximation accurately preserves true game-theoretic feature-ranking fidelity independently of macroscopic grouping. Crucially, as demonstrated in Figure \ref{fig:exact_sii_runtime}, $\mathcal{F}_{syn}$ achieves this high mathematical fidelity while circumventing the severe combinatorial bottleneck of exact Shapley computation. While the exact downsampled SII requires hundreds of seconds per instance, $\mathcal{F}_{syn}$ evaluates the same features in under 10 seconds. This computational acceleration is what enables the large-scale benchmark executed in this study.

\takeaway{Constrained full-enumeration on exact atomic features proves that $\mathcal{F}_{syn}$ maintains near-perfect mathematical fidelity ($\rho=0.93$) to the true Shapley Interaction Index -as does the macro-coalition approximation, while circumventing its severe combinatorial bottleneck to achieve an $88.7\times$ computational speedup.}

\subsection{The granularity-variance trade-off}
The choice of $C$ (the number of background macro-coalitions) dictates a fundamental trade-off between approximation granularity and estimation variance, a recognized challenge in Shapley value estimation \cite{covert2021improving, mitchell2022sampling}. While increasing the number of macro-players (high $C$) yields a higher-resolution approximation of the underlying micro-game, the coalition space grows exponentially ($2^{C+2}$). Consequently, exact computation becomes intractable, forcing reliance on Monte Carlo sampling. For second-order interaction indices, sampling introduces severe estimation variance whenever the game dimensionality exceeds the available evaluation budget \cite{muschalik2024shapiq}. Conversely, minimizing the number of players (low $C$) drastically restricts the state space, permitting exact, zero-variance computation. However, an excessively low $C$ risks introducing aggregation bias: forcing highly heterogeneous background features to activate simultaneously can artificially smooth out localized, non-linear interactions. By setting $C=6$, we strike a practical balance. This configuration guarantees sufficient background perturbation to accurately isolate the cross-modal Harsanyi dividends, while constraining the total state space to $2^8 = 256$ coalitions. Because this space is strictly bounded below our evaluation budget of 400 forward passes, we completely neutralize sampling variance \cite{mitchell2022sampling}. The result is a mathematically rigorous, deterministic ground truth used to validate our proposed $\mathcal{F}_{syn}$ metric.

\textbf{Computational Efficiency vs. Prior Work:} Standard micro-feature Shapley estimation faces an intractable state space (e.g., $2^{\sim 2000}$), forcing researchers to allocate massive evaluation budgets, often exceeding $10^6$ forward passes \cite{banieckiexplaining} merely to mitigate Monte Carlo sampling variance. In contrast, our macro-coalitional formulation bounds the player set to $N=8$, restricting the complete state space to just $2^8 = 256$. By allocating a budget of $B = 400$ forward passes, we strictly over-budget the total power set. This guarantees that the \texttt{shapiq} framework exhausts all permutations, entirely bypassing stochastic approximation. Thus, our approach computes the exact, zero-variance Shapley Interaction Index with optimal efficiency, rendering the massive evaluation budgets of prior work mathematically redundant.

\section{Extended Analysis of Spatial Plausibility vs. Cross-Modal Faithfulness}
\label{spatial_plausibility}
\subsection{Decomposing Visual Salience: Plausibility versus Faithfulness}
\label{app:coco_plausibility}

A recurring question when analyzing multimodal attributions is whether VLM-native explainers genuinely capture cross-modal reasoning or merely default to unimodal visual salience (i.e., functioning as "object detectors"). Directly decomposing a deep VLM's attribution tensor into orthogonal "visual-only" and "cross-modal" sub-matrices is mathematically intractable. As demonstrated by Deng et al. \cite{deng2024unifying}, post-hoc attributions are formally equivalent to weighted allocations of Taylor interaction effects. Because visual and textual modalities become deeply entangled through dozens of self-attention layers, their Taylor interaction terms cannot be algebraically severed without destroying the structural integrity of the attribution mechanism.

To evaluate the "object detector" hypothesis empirically, we measure the spatial plausibility of the explainers on the COCO captioning dataset \cite{lin2014microsoft, chen2015microsoft}. Unlike VQA tasks that require dynamic cross-modal reasoning, COCO primarily requires pure visual localization. We evaluate the explainers using Object Intersection over Union (IoU) and Point Game metrics \cite{li2025token}.

\begin{table}[htbp]
\centering
\caption{Spatial plausibility on the COCO dataset. Specific VLM-native explainers (e.g., TAM) exhibit strong unimodal object detection capabilities.}
\label{tab:coco_plausibility_app}
\resizebox{0.7\linewidth}{!}{
\begin{tabular}{lcc}
\toprule
\textbf{Explainer} & \textbf{Object IoU} $\uparrow$ & \textbf{Point Game} $\uparrow$ \\
\midrule
\textbf{TAM} & \textbf{0.212} & \textbf{0.420} \\
Random & 0.118 & 0.122 \\
LLaVACAM & 0.109 & 0.190 \\
Rollout & 0.100 & 0.156 \\
Grad$\times$Rollout & 0.097 & 0.344 \\
GradCAM & 0.091 & 0.167 \\
IntegratedGradients & 0.070 & 0.109 \\
Input$\times$Gradients & 0.069 & 0.130 \\
AttnLRP & 0.066 & 0.175 \\
\bottomrule
\end{tabular}
}
\end{table}

\begin{figure}[htbp]
  \centering
  % Row 1
  \begin{minipage}[t]{0.48\linewidth}
    \vspace{0pt}
    \centering
    \includegraphics[width=\linewidth]{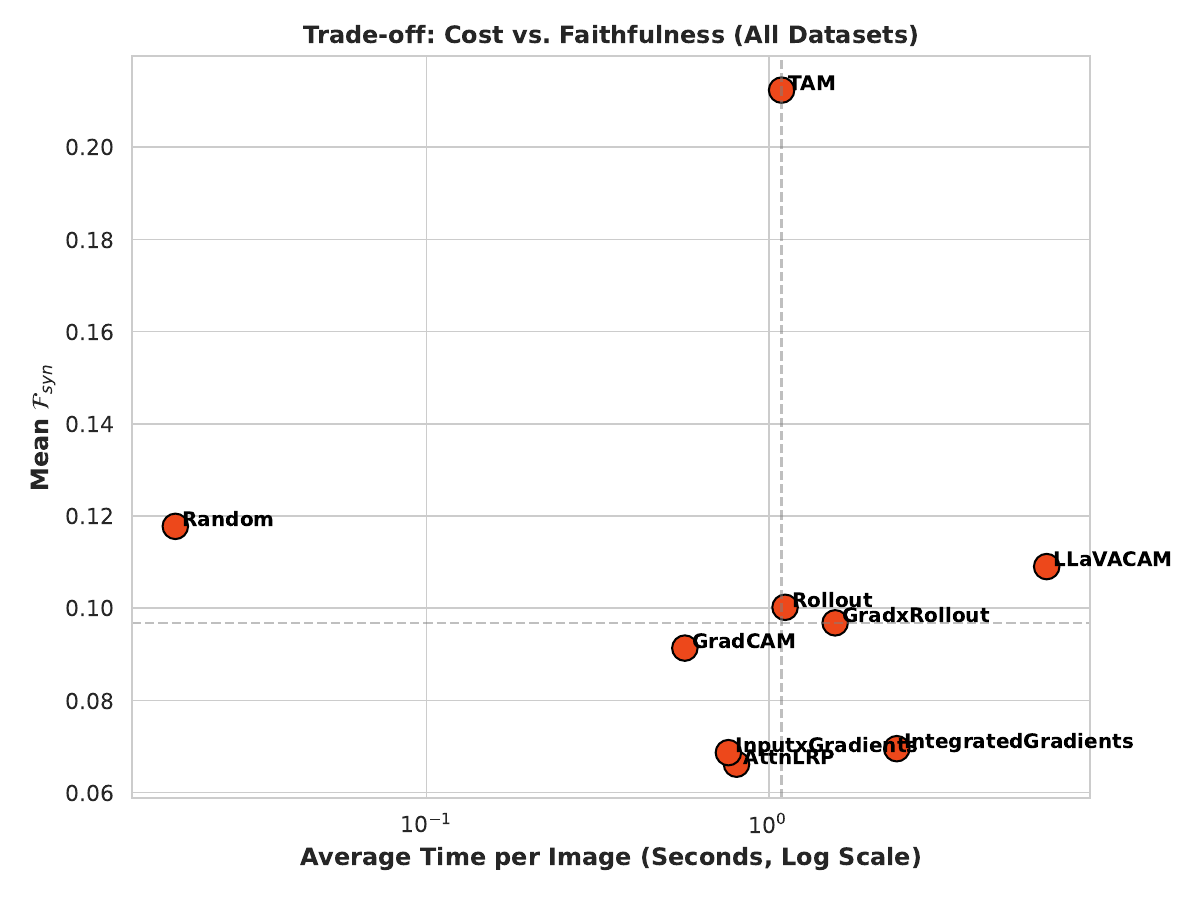}
  \end{minipage}
  \hfill
  \begin{minipage}[t]{0.48\linewidth}
    \vspace{0pt}
    \centering
    \includegraphics[width=\linewidth]{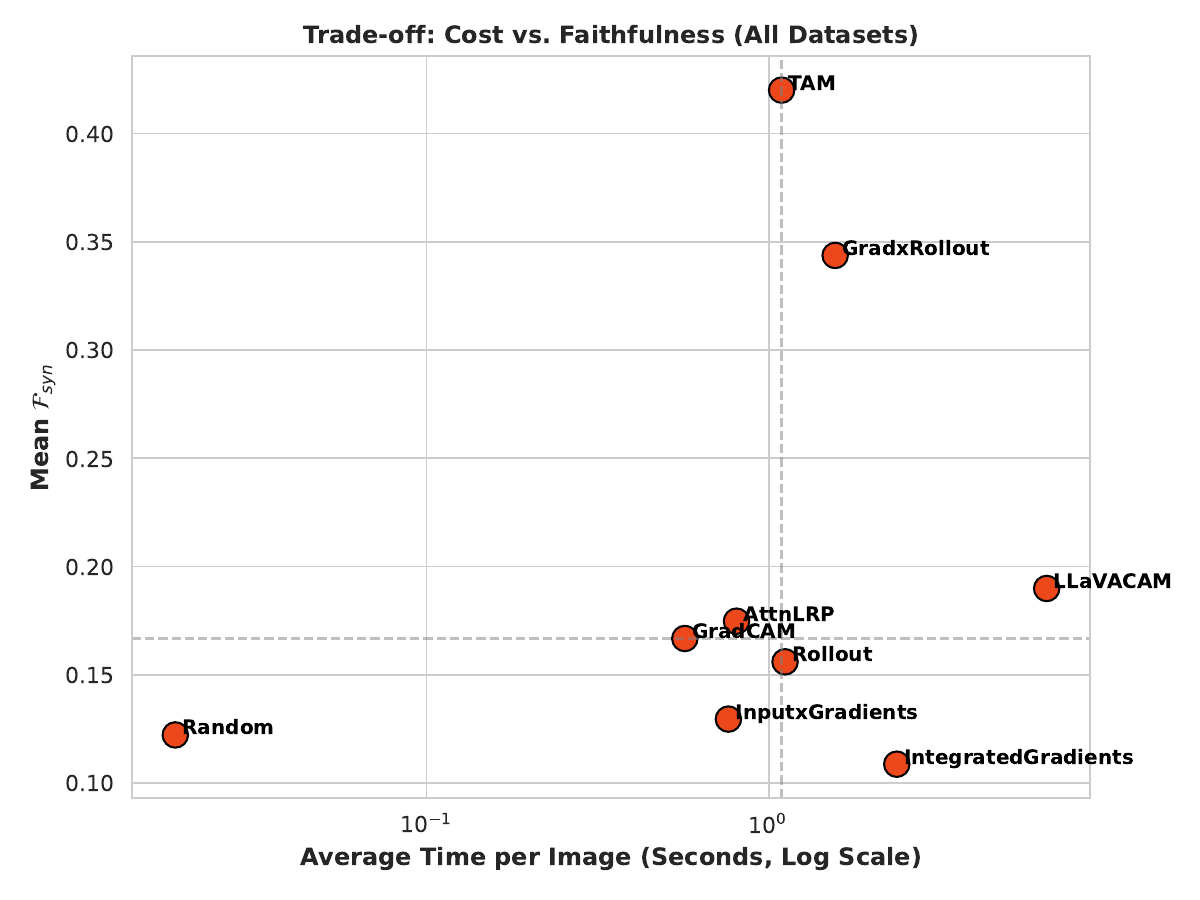}
  \end{minipage}
  
  \vspace{1.5em} % Vertical spacing between rows
  
  % Row 2
  \begin{minipage}[t]{0.48\linewidth}
    \vspace{0pt}
    \centering
    \includegraphics[width=\linewidth]{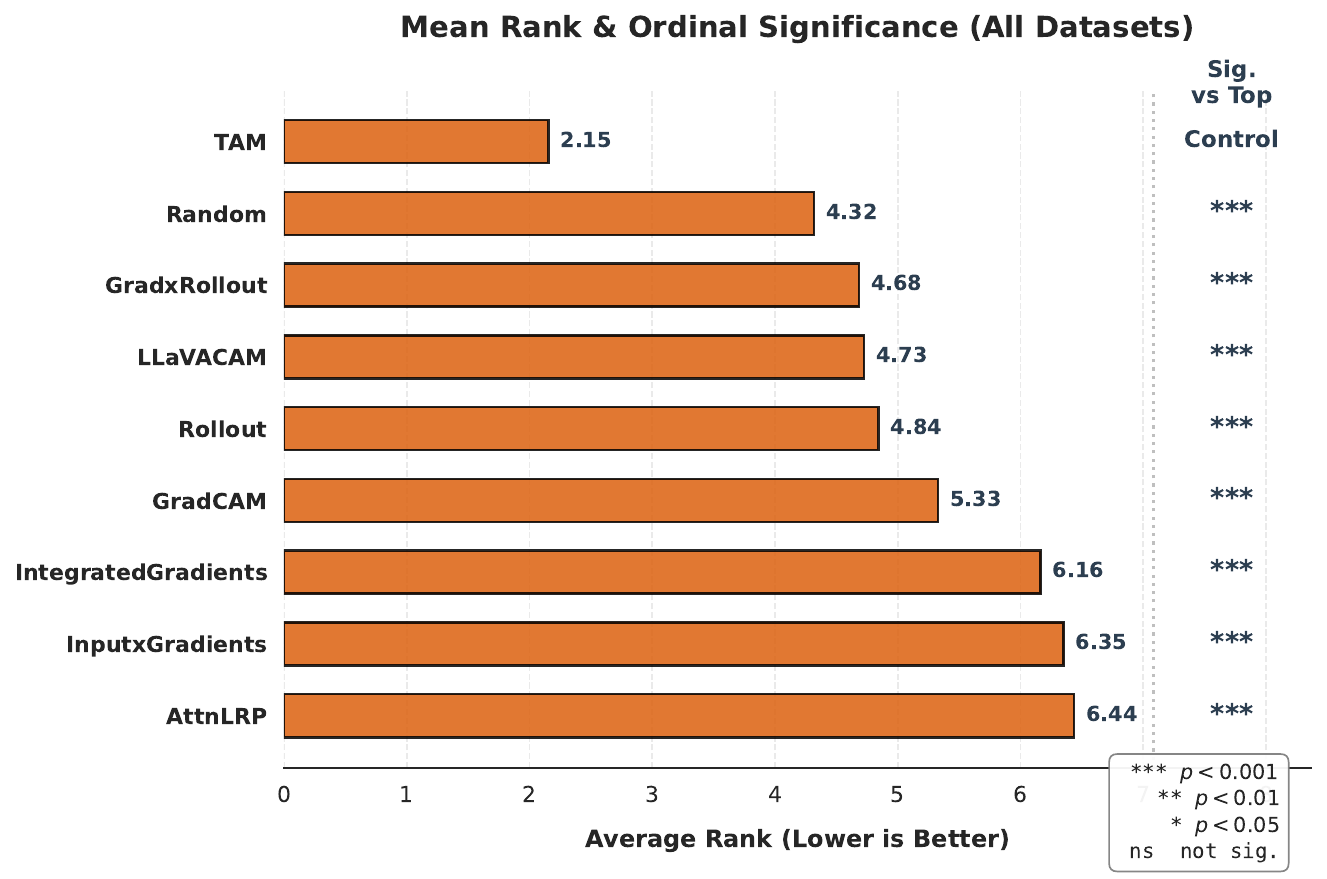}
  \end{minipage}
  \hfill
  \begin{minipage}[t]{0.48\linewidth}
    \vspace{0pt}
    \centering
    \includegraphics[width=\linewidth]{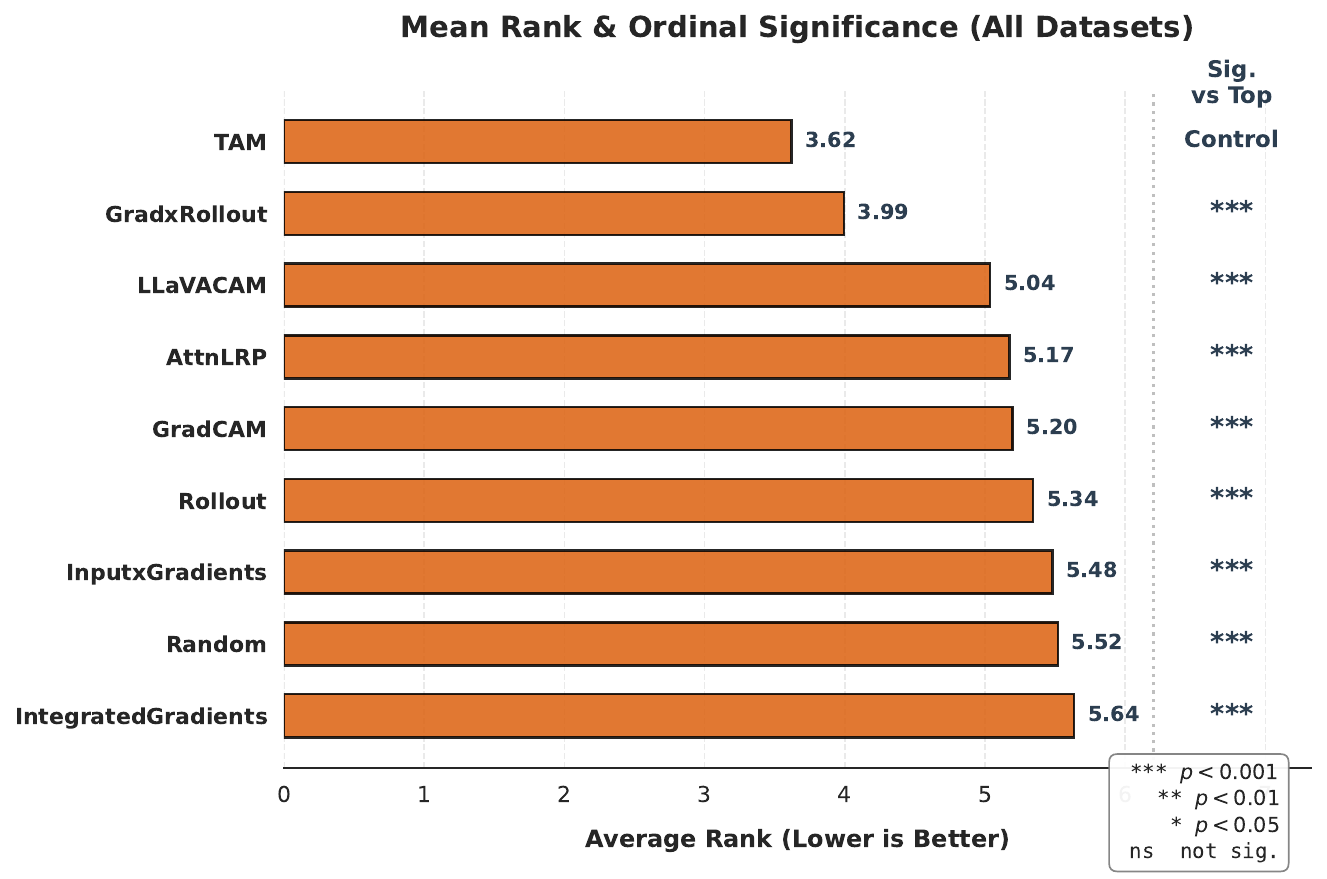}
  \end{minipage}
  \caption{Comprehensive spatial plausibility analysis on the COCO dataset. \textbf{Top Left:} Cost vs. performance trade-off for Object IoU. \textbf{Top Right:} Cost vs. performance trade-off for Point Game. \textbf{Bottom Left:} Mean rank and ordinal significance for Object IoU. \textbf{Bottom Right:} Mean rank and ordinal significance for Point Game.}
  \label{fig:plausibility_grid}
\end{figure}

To further dissect these spatial plausibility results, Figure \ref{fig:plausibility_grid} provides a comprehensive 2$\times$2 analysis illustrating the trade-offs and ordinal significance of the explainers. The top row visualizes the cost-versus-performance trade-offs for Object IoU and Point Game. The bottom row details the average ordinal ranks and statistical significance for the same metrics, confirming TAM's significant dominance as a unimodal localizer. 

As shown in Table \ref{tab:coco_plausibility_app} and Figure \ref{fig:plausibility_grid}, specific methods such as TAM may act as unimodal object detectors, yielding high spatial plausibility metrics. However, as demonstrated in our main benchmark, this visual salience does not translate to cross-modal faithfulness. Despite its high plausibility on COCO, TAM yields a low Synergistic Faithfulness score ($\mathcal{F}_{syn} = 0.114$) on datasets like MMStar that require joint reasoning. This indicates that while optimizing for human-interpretable visual bounding boxes, specific explainers may fail to capture the actual joint interactions utilized by the VLM.

\takeaway{Some VLM-native explainers exhibit a severe multimodal \textit{Clever Hans} effect: they optimize for human-interpretable spatial plausibility but fail to capture the actual cross-modal interactions utilized by the VLM.}

\section{Extensive qualitative case studies}\label{appendix_quality}
\label{app:qualitative_analysis}
\begin{figure*}[htbp] 
  \centering
  
  \textbf{Prompt:} \textit{"How many scarfs are in the image? Select from the following choices. (A) 3 (B) 1 (C) 0 (D) 2 Answer directly with only the letter inside parentheses, and nothing else. Answer :"} \\
  \textbf{Answer:} \textit{(B)}
  \vspace{0.2cm}
  
  % --- ROW 1 (5 Panels) ---
  \begin{minipage}[b]{0.19\linewidth}
    \centering
    % Force the height to exactly match the image block of the others.
    % (Tweak the 4.0cm up or down until it lines up perfectly)
    \includegraphics[width=.7\linewidth]{figs/figs_quality/qual_118_original.pdf} \\
    \vspace{0.8cm} % Tweak this to align "(a)" with "(b)"
    \centerline{\small (a) Original Input}
  \end{minipage}\hfill
  \begin{minipage}[b]{0.19\linewidth}
    \centering
    \includegraphics[width=\linewidth]{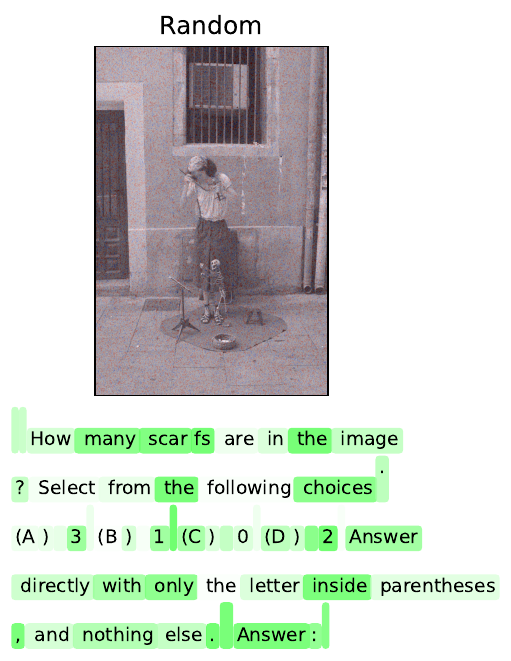}
    \centerline{\small (b) Random}
  \end{minipage}\hfill
  \begin{minipage}[b]{0.19\linewidth}
    \centering
    \includegraphics[width=\linewidth]{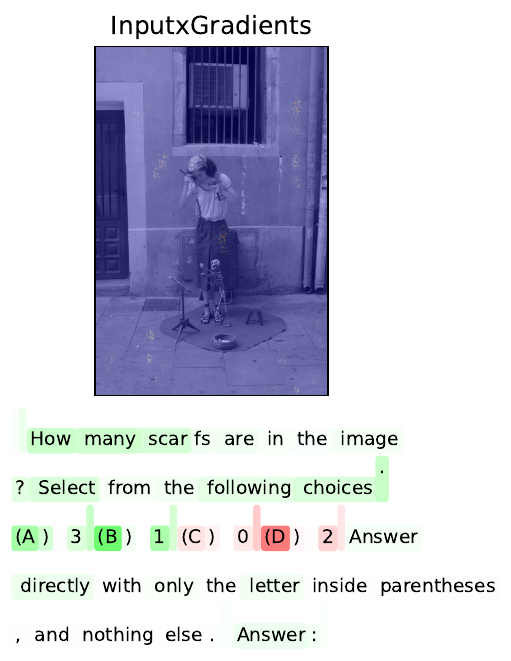}
    \centerline{\small (c) Input$\times$Grad}
  \end{minipage}\hfill
  \begin{minipage}[b]{0.19\linewidth}
    \centering
    \includegraphics[width=\linewidth]{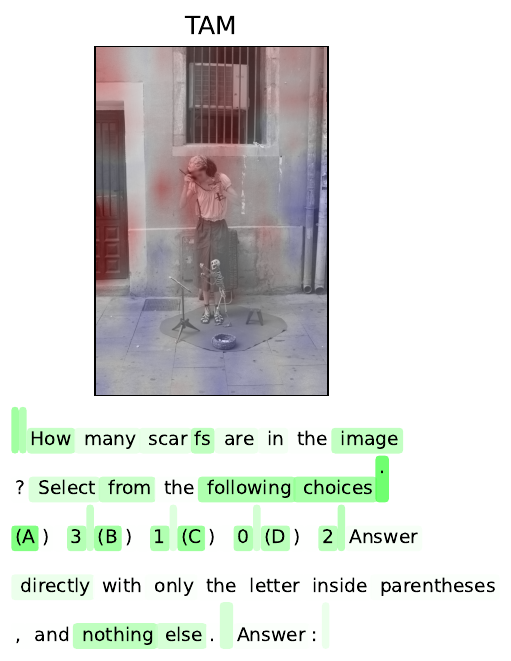}
    \centerline{\small (d) TAM}
  \end{minipage}\hfill
  \begin{minipage}[b]{0.19\linewidth}
    \centering
    \includegraphics[width=\linewidth]{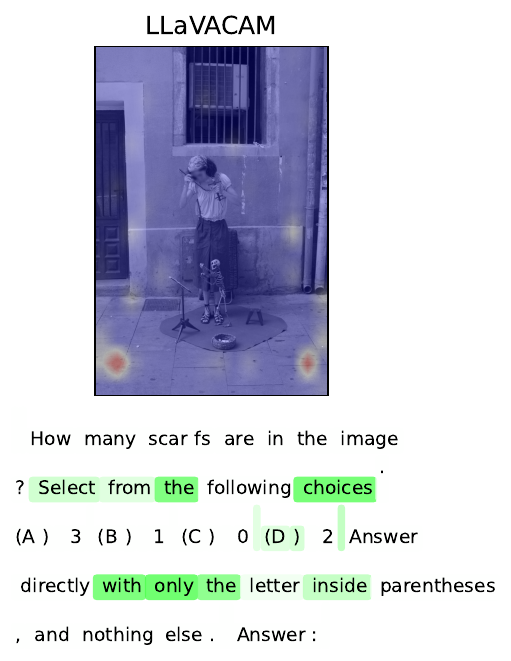}
    \centerline{\small (e) LLaVA-CAM}
  \end{minipage}

  \vspace{0.3cm} % Space between rows

  % --- ROW 2 (5 Panels) ---
  \begin{minipage}[b]{0.19\linewidth}
    \centering
    \includegraphics[width=\linewidth]{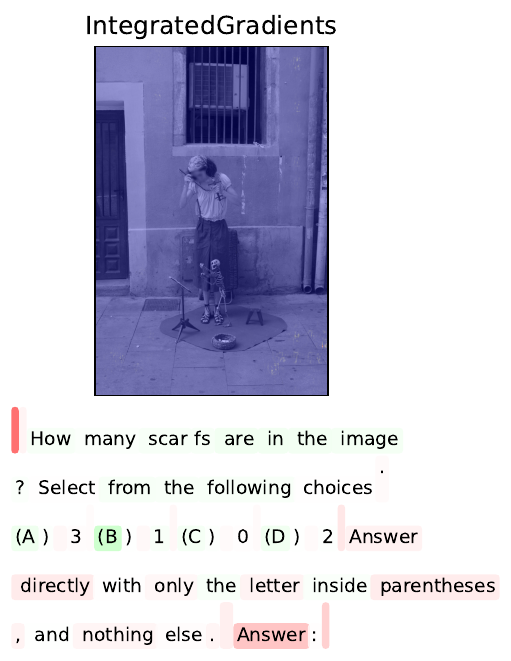}
    \centerline{\small (f) Integrated Gradients}
  \end{minipage}\hfill
  \begin{minipage}[b]{0.19\linewidth}
    \centering
    \includegraphics[width=\linewidth]{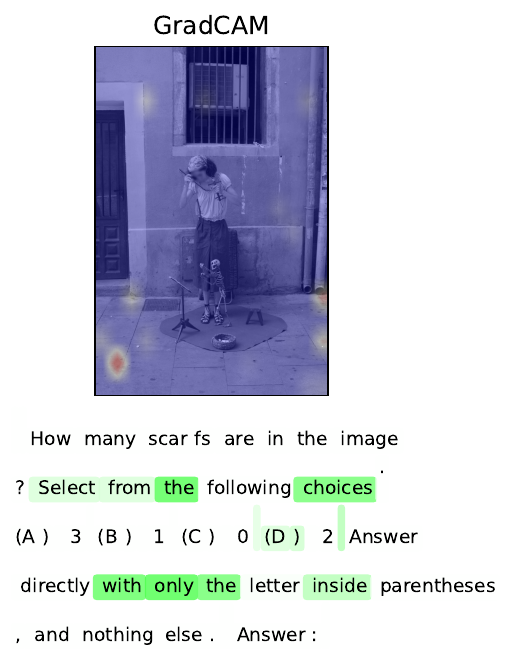}
    \centerline{\small (g) GradCAM}
  \end{minipage}\hfill
  \begin{minipage}[b]{0.19\linewidth}
    \centering
    \includegraphics[width=\linewidth]{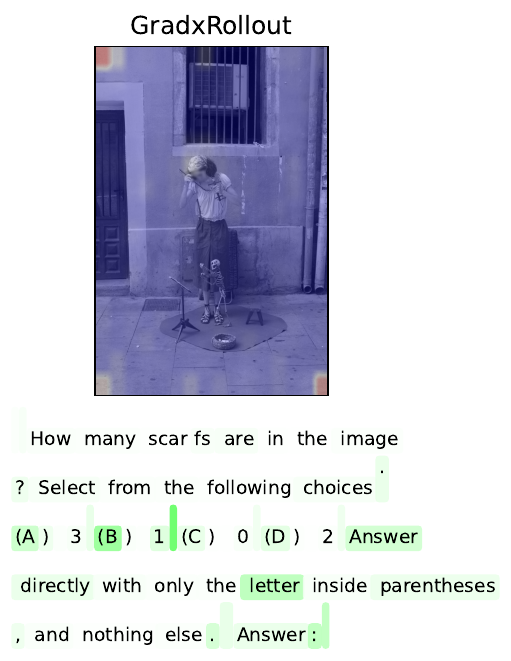}
    \centerline{\small (h) GradxRollout}
  \end{minipage}\hfill
  \begin{minipage}[b]{0.19\linewidth}
    \centering
    \includegraphics[width=\linewidth]{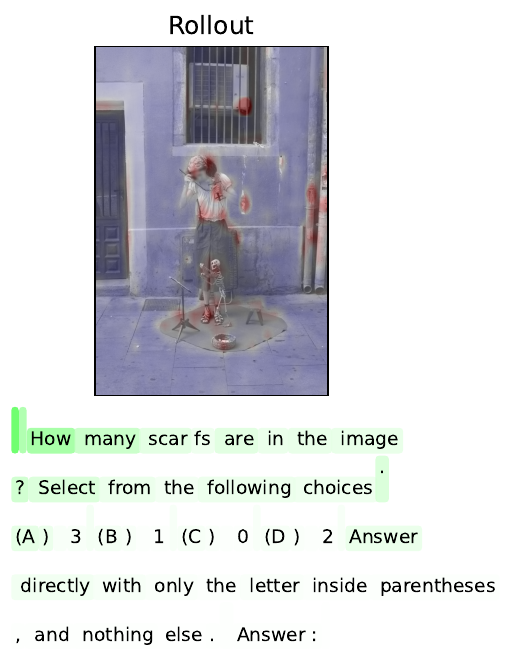}
    \centerline{\small (i) Rollout}
  \end{minipage}\hfill
  \begin{minipage}[b]{0.19\linewidth}
    \centering
    \includegraphics[width=\linewidth]{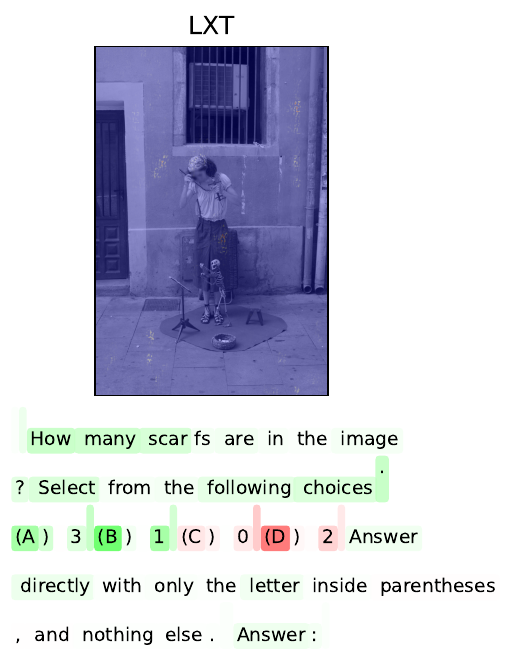}
    \centerline{\small (j) AttnLRP}
  \end{minipage}
  
  %\caption{Example from CVBench dataset with InternVL model.}
  \caption{Example on CVBench (InternVL2-2B). The prompt queries the quantity of "scarfs" (a small, localized accessory). However, VLM-native explainers like TAM and LLaVA-CAM completely fail to ground the textual concept. Instead, they are distracted by the most visually prominent elements in the scene, placing massive attribution weight on the bright red circular mat on the ground and the central torso. This demonstrates a bias toward raw visual salience (color contrast and size) over true cross-modal synergistic grounding.}  
  \label{fig:qual_cvbench_internvl}
\end{figure*}

\begin{figure*}[htbp] 
  \centering
  
  \textbf{Prompt:} \textit{"Is there a person in the image?"} \\
  \textbf{Answer:} \textit{Yes}
  \vspace{0.2cm}
  
  % --- ROW 1 (5 Panels) ---
  \begin{minipage}[b]{0.19\linewidth}
    \centering
    % Force the height to exactly match the image block of the others.
    % (Tweak the 4.0cm up or down until it lines up perfectly)
    \includegraphics[width=\linewidth]{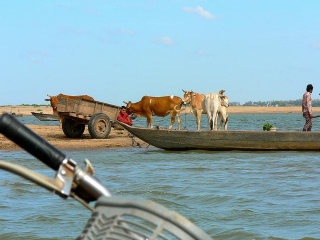} \\
    \vspace{0.8cm} % Tweak this to align "(a)" with "(b)"
    \centerline{\small (a) Original Input}
  \end{minipage}\hfill
  \begin{minipage}[b]{0.19\linewidth}
    \centering
    \includegraphics[width=\linewidth]{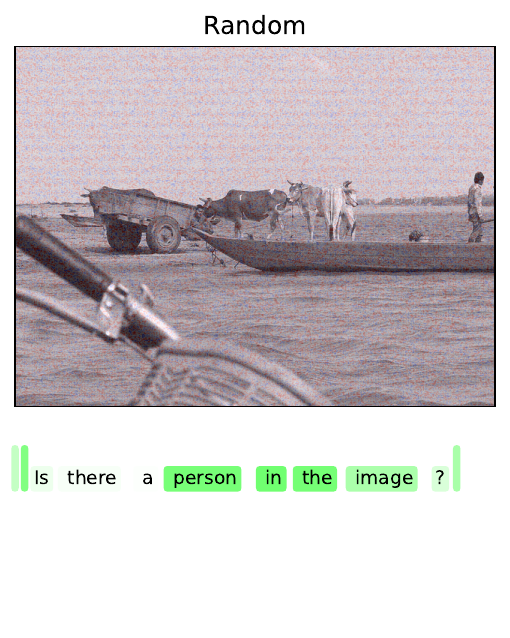}
    \centerline{\small (b) Random}
  \end{minipage}\hfill
  \begin{minipage}[b]{0.19\linewidth}
    \centering
    \includegraphics[width=\linewidth]{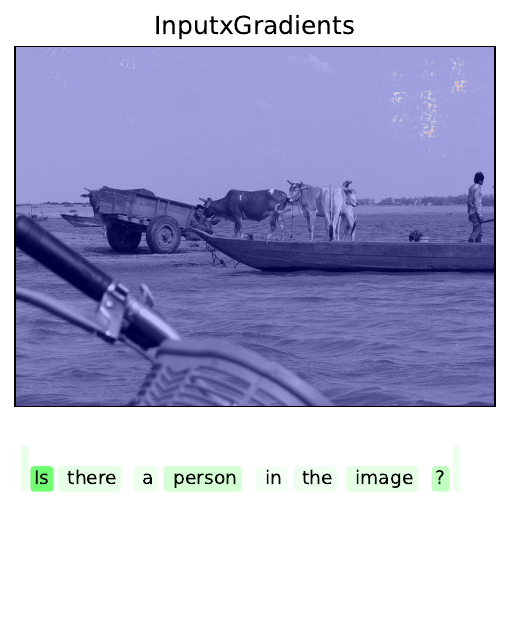}
    \centerline{\small (c) Input$\times$Grad}
  \end{minipage}\hfill
  \begin{minipage}[b]{0.19\linewidth}
    \centering
    \includegraphics[width=\linewidth]{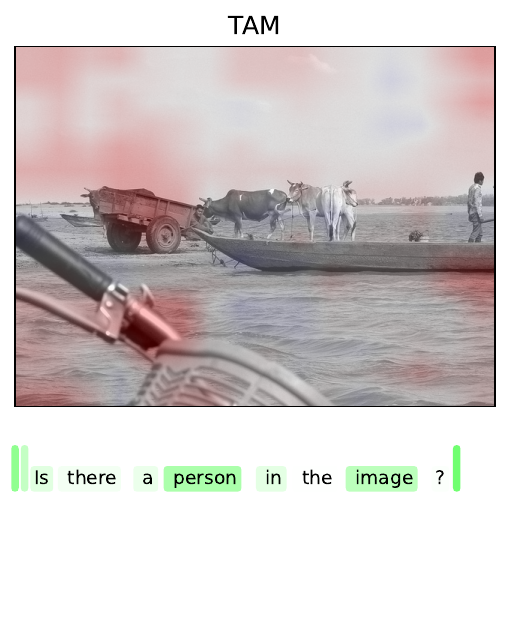}
    \centerline{\small (d) TAM}
  \end{minipage}\hfill
  \begin{minipage}[b]{0.19\linewidth}
    \centering
    \includegraphics[width=\linewidth]{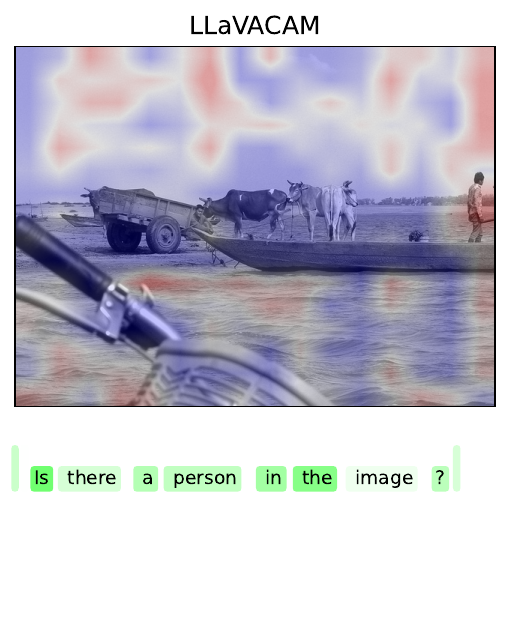}
    \centerline{\small (e) LLaVA-CAM}
  \end{minipage}

  \vspace{0.3cm} % Space between rows

  % --- ROW 2 (5 Panels) ---
  \begin{minipage}[b]{0.19\linewidth}
    \centering
    \includegraphics[width=\linewidth]{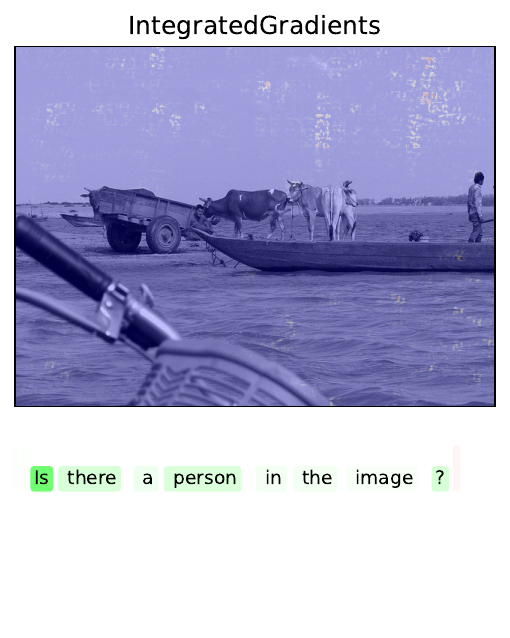}
    \centerline{\small (f) Integrated Gradients}
  \end{minipage}\hfill
  \begin{minipage}[b]{0.19\linewidth}
    \centering
    \includegraphics[width=\linewidth]{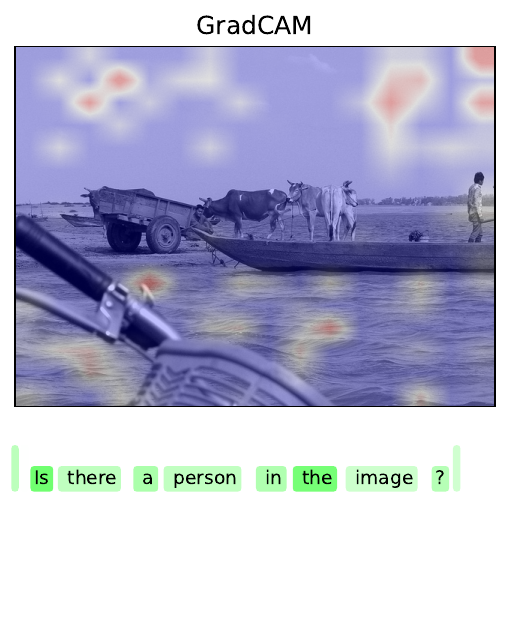}
    \centerline{\small (g) GradCAM}
  \end{minipage}\hfill
  \begin{minipage}[b]{0.19\linewidth}
    \centering
    \includegraphics[width=\linewidth]{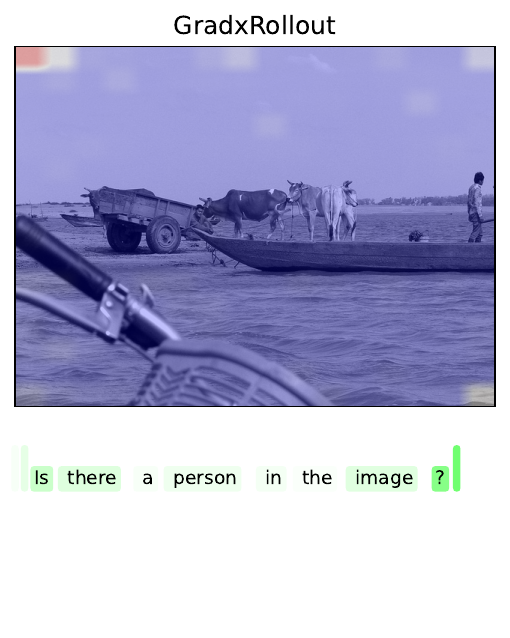}
    \centerline{\small (h) GradxRollout}
  \end{minipage}\hfill
  \begin{minipage}[b]{0.19\linewidth}
    \centering
    \includegraphics[width=\linewidth]{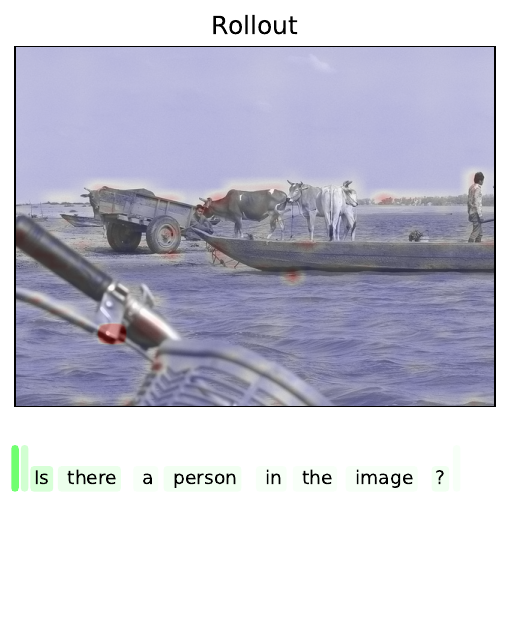}
    \centerline{\small (i) Rollout}
  \end{minipage}\hfill
  \begin{minipage}[b]{0.19\linewidth}
    \centering
    \includegraphics[width=\linewidth]{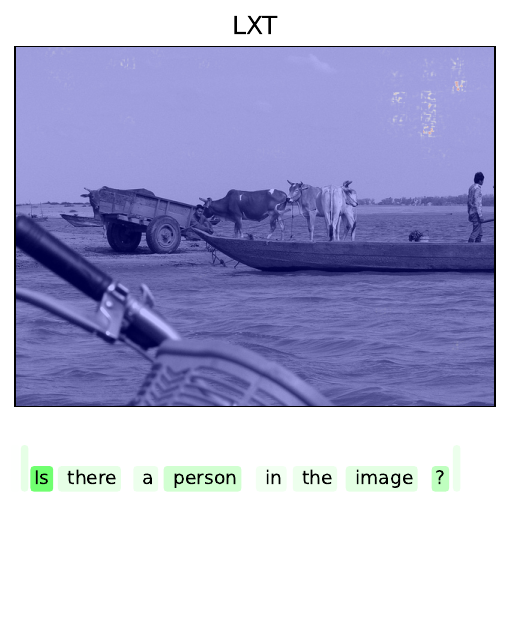}
    \centerline{\small (j) AttnLRP}
  \end{minipage}
  \caption{Example from RePOPE (InternVL-2B). The prompt asks to verify the presence of a "person," who is located on the far right periphery of the boat. Baseline explainers exhibit a massive center-object bias. TAM and LLaVA-CAM place almost all attribution weight on the central cattle, while Rollout focuses on the high-contrast foreground mechanism.}
  % \caption{Example from RePOPE dataset with InternVL model.}
  \label{fig:qual_repope_all}
\end{figure*}

\begin{figure*}[htbp] 
  \centering
  
  \textbf{Prompt:} \textit{"How many light switchs are in the image? Select from the following choices. (A) 1 (B) 0 (C) 2 (D) Answer directly with only the letter inside parentheses, and nothing else. Answer :"} \\
  \textbf{Answer:} \textit{B}
  \vspace{0.2cm}
  
  % --- ROW 1 (5 Panels) ---
  \begin{minipage}[b]{0.19\linewidth}
    \centering
    % Force the height to exactly match the image block of the others.
    % (Tweak the 4.0cm up or down until it lines up perfectly)
    \includegraphics[width=\linewidth]{figs/figs_quality/qual_19_original.pdf} \\
    \vspace{0.8cm} % Tweak this to align "(a)" with "(b)"
    \centerline{\small (a) Original Input}
  \end{minipage}\hfill
  \begin{minipage}[b]{0.19\linewidth}
    \centering
    \includegraphics[width=\linewidth]{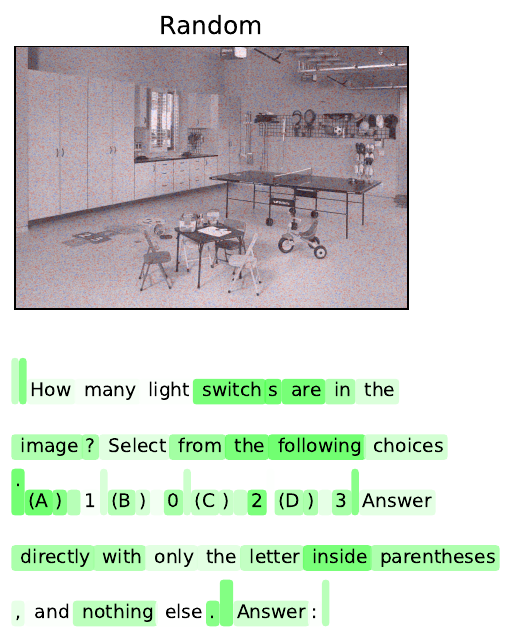}
    \centerline{\small (b) Random}
  \end{minipage}\hfill
  \begin{minipage}[b]{0.19\linewidth}
    \centering
    \includegraphics[width=\linewidth]{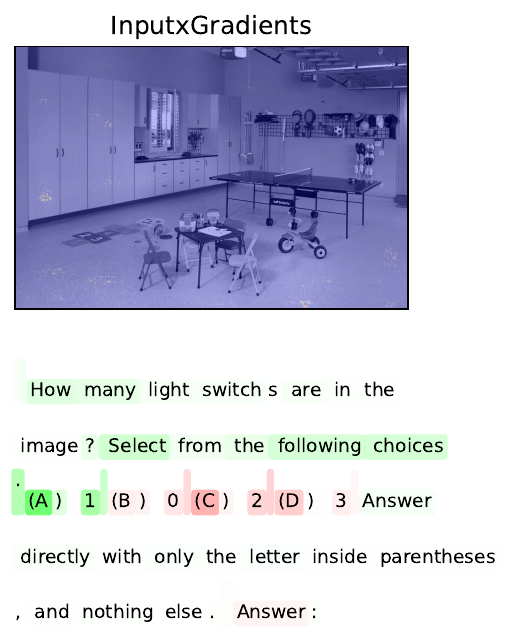}
    \centerline{\small (c) Input$\times$Grad}
  \end{minipage}\hfill
  \begin{minipage}[b]{0.19\linewidth}
    \centering
    \includegraphics[width=\linewidth]{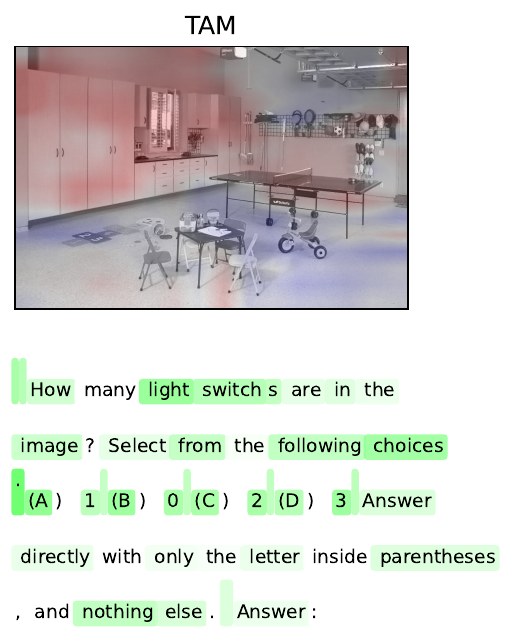}
    \centerline{\small (d) TAM}
  \end{minipage}\hfill
  \begin{minipage}[b]{0.19\linewidth}
    \centering
    \includegraphics[width=\linewidth]{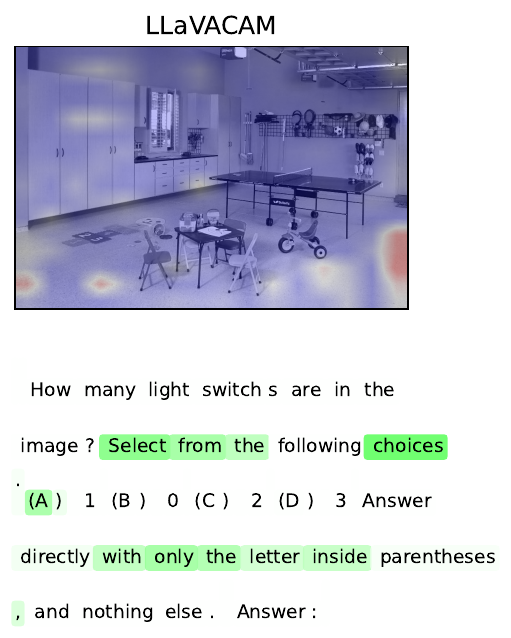}
    \centerline{\small (e) LLaVA-CAM}
  \end{minipage}

  \vspace{0.3cm} % Space between rows

  % --- ROW 2 (5 Panels) ---
  \begin{minipage}[b]{0.19\linewidth}
    \centering
    \includegraphics[width=\linewidth]{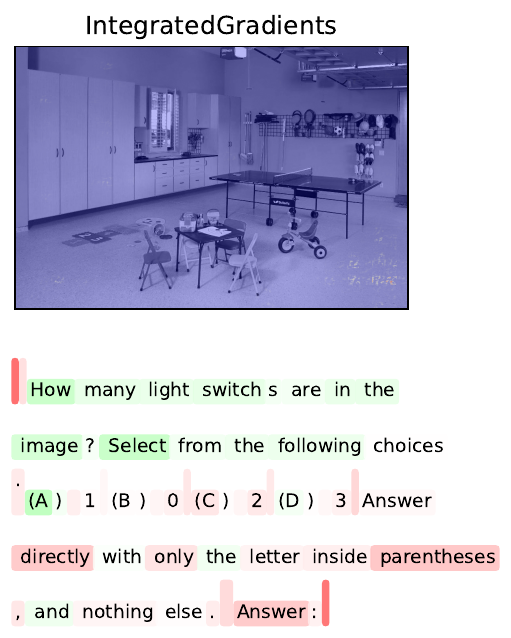}
    \centerline{\small (f) Integrated Gradients}
  \end{minipage}\hfill
  \begin{minipage}[b]{0.19\linewidth}
    \centering
    \includegraphics[width=\linewidth]{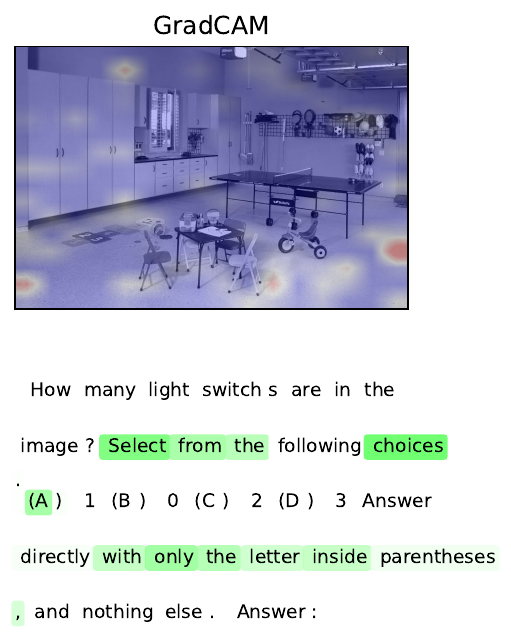}
    \centerline{\small (g) GradCAM}
  \end{minipage}\hfill
  \begin{minipage}[b]{0.19\linewidth}
    \centering
    \includegraphics[width=\linewidth]{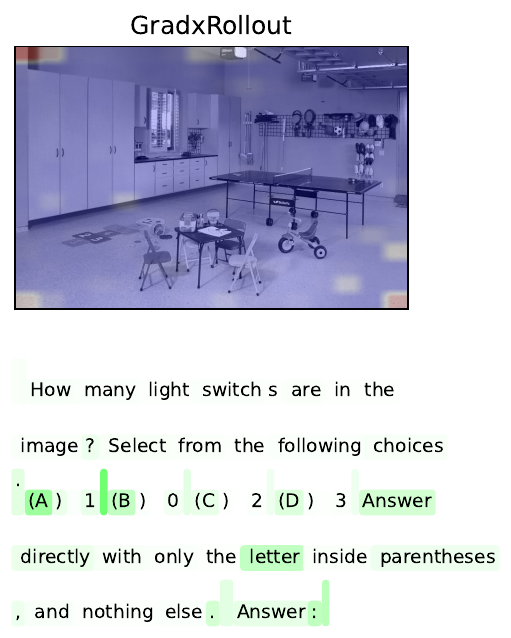}
    \centerline{\small (h) GradxRollout}
  \end{minipage}\hfill
  \begin{minipage}[b]{0.19\linewidth}
    \centering
    \includegraphics[width=\linewidth]{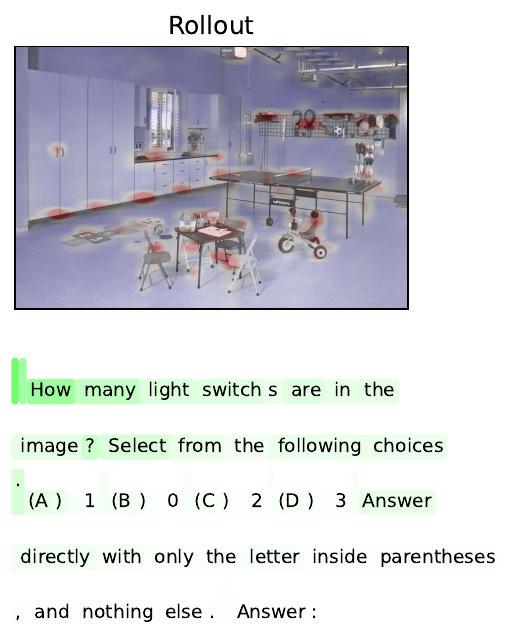}
    \centerline{\small (i) Rollout}
  \end{minipage}\hfill
  \begin{minipage}[b]{0.19\linewidth}
    \centering
    \includegraphics[width=\linewidth]{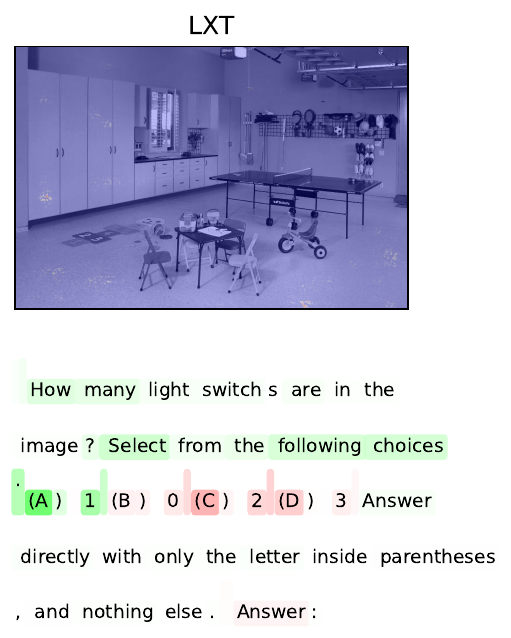}
    \centerline{\small (j) AttnLRP}
  \end{minipage}
  
  %\caption{Contradiction example on CVBench (InternVL2-2B)}
   \caption{Example from CVBench (InternVL2-2B). The prompt requires localizing highly specific, small-scale objects ("light switches"). However, gradient-based baseline methods (e.g., Input$\times$Grad, Integrated Gradients) devolve into naive edge detectors. They trace the broad structural contours of the room, highlighting the edges of the cabinets and tables rather than the queried objects.}
  \label{fig:qual_cv_bench_contradict}
\end{figure*}

\begin{figure*}[htbp] 
  \centering
  
  \textbf{Prompt:} \textit{"Which option describe the object relationship in the image correctly?Options: A: The suitcase is on the book., B: The suitcase is beneath the cat., C: The suitcase is beneath the bed., D: The suitcase is beneath the book Answer directly with only the letter inside parentheses, and nothing else. Answer :"} \\
  \textbf{Answer:} \textit{A.}
  \vspace{0.2cm}
  
  % --- ROW 1 (5 Panels) ---
  \begin{minipage}[b]{0.19\linewidth}
    \centering
    % Force the height to exactly match the image block of the others.
    % (Tweak the 4.0cm up or down until it lines up perfectly)
    \includegraphics[width=\linewidth]{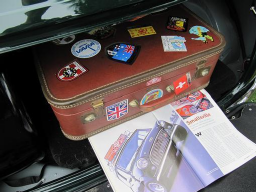} \\
    \vspace{0.8cm} % Tweak this to align "(a)" with "(b)"
    \centerline{\small (a) Original Input}
  \end{minipage}\hfill
  \begin{minipage}[b]{0.19\linewidth}
    \centering
    \includegraphics[width=\linewidth]{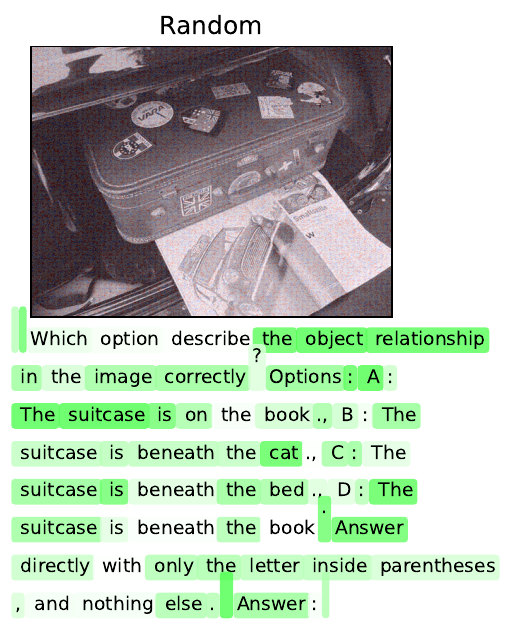}
    \centerline{\small (b) Random}
  \end{minipage}\hfill
  \begin{minipage}[b]{0.19\linewidth}
    \centering
    \includegraphics[width=\linewidth]{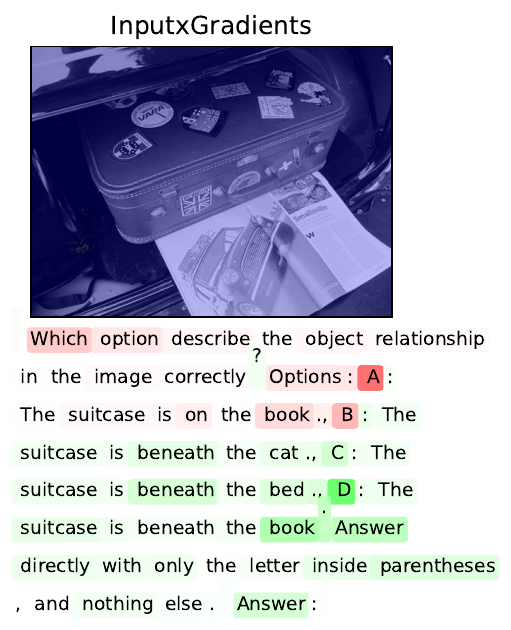}
    \centerline{\small (c) Input$\times$Grad}
  \end{minipage}\hfill
  \begin{minipage}[b]{0.19\linewidth}
    \centering
    \includegraphics[width=\linewidth]{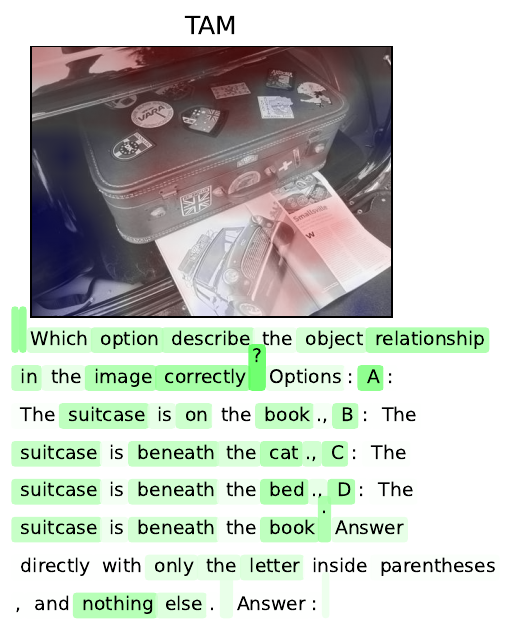}
    \centerline{\small (d) TAM}
  \end{minipage}\hfill
  \begin{minipage}[b]{0.19\linewidth}
    \centering
    \includegraphics[width=\linewidth]{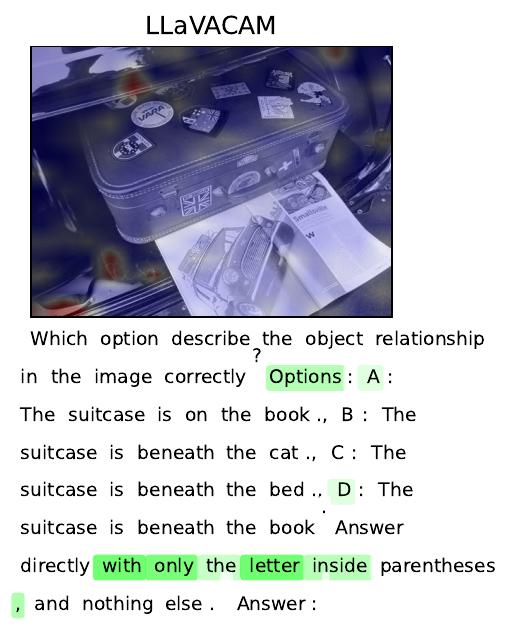}
    \centerline{\small (e) LLaVA-CAM}
  \end{minipage}

  \vspace{0.3cm} % Space between rows

  % --- ROW 2 (5 Panels) ---
  \begin{minipage}[b]{0.19\linewidth}
    \centering
    \includegraphics[width=\linewidth]{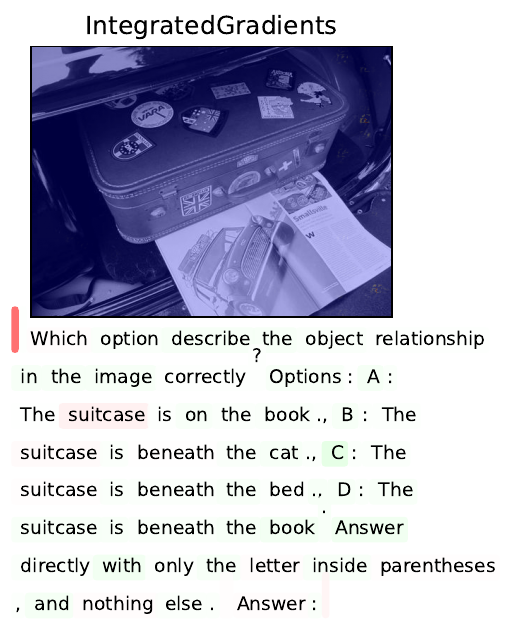}
    \centerline{\small (f) Integrated Gradients}
  \end{minipage}\hfill
  \begin{minipage}[b]{0.19\linewidth}
    \centering
    \includegraphics[width=\linewidth]{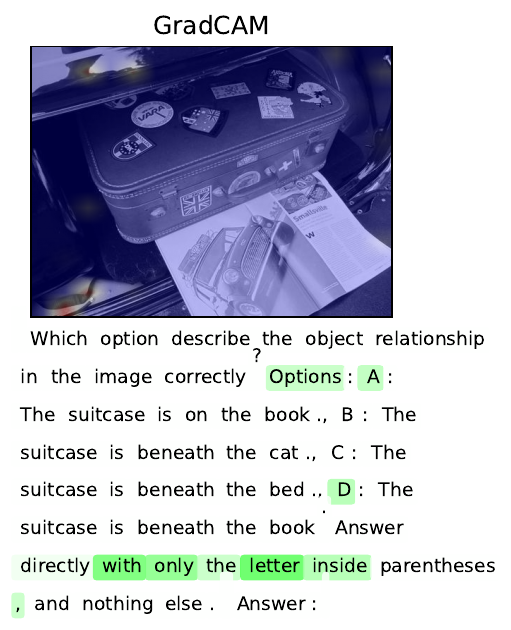}
    \centerline{\small (g) GradCAM}
  \end{minipage}\hfill
  \begin{minipage}[b]{0.19\linewidth}
    \centering
    \includegraphics[width=\linewidth]{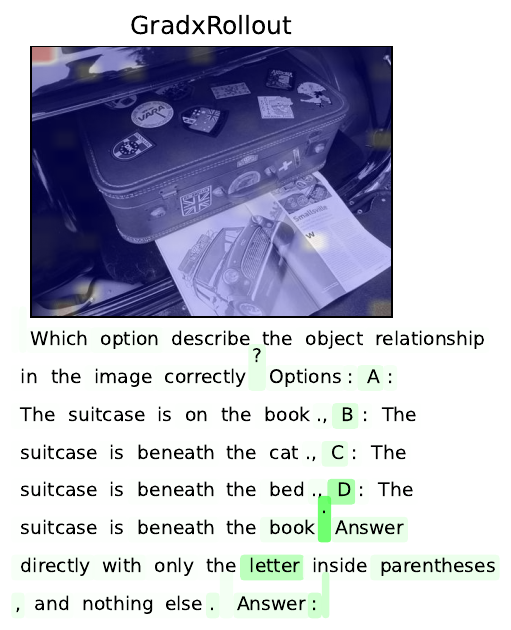}
    \centerline{\small (h) GradxRollout}
  \end{minipage}\hfill
  \begin{minipage}[b]{0.19\linewidth}
    \centering
    \includegraphics[width=\linewidth]{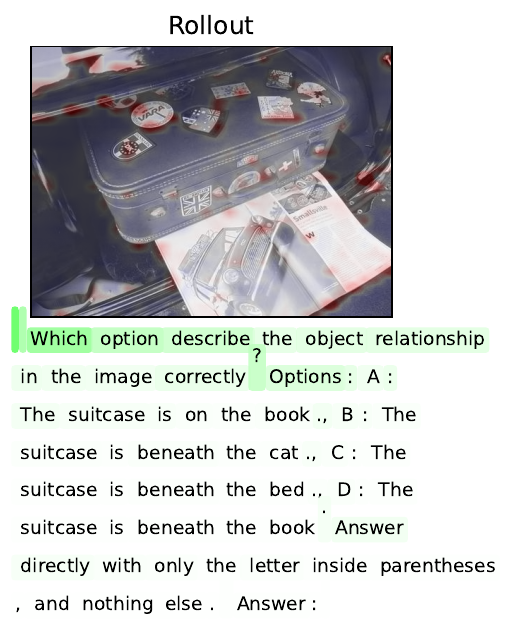}
    \centerline{\small (i) Rollout}
  \end{minipage}\hfill
  \begin{minipage}[b]{0.19\linewidth}
    \centering
    \includegraphics[width=\linewidth]{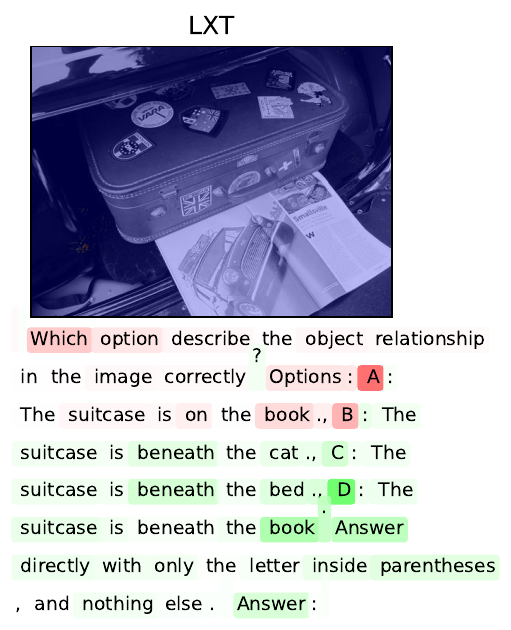}
    \centerline{\small (j) AttnLRP}
  \end{minipage}
  
  \caption{Failure example from MMStar dataset (InternVL).}
    \caption{Failure on Spatial/Relational Reasoning (MMStar, InternVL2-2B). The prompt requires evaluating an abstract spatial relationship between two entities ("suitcase" and "book"). Unlike simple object identification tasks, explaining relational logic requires grounding the physical intersection and relative positioning of the objects. Here, all evaluated explainers universally fail to provide localized reasoning.}
  \label{fig:qual_failure}
\end{figure*}

To contextualize the quantitative results of our synergistic faithfulness metric ($\mathcal{F}_{syn}$), we present a qualitative analysis of the attributions generated by the evaluated explainers. By visualizing the exact cross-modal grounding, we identify three recurring behavioral archetypes that explain the divergence between unimodal and multimodal evaluation metrics: the vulnerability of VLM-native methods to visual salience traps (see Fig. \ref{fig:qual_cvbench_internvl}, the tendency of gradient methods to act as edge-detectors (yielding unimodal contradictions) (see Fig. \ref{fig:qual_cv_bench_contradict} and Fig. \ref{fig:qual_repope_all}), and the failure of current XAI techniques on highly abstract reasoning tasks (see Fig. \ref{fig:qual_failure}).
Finally, we note that while AttnLRP achieves strong synergistic grounding, its raw visual heatmaps often lack sharp localized resolution and appear visually unappealing; this aesthetic limitation is a direct artifact of the implementation constraints detailed in Appendix \ref{appendix_reproducibility}.

\clearpage
\section*{NeurIPS Paper Checklist}

\begin{enumerate}

\item {\bf Claims}
    \item[] Question: Do the main claims made in the abstract and introduction accurately reflect the paper's contributions and scope?
    \item[] Answer: \answerYes{}
    \item[] Justification: The empirical results in Section \ref{experiments} support our claims in the abstract and introduction.

\item {\bf Limitations}
    \item[] Question: Does the paper discuss the limitations of the work performed by the authors?
    \item[] Answer: \answerYes{}
    \item[] Justification: We discuss the limitation of the works in Section \ref{discussion}.

\item {\bf Theory assumptions and proofs}
    \item[] Question: For each theoretical result, does the paper provide the full set of assumptions and a complete (and correct) proof?
    \item[] Answer: \answerNA{} % Replace by \answerYes{}, \answerNo{}, or \answerNA{}.
    \item[] Justification: The paper does not include theoretical results.

    \item {\bf Experimental result reproducibility}
    \item[] Question: Does the paper fully disclose all the information needed to reproduce the main experimental results of the paper to the extent that it affects the main claims and/or conclusions of the paper (regardless of whether the code and data are provided or not)?
    \item[] Answer: \answerYes{} % Replace by \answerYes{}, \answerNo{}, or \answerNA{}.
    \item[] Justification: Full reproducibility details, including explainer configurations, model adaptations, generation parameters, and dataset sampling are rigorously provided in Appendix \ref{appendix_reproducibility}.

\item {\bf Open access to data and code}
    \item[] Question: Does the paper provide open access to the data and code, with sufficient instructions to faithfully reproduce the main experimental results, as described in supplemental material?
    \item[] Answer: \answerYes{} % Replace by \answerYes{}, \answerNo{}, or \answerNA{}.
    \item[] Justification: We provide the full code and the data to faithfully reproduce the experimental results.

\item {\bf Experimental setting/details}
    \item[] Question: Does the paper specify all the training and test details (e.g., data splits, hyperparameters, how they were chosen, type of optimizer) necessary to understand the results?
    \item[] Answer: \answerYes{} % Replace by \answerYes{}, \answerNo{}, or \answerNA{}.
    \item[] Justification: All hyperparameter details are specified in Appendix \ref{appendix_reproducibility}.

\item {\bf Experiment statistical significance}
    \item[] Question: Does the paper report error bars suitably and correctly defined or other appropriate information about the statistical significance of the experiments?
    \item[] Answer: \answerYes{} % Replace by \answerYes{}, \answerNo{}, or \answerNA{}.
    \item[] Justification: We report error bars in the figures and conduct statistical significance tests for the metric rankings or the explainer ranking using LMM plots (with p-values). See Appendix \ref{appendix_statistical} for additional results.

\item {\bf Experiments compute resources}
    \item[] Question: For each experiment, does the paper provide sufficient information on the computer resources (type of compute workers, memory, time of execution) needed to reproduce the experiments?
    \item[] Answer: \answerYes{} % Replace by \answerYes{}, \answerNo{}, or \answerNA{}.
    \item[] Justification: Hardware specifications (e.g., GPU models) and the estimated computational requirements are detailed in Appendix \ref{appendix_reproducibility}.
    
\item {\bf Code of ethics}
    \item[] Question: Does the research conducted in the paper conform, in every respect, with the NeurIPS Code of Ethics \url{https://neurips.cc/public/EthicsGuidelines}?
    \item[] Answer: \answerYes{} % Replace by \answerYes{}, \answerNo{}, or \answerNA{}.
    \item[] Justification: The research methodology, dataset usage, and evaluations comply fully with the NeurIPS Code of Ethics.

\item {\bf Broader impacts}
    \item[] Question: Does the paper discuss both potential positive societal impacts and negative societal impacts of the work performed?
    \item[] Answer: \answerYes{} % Replace by \answerYes{}, \answerNo{}, or \answerNA{}.
    \item[] Justification: We provide the information in Section \ref{discussion}.
    
\item {\bf Safeguards}
    \item[] Question: Does the paper describe safeguards that have been put in place for responsible release of data or models that have a high risk for misuse (e.g., pre-trained language models, image generators, or scraped datasets)?
    \item[] Answer: \answerNA{} % Replace by \answerYes{}, \answerNo{}, or \answerNA{}.
    \item[] Justification: \answerNA{}

\item {\bf Licenses for existing assets}
    \item[] Question: Are the creators or original owners of assets (e.g., code, data, models), used in the paper, properly credited and are the license and terms of use explicitly mentioned and properly respected?
    \item[] Answer: \answerYes{} % Replace by \answerYes{}, \answerNo{}, or \answerNA{}.
    \item[] Justification: All evaluated open-weight models and utilized datasets are properly cited, and our codebase is packaged with a CC BY 4.0 license respecting upstream dependencies.

\item {\bf New assets}
    \item[] Question: Are new assets introduced in the paper well documented and is the documentation provided alongside the assets?
    \item[] Answer: \answerYes{} % Replace by \answerYes{}, \answerNo{}, or \answerNA{}.
    \item[] Justification: The proposed benchmark suite and metric are thoroughly documented via comprehensive README files, configuration files, and instructions provided in the supplemental code repository.

\item {\bf Crowdsourcing and research with human subjects}
    \item[] Question: For crowdsourcing experiments and research with human subjects, does the paper include the full text of instructions given to participants and screenshots, if applicable, as well as details about compensation (if any)? 
    \item[] Answer: \answerNA{} % Replace by \answerYes{}, \answerNo{}, or \answerNA{}.
    \item[] Justification: \answerNA{}

\item {\bf Institutional review board (IRB) approvals or equivalent for research with human subjects}
    \item[] Question: Does the paper describe potential risks incurred by study participants, whether such risks were disclosed to the subjects, and whether Institutional Review Board (IRB) approvals (or an equivalent approval/review based on the requirements of your country or institution) were obtained?
    \item[] Answer: \answerNA{} % Replace by \answerYes{}, \answerNo{}, or \answerNA{}.
    \item[] Justification: \answerNA{}

\item {\bf Declaration of LLM usage}
    \item[] Question: Does the paper describe the usage of LLMs if it is an important, original, or non-standard component of the core methods in this research? Note that if the LLM is used only for writing, editing, or formatting purposes and does \emph{not} impact the core methodology, scientific rigor, or originality of the research, declaration is not required.
    %this research? 
    \item[] Answer: \answerNA{} % Replace by \answerYes{}, \answerNo{}, or \answerNA{}.
    \item[] Justification: \answerNA{}

\end{enumerate}

\end{document}